%% file: main.tex
\documentclass{article}
\let\CBPOAddContentsLine\addcontentsline
\usepackage{iclr2027_conference,times}
\let\addcontentsline\CBPOAddContentsLine
\iclrfinalcopy

\usepackage[T1]{fontenc}
\usepackage[utf8]{inputenc}
\usepackage[hyphens]{url}
\usepackage{graphicx}
\usepackage{amsmath}
\usepackage{amsfonts}
\usepackage{booktabs}
\usepackage{array}
\usepackage{tabularx}
\usepackage{multirow}
\usepackage{placeins}
\usepackage{float}
\usepackage{wrapfig}
\usepackage{algorithm}
\usepackage{algpseudocode}
\usepackage{fvextra}
\usepackage[most]{tcolorbox}
\usepackage{hyperref}
\hypersetup{
  colorlinks=true,
  citecolor=green!40!black,
  linkcolor=black,
  urlcolor=black
}

\makeatletter
\renewcommand\section{\@startsection{section}{1}{\z@}%
  {-0.55ex plus -0.10ex minus -0.05ex}%
  {0.20ex plus 0.05ex minus 0.05ex}%
  {\large\sc\raggedright}}
\renewcommand\subsection{\@startsection{subsection}{2}{\z@}%
  {-0.50ex plus -0.10ex minus -0.05ex}%
  {0.18ex plus 0.05ex minus 0.05ex}%
  {\normalsize\sc\raggedright}}
\renewcommand\paragraph{\@startsection{paragraph}{4}{\z@}%
  {1.0ex plus 0.2ex minus 0.1ex}%
  {-1em}%
  {\normalsize\bf}}
\makeatother

\renewcommand{\arraystretch}{1.15}
\newcommand{\metrichead}[1]{\scalebox{0.85}{\texttt{#1}}}
\newfont{\seedpmfont}{cmsy5 at 4pt}
\newcommand{\seedscore}[2]{\mbox{\fontsize{6}{7}\selectfont#1{\seedpmfont\char6}#2}}
\newsavebox{\CBPOFirstFigureBox}
\newlength{\CBPOFirstFigureBottom}

\DeclareUnicodeCharacter{00B1}{\ensuremath{\pm}}
\DeclareUnicodeCharacter{00B2}{\ensuremath{{}^{2}}}
\DeclareUnicodeCharacter{00D7}{\ensuremath{\times}}
\DeclareUnicodeCharacter{00F7}{\ensuremath{\div}}
\DeclareUnicodeCharacter{03C0}{\ensuremath{\pi}}
\DeclareUnicodeCharacter{2081}{\ensuremath{{}_{1}}}
\DeclareUnicodeCharacter{2082}{\ensuremath{{}_{2}}}
\DeclareUnicodeCharacter{2083}{\ensuremath{{}_{3}}}
\DeclareUnicodeCharacter{2192}{\ensuremath{\to}}
\DeclareUnicodeCharacter{21D2}{\ensuremath{\Rightarrow}}
\DeclareUnicodeCharacter{2208}{\ensuremath{\in}}
\DeclareUnicodeCharacter{221A}{\ensuremath{\surd}}
\DeclareUnicodeCharacter{2260}{\ensuremath{\ne}}
\DeclareUnicodeCharacter{2264}{\ensuremath{\le}}
\DeclareUnicodeCharacter{2265}{\ensuremath{\ge}}

\title{Cross-Behavior Policy Optimization: Unified Learning from Offline Demonstrations and On- and Off-Policy Rollouts via a Shared Regression Target}

\author{
    Du Cheng, Junyu Lu, Jingfan Zhang, Haoxuan Jia, Shichao Weng
}

\begin{document}
\maketitle
\fancyhead{}

\addtocontents{toc}{\protect\setcounter{tocdepth}{-1}}

\input{sections/0_abstract}

\vspace{-20pt}
\sbox{\CBPOFirstFigureBox}{\includegraphics[width=0.90\columnwidth]{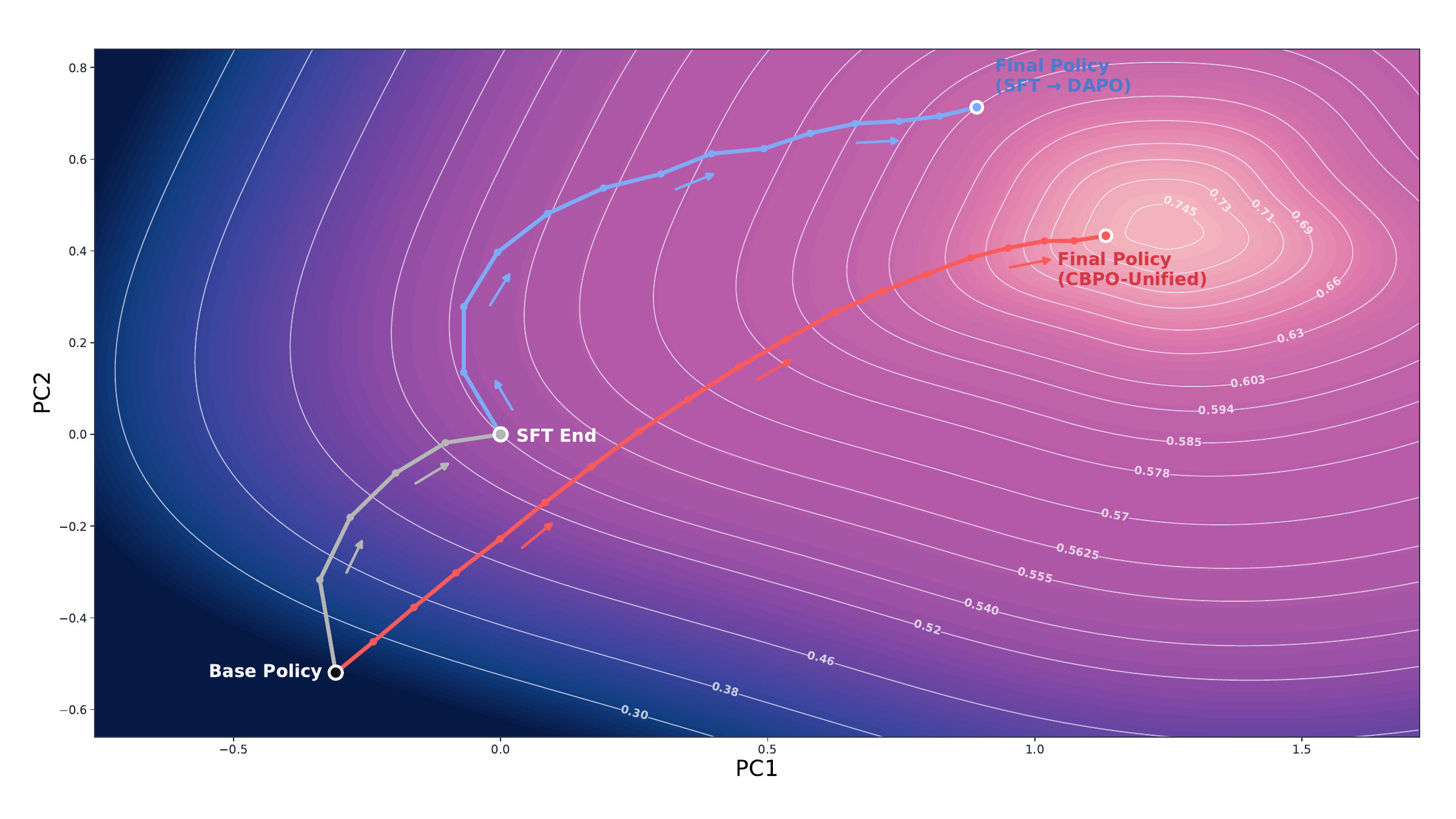}}
\ifdim\dimexpr\pagegoal-\pagetotal\relax<\dimexpr\ht\CBPOFirstFigureBox+2\intextsep+3\baselineskip\relax
  \newpage
\fi
\setlength{\CBPOFirstFigureBottom}{\dimexpr\pagetotal+\ht\CBPOFirstFigureBox-18pt\relax}
\AddToShipoutPictureBG*{%
  \AtTextUpperLeft{%
    \put(\LenToUnit{0.05\columnwidth},\LenToUnit{-\CBPOFirstFigureBottom}){\usebox{\CBPOFirstFigureBox}}%
  }%
}
\begin{figure}[H]
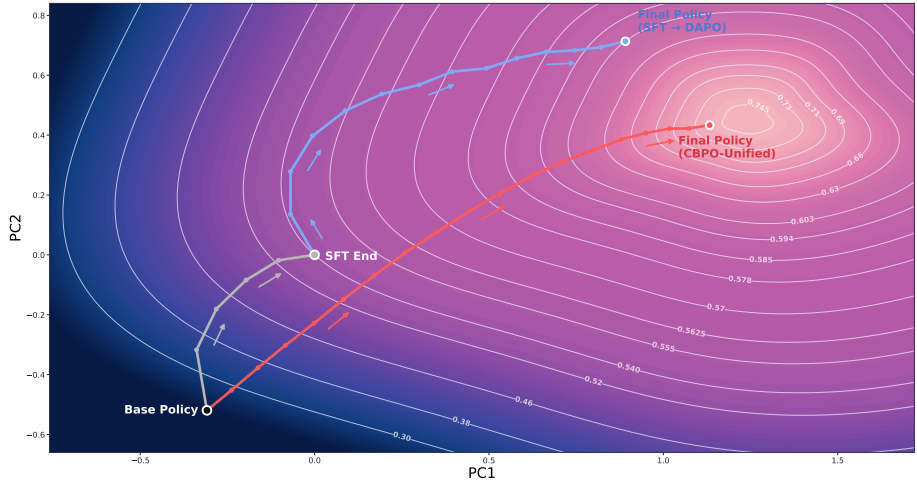

  \centering
  \phantom{\usebox{\CBPOFirstFigureBox}}
  \vspace{-20pt}
  \caption{CBPO-Unified/SFT$\rightarrow$DAPO: $L/D=1.0205/1.2207$, with total turn $=58.20^\circ/368.75^\circ$.}
  \label{fig:optimization_trajectories}
\end{figure}
\vspace{-15pt}

\input{sections/1_introduction}

\input{sections/2_related_work}

\input{sections/3_preliminaries}

\input{sections/4_methods}

\input{sections/5_experiments}

\input{sections/6_conclusion}

\input{sections/7_limitations}

\clearpage
\subsection*{Ethics Statement}
This work uses existing mathematical-reasoning datasets and model-generated trajectories. No new human-subject data were collected.
Confirmed overlaps between training questions and held-out evaluation problems were removed before optimization (Appendix~\ref{app:decontamination}).
Use of the source models and datasets is subject to their respective licenses and terms.

\subsection*{Reproducibility Statement}
Section~\ref{sec:experimental_setup} specifies the model scales, data sources, reward values, training budgets, evaluation sets, and metrics.
Appendix~\ref{app:prompt_reward} describes the artifact sources and verifier, Appendix~\ref{app:decontamination} gives the overlap-screening protocol and retained counts, and Appendices~\ref{app:baseline_details} and~\ref{app:training_system_settings} provide baseline settings, decoding, hardware, and seed usage.
Reproduction requires access to the cited model and dataset artifacts under their respective terms.

\subsection*{AI Use Statement}
Generative AI tools were used for language editing and for assistance with code generation, editing, and debugging. All AI-assisted manuscript edits and code changes were reviewed and verified by the authors.

\bibliography{references}
\bibliographystyle{iclr2027_conference}

\clearpage
\appendix
\input{appendix}

\end{document}

%% file: sections/0_abstract.tex
\begin{abstract}
Reasoning post-training draws on offline demonstrations and policy-generated rollouts, yet conventional supervised fine-tuning (SFT) followed by reinforcement learning (RL) separates their objectives and limits continued reuse of collected responses. We propose \textbf{Cross-Behavior Policy Optimization} (CBPO), which makes these sources compatible with one regression loss through a fixed-reference target. Because the generating policy does not enter this target, positive and negative offline completions, reference responses, and current and historical rollouts can share a common reward-labeled record format and replay buffer. As new responses enter the buffer, optimization retains the same regression objective without coordinating separate SFT and RL objectives or requiring stored behavior-policy probabilities, importance weighting, or a learned critic. Restricting the available sources gives offline, on-policy, and off-policy variants, while CBPO-Unified combines them. Across two model scales and four mathematical reasoning benchmarks, CBPO delivers leading \texttt{avg@32} and \texttt{pass@32} performance across benchmarks and data regimes. On Qwen3-1.7B, CBPO-Unified improves benchmark-averaged \texttt{avg@32} and \texttt{pass@32} over SFT$\rightarrow$DAPO by 2.33 and 4.44 percentage points, respectively. In the replay ablation, retaining historical rollouts improves these metrics by 1.87 and 3.72 percentage points over the no-history configuration.
\end{abstract}

%% file: sections/1_introduction.tex
\section{Introduction}
Large language models (LLMs) have advanced rapidly in mathematical reasoning, code generation, and theorem proving.
Reinforcement learning from verifiable rewards (RLVR) increasingly drives this progress by scoring reasoning trajectories with automatic verifiers \citep{shao2024deepseekmath,deepseekai2025r1,xin2024deepseekprover}.

Reasoning post-training can draw on three data regimes.
\textbf{Offline Demonstrations} are collected before optimization and commonly used for supervised fine-tuning (SFT).
\textbf{On-Policy Rollouts} are sampled from the current model and drive methods including PPO, DAPO, REINFORCE++, and GSPO \citep{schulman2017ppo,yu2025dapo,hu2025reinforceplusplus,zheng2025gspo}.
\textbf{Off-Policy Rollouts} come from earlier policies and can be reused by methods such as TBRM \citep{yuan2025tbrm}.
In practice, conventional reasoning post-training commonly follows a two-stage pipeline consisting of cold-start SFT on positive offline chain-of-thought (CoT) demonstrations followed by RLVR on model-generated rollouts, as in DeepSeekMath and DeepSeek-R1 \citep{shao2024deepseekmath,deepseekai2025r1}.

This pipeline fragments optimization and data use. It assigns offline demonstrations and RL rollouts to different objectives, excludes negative demonstrations from SFT, retires positive demonstrations once RLVR begins, and replaces earlier rollout batches even when their verifier rewards remain informative.
Fig.~\ref{fig:optimization_trajectories} shows the projected paths: SFT$\rightarrow$DAPO turns sharply at the stage boundary (path ratio 1.2207, turning $368.75^\circ$) before reaching a lower-reward endpoint. CBPO-Unified instead follows a more direct path (1.0205, $58.20^\circ$) to a higher-reward endpoint.

The closed-form optimal policy of KL-regularized RL induces a fixed-reference regression target that combines a completion's reference-policy log-probability, scaled verifier reward, and frozen prompt-level normalizer, yet is independent of the behavior policy that generated the completion \citep{rafailov2023dpo,brantley2025astarpo}.

Building on this property, we propose \textbf{Cross-Behavior Policy Optimization} (CBPO), which applies one regression loss to Offline Demonstrations and On- and Off-Policy Rollouts through common reward-labeled records.
CBPO-Offline uses offline demonstrations and reference responses. CBPO-OnPolicy trains on current rollouts only. CBPO-OffPolicy combines reference responses with current and historical rollouts. CBPO-Unified uses all of these sources.
On Qwen3-1.7B and Qwen3-8B, each restricted CBPO variant improves aggregate \texttt{avg@32} over its regime-matched baselines, while CBPO-Unified achieves the best \texttt{avg@32} and \texttt{pass@32} on all four benchmarks.

%% file: sections/2_related_work.tex
\section{Related Work}

\paragraph{KL-regularized objectives and response-level learning.}
KL-regularized control relates rewards to reference policies through exponential reweighting \citep{todorov2007lmdp,levine2018inference}; distribution matching and DPO apply this connection to language-model alignment \citep{korbak2022distributionmatching,rafailov2023dpo}. Both A$^\star$-PO and CBPO use response-level least-squares regression derived from the KL-regularized optimum. In the A$^\star$-PO, value estimation and policy regression use separate coefficients, $\beta_1=0.5$ and $\beta_2=10^{-3}$, whereas CBPO uses one $\beta$ for both. A$^\star$-PO estimates $V^\star(x)$ from reference-policy samples and updates the policy using one newly collected on-policy response per prompt \citep{brantley2025astarpo}. CBPO applies its regression loss to a persistent buffer of complete response records from offline demonstrations, reference responses, and fresh and historical rollouts. DRO offers value- and policy-only objectives for single-trajectory rewards, while Path Consistency Learning (PCL) enforces entropy-regularized consistency over multi-step on- and off-policy paths \citep{richemond-etal-2024-dro,nachum2017bridging}. Preference optimization and relative-reward regression use within-prompt comparisons instead \citep{tang2024gpo,ji2024exo,gao2024rebel}.

\paragraph{SFT--RL post-training.}
SFT on CoT data and distillation are common cold starts; DeepSeek-R1 combines such supervision with RLVR, whereas DAPO applies RL to the base model \citep{shao2024deepseekmath,deepseekai2025r1,yu2025dapo}. BRIDGE coordinates SFT and RL through joint bilevel updates \citep{chen2026bridge}. Prefix-RFT replaces one current-policy rollout with a demonstration prefix followed by a policy-generated continuation \citep{huang2025prefixrft}. CBPO keeps demonstrations and rollouts as complete response records and trains them under one verifier-reward objective.

\paragraph{Offline and off-policy RLVR.}
Trajectory-balance and value-based methods exploit non-current rollouts through asynchronous collection or learned value functions \citep{bartoldson2025tba,yuan2025tbrm,wang2026reval}. AGRO and SPO support both on- and off-policy data \citep{tang2025agro,cohen2025spo}; OAPL addresses trainer--inference lag with asynchronous generation and resets its buffer at policy synchronization \citep{ritter2026oapl}. RePO and BAPO also reuse off-policy samples \citep{li2025repo,wan2026bapo}. CBPO retains positive and negative demonstrations, reference responses, and fresh and historical rollouts in one persistent buffer under its frozen target.

%% file: sections/3_preliminaries.tex
\begin{figure*}[t]
    \centering
    \includegraphics[width=\linewidth]{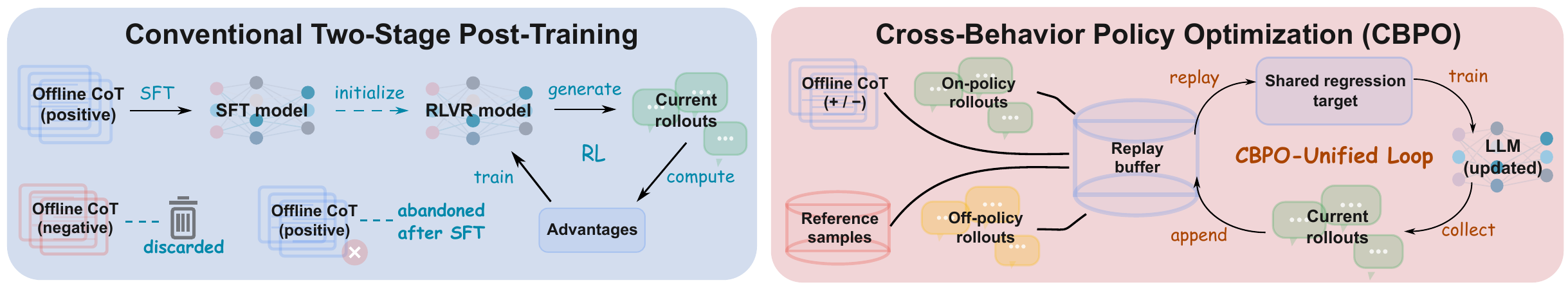}
    \vspace{-25pt}
    \caption{SFT$\rightarrow$RLVR vs.\ shared-target CBPO-Unified on offline and on-/off-policy data.}
    \vspace{-21pt}
    \label{fig:off_policy_training_loop}
\end{figure*}

\section{Preliminaries}
\label{sec:preliminaries}

We first distinguish the three data regimes considered in this work and review the conventional two-stage reasoning post-training pipeline.
Let $x$ denote a prompt, $y=(y_1,\ldots,y_T)$ a completion, and $r(x,y)\in\mathbb R$ the scalar reward assigned by a verifier.
For an autoregressive policy, we use
$\log \pi(y\!\mid\!x)=\sum_{j=1}^{T}\log \pi(y_j\!\mid\!x,y_{<j})$
for the sequence log-probability.
At optimization round $t$, $\pi_{\theta_t}$ denotes the current policy. A rollout collected at round $k$ is sampled from $\pi_{\theta_k}$. It is on-policy when used at round $k=t$ and off-policy when reused at a later round $t>k$.

\subsection{Three RL Data Regimes}
\label{sec:data_regimes}

We distinguish three data regimes by when trajectories are collected and if they remain eligible for subsequent updates.
\textbf{Offline Demonstrations} constitute the dataset $\mathcal D_{\mathrm{off}}=\{(x_i,y_i,r(x_i,y_i))\}_{i=1}^{N}$, which is available before policy optimization. Its completions may come from external policies whose identities need not be known.
\textbf{On-Policy Rollouts} are the fresh batch $\mathcal D_t^{\mathrm{roll}}$ sampled from $\pi_{\theta_t}(\cdot\!\mid\!x)$ at round $t$.
\textbf{Off-Policy Rollouts} at round $t$ were sampled from $\pi_{\theta_k}$ with $k<t$.
A replay buffer combines them with current rollouts as $\mathcal B_t^{\mathrm{roll}}=\bigcup_{k=0}^{t}\mathcal D_k^{\mathrm{roll}}$.

\subsection{Conventional Two-Stage Reasoning Post-Training}
\label{sec:two_stage_post_training}

Reasoning post-training commonly follows a two-stage pipeline in which cold-start SFT on positive offline chain-of-thought demonstrations precedes RLVR on model-generated rollouts \citep{shao2024deepseekmath,deepseekai2025r1}.
\begingroup
\setlength{\abovedisplayskip}{-4pt}
\setlength{\belowdisplayskip}{-2pt}
\setlength{\abovedisplayshortskip}{-4pt}
\setlength{\belowdisplayshortskip}{-2pt}
\begin{equation*}
\theta_{\mathrm{base}}
\xrightarrow[\text{cold-start SFT}]
{\mathcal D_{\mathrm{off}}^{+}}
\theta_{\mathrm{SFT}}
\xrightarrow[\text{reinforcement learning}]
{\mathcal D_t^{\mathrm{roll}}}
\theta_{\mathrm{RL}}.
\end{equation*}
\endgroup

The SFT stage minimizes the negative log-likelihood of demonstrations selected as correct or high quality.
\begingroup
\setlength{\abovedisplayskip}{-2pt}
\setlength{\belowdisplayskip}{-4pt}
\setlength{\abovedisplayshortskip}{-2pt}
\setlength{\belowdisplayshortskip}{-4pt}
\begin{equation*}
\mathcal L_{\mathrm{SFT}}(\theta)
=
-\mathbb E_{(x,y)\sim\mathcal D_{\mathrm{off}}^{+}}
\bigl[\log\pi_\theta(y\!\mid\!x)\bigr].
\end{equation*}
\endgroup

Because this objective increases the likelihood of every target completion regardless of reward, incorrect or low-reward demonstrations are typically filtered rather than used as ordinary SFT targets. The RLVR stage then initializes from $\theta_{\mathrm{SFT}}$, switches to an RL objective, and trains on newly generated on-policy rollouts. The offline demonstrations are typically not reused.

%% file: sections/4_methods.tex
\section{Cross-Behavior Policy Optimization}
\label{sec:methods}

We first derive the fixed-reference regression target and establish its per-completion invariance to the generating policy.
We then describe unified replay-buffer training, define data-restricted variants, and analyze the finite-sample bias of the frozen normalizer.

\subsection{Fixed-Reference Regression Target}
\label{sec:fixed_reference_target}
\begingroup
\setlength{\abovedisplayskip}{5pt plus 1pt minus 2pt}
\setlength{\belowdisplayskip}{5pt plus 1pt minus 2pt}
\setlength{\abovedisplayshortskip}{0pt plus 1pt}
\setlength{\belowdisplayshortskip}{3pt plus 1pt minus 1pt}

Let $\pi_{\mathrm{ref}}$ be a fixed reference policy and $\beta>0$ a reward-temperature coefficient.
For a prompt $x$, consider the KL-regularized objective
\begin{equation*}
\max_{\pi(\cdot\mid x)}
\left\{
\mathbb E_{y\sim\pi(\cdot\mid x)}[r(x,y)]
-\beta\,
D_{\mathrm{KL}}\!\left(
\pi(\cdot\!\mid\!x)\,\|\,\pi_{\mathrm{ref}}(\cdot\!\mid\!x)
\right)
\right\}.
\end{equation*}
Its closed-form optimizer is the reward-tilted reference distribution \citep{rafailov2023dpo}
\begin{equation}
\label{eq:cbpo_opt_policy}
\pi^\star(y\!\mid\!x)
=
\frac{\pi_{\mathrm{ref}}(y\!\mid\!x)
\exp\!\left(r(x,y)/\beta\right)}{Z_{\mathrm{ref}}(x)},
\qquad
Z_{\mathrm{ref}}(x)
=
\mathbb E_{y\sim\pi_{\mathrm{ref}}(\cdot\mid x)}
\!\left[\exp\!\left(r(x,y)/\beta\right)\right].
\end{equation}
To avoid the additional log-normalizer bias that top-$p$ truncation can introduce (Appendix~\ref{app:reference_decoding_bias}), we use temperature $1.0$ and top-$p=1$ for reference sampling.
We estimate $Z_{\mathrm{ref}}(x)$ from $K_{\mathrm{ref}}$ reference samples,
\begin{equation}
\label{eq:cbpo_z_hat}
\widehat Z_{\mathrm{ref}}(x)
=
\frac{1}{K_{\mathrm{ref}}}
\sum_{i=1}^{K_{\mathrm{ref}}}
\left[
\exp\!\left(\frac{r(x,y_i)}{\beta}\right)
\right]_{y_i\sim\pi_{\mathrm{ref}}(\cdot\mid x)},
\end{equation}
and freeze it before optimization.
Taking logs in Eq.~\eqref{eq:cbpo_opt_policy} and substituting the frozen normalizer estimate gives the target
\begin{equation}
\label{eq:cbpo_target}
\widehat t(x,y)
=
\log\pi_{\mathrm{ref}}(y\!\mid\!x)
+\frac{r(x,y)}{\beta}
-\log\widehat Z_{\mathrm{ref}}(x).
\end{equation}
CBPO fits the completion log-probability to the target with the following response-level loss.
\begin{equation}
\label{eq:cbpo_loss}
\ell_{\mathrm{CBPO}}(x,y;\theta)
:=
\left[
\log\frac{\pi_\theta(y\!\mid\!x)}{\pi_{\mathrm{ref}}(y\!\mid\!x)}
-\frac{r(x,y)}{\beta}
+\log\widehat Z_{\mathrm{ref}}(x)
\right]^2
\end{equation}
Each completion receives one target constructed from its sequence-level log-probability.
\endgroup

\paragraph{Behavior-policy independence.}
For any completion $y$ sampled from a behavior policy $\mu(\cdot\mid x)$, the target in Eq.~\eqref{eq:cbpo_target} depends only on its verifier reward, reference log-probability, and frozen prompt normalizer, and is therefore independent of $\mu$. Consequently, completions collected by different policies can be optimized with the same per-sample regression loss. This independence applies to the per-record target and loss, not to the expected training objective. The behavior policy determines which completions are observed, while the buffer-sampling distribution determines their weights in the aggregate objective.

\subsection{CBPO-Unified Across Three Data Regimes}
\label{sec:cbpo_unified}

CBPO stores each completion as a record of $(x,y)$, its verifier reward, and reference log-probability, with one frozen normalizer cached per prompt.

\paragraph{Replay initialization.}
We rescore offline demonstrations with the rollout verifier, yielding positive and negative subsets ($\mathcal D_{\mathrm{off}}=\mathcal D_{\mathrm{off}}^{+}\cup\mathcal D_{\mathrm{off}}^{-}$) with reference log-probabilities.
The $K_{\mathrm{ref}}$ reference responses per prompt used in Eq.~\eqref{eq:cbpo_z_hat} form $\mathcal D_{\mathrm{ref}}$ and also enter training.
Before optimization, we freeze $\log\widehat Z_{\mathrm{ref}}(x)$ and set $\mathcal B_0=\mathcal D_{\mathrm{off}}\cup\mathcal D_{\mathrm{ref}}$.

\paragraph{Interleaved replay and collection.}
\label{sec:pr_schedule}
At round $t$, $\mathcal B_t=\mathcal B_0\cup\mathcal B_t^{\mathrm{roll}}$ contains $N_t$ indexed records $(x_i,y_i)$ with uniform replay weights $q_t(i)=1/N_t$.
CBPO minimizes
\begingroup
\setlength{\abovedisplayskip}{0pt}
\setlength{\belowdisplayskip}{0pt}
\setlength{\abovedisplayshortskip}{0pt}
\setlength{\belowdisplayshortskip}{0pt}
\begin{equation*}
\mathcal L_t(\theta)
=
\mathbb E_{i\sim q_t}
\left[\ell_{\mathrm{CBPO}}(x_i,y_i;\theta)\right]
=\frac{1}{N_t}\sum_{i=1}^{N_t}
\ell_{\mathrm{CBPO}}(x_i,y_i;\theta).
\end{equation*}
\endgroup
After replay, $\pi_{\theta_{t+1}}$ generates $\mathcal D_{t+1}^{\mathrm{roll}}$.
The verifier scores each completion, CBPO attaches its reference log-probability and cached normalizer, and $\mathcal B_{t+1}^{\mathrm{roll}}=\mathcal B_t^{\mathrm{roll}}\cup\mathcal D_{t+1}^{\mathrm{roll}}$.
Fresh batches are on-policy, retained batches become off-policy, and offline demonstrations remain eligible throughout.
Fig.~\ref{fig:off_policy_training_loop} illustrates this process, and Algorithm~\ref{alg:cbpo} specifies the schedule.

\begingroup
\makeatletter
\let\cbpo@original@fs@ruled\fs@ruled
\def\fs@ruled{%
  \cbpo@original@fs@ruled
  \def\@fs@post{\kern2pt\hrule\relax\vspace{-20pt}}%
}
\makeatother
\begin{algorithm}[t]
\caption{CBPO-Unified Training Schedule}
\label{alg:cbpo}
\begin{algorithmic}[1]
\Require Prompt source $\mathcal X$, offline data $\mathcal D_{\mathrm{off}}$, reference policy $\pi_{\mathrm{ref}}$, verifier $v$, prompt batch size $P$, responses per prompt $G$, replay minibatch size $Q$, replay updates per trainable iteration $M$, reference samples per prompt $K_{\mathrm{ref}}$
\Statex \textbf{Reference preprocessing}
\State Initialize $\pi_\theta\leftarrow\pi_{\mathrm{ref}}$
\State Score $\mathcal D_{\mathrm{off}}$ with $v$ and attach reference log-probabilities
\State Collect $\mathcal D_{\mathrm{ref}}$ with $K_{\mathrm{ref}}$ responses per prompt from $\pi_{\mathrm{ref}}$
\State Score $\mathcal D_{\mathrm{ref}}$, attach reference log-probabilities, and initialize $\mathcal B\leftarrow\mathcal D_{\mathrm{off}}\cup\mathcal D_{\mathrm{ref}}$
\State Freeze $\log\widehat Z_{\mathrm{ref}}(x)$ from the $K_{\mathrm{ref}}$ reference responses per prompt
\Statex \textbf{Unified optimization loop}
\For{epochs after preprocessing}
  \State Reset the within-epoch prompt-exclusion set
  \For{prompt batches $b=1,2,\ldots$}
    \For{$s=1$ to $M$}
      \State Draw $Q$ eligible response records from $\mathcal B$ and update $\theta$ using Eq.~\eqref{eq:cbpo_loss}
    \EndFor
    \State Sample $P$ unused prompts and collect $G$ responses per prompt from $\pi_\theta$
    \State Score them with $v$, attach reference log-probabilities, and append them to $\mathcal B$
  \EndFor
\EndFor
\end{algorithmic}
\end{algorithm}
\endgroup

\subsection{Special Forms of CBPO}
\label{sec:special_forms}
Restricting the trainable data sources yields three controlled variants under the same target.
\textbf{CBPO-Offline} fixes $\mathcal D_{\mathrm{off}}\cup\mathcal D_{\mathrm{ref}}$ and performs no rollout collection.
\textbf{CBPO-OnPolicy} uses reference responses to construct the target, trains on the latest current-policy rollouts, and discards earlier rollout rounds.
\textbf{CBPO-OffPolicy} omits offline demonstrations, initializes replay with $\mathcal D_{\mathrm{ref}}$, and retains both current and historical rollouts.
\textbf{CBPO-Unified} combines Offline Demonstrations and On- and Off-Policy Rollouts in the evolving buffer.

\subsection{Finite-Sample Normalizer Analysis}
\label{sec:theoretical_analysis}
For a fixed record, the verifier reward and reference log-probability are directly evaluated. The Monte Carlo normalizer is the only sampled term in the target.
Although Eq.~\eqref{eq:cbpo_z_hat} is unbiased for $Z_{\mathrm{ref}}(x)$, Jensen's inequality gives $\mathbb E[\log\widehat Z_{\mathrm{ref}}(x)]\le\log Z_{\mathrm{ref}}(x)$.
Thus, the frozen log-normalizer is downward biased at finite $K_{\mathrm{ref}}$.
Let
$\sigma_x^2=\operatorname{Var}_{y\sim\pi_{\mathrm{ref}}}
[\exp(r(x,y)/\beta)]$
and assume $\exp(r(x,y)/\beta)\ge a_x>0$.
A second-order expansion yields
\begingroup
\setlength{\abovedisplayskip}{0pt}
\setlength{\belowdisplayskip}{4pt}
\setlength{\abovedisplayshortskip}{0pt}
\setlength{\belowdisplayshortskip}{4pt}
\begin{equation}
\label{eq:logz_bias_bound_main}
0
\le
\log Z_{\mathrm{ref}}(x)
-\mathbb E[\log\widehat Z_{\mathrm{ref}}(x)]
\le
\frac{\sigma_x^2}{2K_{\mathrm{ref}}a_x^2}.
\end{equation}
\endgroup
Our rewards lie in $\{0,0.1,1.0\}$, so $a_x=1$.
At fixed finite model parameters and finite activations, bounded response length gives finite score norms, and the induced gradient bias is $O(1/K_{\mathrm{ref}})$.
If reference samples used in the estimate are replayed for training, an additional covariance term appears. It remains $O(1/K_{\mathrm{ref}})$ under bounded score gradients, while a leave-one-out estimate removes this dependence.
Appendix~\ref{app:mc-logz-bias} proves these bounds and relates regime-specific regression error to the KL-regularized objective gap through normalizer, sampling, approximation, and optimization errors.
Hence $K_{\mathrm{ref}}$ trades normalizer accuracy against reference-sampling cost and, through the length of reference preprocessing, against the number of current-policy rollouts.

%% file: sections/5_experiments.tex
\section{Experiments}
\label{sec:experiments}

\subsection{Experimental Setup}
\label{sec:experimental_setup}

\paragraph{Training.}
We use Qwen3-1.7B as the base and reference model and Qwen3-8B for the scale study.
Training uses 16,254 decontaminated RLVR-Math-16k prompts and, per prompt, two positive and one negative DPSK-CoT trajectories from DeepSeek-R1-Distill-Qwen-32B \citep{deepseekai2025r1} (Appendix~\ref{app:decontamination}).
Rewards are $1.0$, $0.1$, and $0$ for correct, parseable but incorrect, and unparseable responses, respectively.
Core Qwen3-1.7B training uses 650,160 generated responses, 1,280 optimizer steps, and $K_{\mathrm{ref}}=24$ reference samples per prompt (Appendix~\ref{app:training_system_settings}).

\paragraph{Baselines.}
We compare CBPO-Offline with SFT and offline DPO \citep{rafailov2023dpo}, CBPO-OnPolicy with PPO \citep{schulman2017ppo}, DAPO \citep{yu2025dapo}, REINFORCE++ \citep{hu2025reinforceplusplus}, and GSPO \citep{zheng2025gspo}, and CBPO-OffPolicy with TBRM \citep{yuan2025tbrm}.
CBPO-Unified is compared with BRIDGE \citep{chen2026bridge} at both model sizes.
Inspired by~\citet{liu2026paying}, we adopt a compromise SFT:DAPO step ratio of 1:2.6.
Comparisons match models, prompts, positive-trajectory counts, actor learning rates, generation and optimizer-step budgets, decoding, verifier, and evaluation protocol (Appendix~\ref{app:baseline_details}).

\paragraph{Evaluation.}
We evaluate 842 prompts from AIME25, MinervaMath, AMC23, and MATH500, with 32 responses each.
\texttt{avg@32} is mean response correctness, and \texttt{pass@32} is the percentage of prompts with at least one correct response.
For comparisons across five seeds, aggregate scores average the four benchmarks equally within each seed before computing the mean and sample standard deviation.
Bold and underlined scores mark the best overall and within-regime scores, respectively.

\begin{table}[t]
  \centering
  \scriptsize
  \setlength{\tabcolsep}{0.25pt}
  \renewcommand{\arraystretch}{1.0}
  \vspace{-5pt}
  \caption{Qwen3-1.7B performance across data regimes and joint SFT/RL over five training seeds.}
  \label{tab:core_cross_regime}
  \begin{tabularx}{\linewidth}{@{}>{\fontsize{6}{7}\selectfont}l>{\fontsize{6}{7}\selectfont}l *{10}{>{\centering\arraybackslash}X}@{}}
    \toprule
    \multirow[c]{2}{*}{\raisebox{-2pt}{\textbf{Regime}}} & \multirow[c]{2}{*}{\raisebox{-2pt}{\textbf{Methods}}}
      & \multicolumn{2}{c}{\textbf{AIME25}}
      & \multicolumn{2}{c}{\textbf{MinervaMath}}
      & \multicolumn{2}{c}{\textbf{AMC23}}
      & \multicolumn{2}{c}{\textbf{MATH500}}
      & \multicolumn{2}{c}{\textbf{Avg.}} \\
    \cmidrule(lr){3-4}\cmidrule(lr){5-6}\cmidrule(lr){7-8}\cmidrule(lr){9-10}\cmidrule(lr){11-12}
      & & \metrichead{avg@32} & \metrichead{pass@32} & \metrichead{avg@32} & \metrichead{pass@32} & \metrichead{avg@32} & \metrichead{pass@32} & \metrichead{avg@32} & \metrichead{pass@32} & \metrichead{avg@32} & \metrichead{pass@32} \\
    \midrule
    \multirow{3}{*}{Offline} & SFT & \seedscore{14.23}{0.06} & \seedscore{\underline{29.33}}{1.49} & \seedscore{19.29}{0.02} & \seedscore{33.53}{0.16} & \seedscore{48.77}{0.03} & \seedscore{75.50}{1.12} & \seedscore{64.99}{0.01} & \seedscore{77.96}{0.09} & \seedscore{36.82}{0.02} & \seedscore{54.08}{0.11} \\
     & DPO & \seedscore{14.50}{0.05} & \seedscore{\underline{29.33}}{1.49} & \seedscore{19.32}{0.01} & \seedscore{33.68}{0.20} & \seedscore{48.97}{0.03} & \seedscore{74.50}{1.12} & \seedscore{65.34}{0.01} & \seedscore{78.16}{0.09} & \seedscore{37.03}{0.02} & \seedscore{53.92}{0.41} \\
     & \textbf{CBPO-Offline} & \seedscore{\underline{15.81}}{0.05} & \seedscore{\underline{29.33}}{1.49} & \seedscore{\underline{20.29}}{0.01} & \seedscore{\underline{35.22}}{0.16} & \seedscore{\underline{50.11}}{0.04} & \seedscore{\underline{78.00}}{1.12} & \seedscore{\underline{66.99}}{0.01} & \seedscore{\underline{80.44}}{0.09} & \seedscore{\underline{38.30}}{0.02} & \seedscore{\underline{55.75}}{0.54} \\
    \midrule
    \multirow{5}{*}{On-policy} & PPO & \seedscore{\underline{17.48}}{0.05} & \seedscore{32.67}{1.49} & \seedscore{21.35}{0.01} & \seedscore{38.82}{0.20} & \seedscore{56.47}{0.03} & \seedscore{84.50}{1.12} & \seedscore{69.71}{0.01} & \seedscore{83.68}{0.11} & \seedscore{41.25}{0.02} & \seedscore{59.92}{0.42} \\
     & DAPO & \seedscore{17.27}{0.05} & \seedscore{38.00}{1.83} & \seedscore{21.58}{0.01} & \seedscore{\underline{39.78}}{0.16} & \seedscore{56.23}{0.03} & \seedscore{85.50}{1.12} & \seedscore{70.38}{0.01} & \seedscore{84.64}{0.09} & \seedscore{41.37}{0.01} & \seedscore{61.98}{0.67} \\
     & GSPO & \seedscore{17.21}{0.05} & \seedscore{37.33}{1.49} & \seedscore{21.86}{0.01} & \seedscore{\underline{39.78}}{0.16} & \seedscore{55.64}{0.03} & \seedscore{81.50}{1.37} & \seedscore{70.51}{0.01} & \seedscore{84.64}{0.09} & \seedscore{41.30}{0.02} & \seedscore{60.81}{0.33} \\
     & REINFORCE++ & \seedscore{16.90}{0.05} & \seedscore{32.67}{1.49} & \seedscore{21.19}{0.01} & \seedscore{38.31}{0.16} & \seedscore{52.59}{0.03} & \seedscore{79.50}{1.12} & \seedscore{68.58}{0.02} & \seedscore{81.84}{0.09} & \seedscore{39.81}{0.02} & \seedscore{58.08}{0.41} \\
     & \textbf{CBPO-OnPolicy} & \seedscore{17.46}{0.06} & \seedscore{\underline{39.33}}{1.49} & \seedscore{\underline{22.02}}{0.02} & \seedscore{\underline{39.78}}{0.16} & \seedscore{\underline{57.20}}{0.03} & \seedscore{\underline{88.00}}{1.12} & \seedscore{\underline{70.69}}{0.01} & \seedscore{\underline{84.96}}{0.09} & \seedscore{\underline{41.84}}{0.01} & \seedscore{\underline{63.02}}{0.52} \\
    \midrule
    \multirow{2}{*}{Off-policy} & TBRM & \seedscore{18.15}{0.05} & \seedscore{\underline{42.00}}{1.83} & \seedscore{22.10}{0.01} & \seedscore{\underline{40.37}}{0.16} & \seedscore{57.53}{0.04} & \seedscore{90.50}{1.12} & \seedscore{\underline{71.46}}{0.01} & \seedscore{83.56}{0.09} & \seedscore{42.31}{0.02} & \seedscore{64.11}{0.64} \\
     & \textbf{CBPO-OffPolicy} & \seedscore{\underline{18.67}}{0.05} & \seedscore{41.33}{1.83} & \seedscore{\underline{22.64}}{0.01} & \seedscore{40.29}{0.20} & \seedscore{\underline{57.98}}{0.03} & \seedscore{\underline{91.00}}{1.37} & \seedscore{\underline{71.46}}{0.01} & \seedscore{\underline{85.84}}{0.09} & \seedscore{\underline{42.69}}{0.02} & \seedscore{\underline{64.62}}{0.38} \\
    \midrule
    Joint SFT/RL\hspace{3pt} & BRIDGE & \seedscore{18.98}{0.05} & \seedscore{44.00}{1.49} & \seedscore{23.43}{0.01} & \seedscore{41.47}{0.16} & \seedscore{57.98}{0.03} & \seedscore{87.00}{1.12} & \seedscore{71.99}{0.02} & \seedscore{85.76}{0.09} & \seedscore{43.10}{0.01} & \seedscore{64.56}{0.47} \\
    \midrule
    Unified & \textbf{CBPO-Unified} & \seedscore{\textbf{20.67}}{0.06} & \seedscore{\textbf{49.33}}{1.49} & \seedscore{\textbf{24.47}}{0.01} & \seedscore{\textbf{42.35}}{0.16} & \seedscore{\textbf{60.17}}{0.03} & \seedscore{\textbf{94.50}}{1.12} & \seedscore{\textbf{72.99}}{0.01} & \seedscore{\textbf{87.64}}{0.09} & \seedscore{\textbf{44.57}}{0.02} & \seedscore{\textbf{68.46}}{0.59} \\
    \bottomrule
  \end{tabularx}
  \vspace{-15pt}
\end{table}

\begin{table}[t]
  \centering
  \scriptsize
  \setlength{\tabcolsep}{0.25pt}
  \renewcommand{\arraystretch}{1.0}
  \vspace{-5pt}
  \caption{Qwen3-8B performance across data regimes and joint SFT/RL over five training seeds.}
  \label{tab:core_cross_regime_8b}
  \begin{tabularx}{\linewidth}{@{}>{\fontsize{6}{7}\selectfont}l>{\fontsize{6}{7}\selectfont}l *{10}{>{\centering\arraybackslash}X}@{}}
    \toprule
    \multirow[c]{2}{*}{\raisebox{-2pt}{\textbf{Regime}}} & \multirow[c]{2}{*}{\raisebox{-2pt}{\textbf{Methods}}}
      & \multicolumn{2}{c}{\textbf{AIME25}}
      & \multicolumn{2}{c}{\textbf{MinervaMath}}
      & \multicolumn{2}{c}{\textbf{AMC23}}
      & \multicolumn{2}{c}{\textbf{MATH500}}
      & \multicolumn{2}{c}{\textbf{Avg.}} \\
    \cmidrule(lr){3-4}\cmidrule(lr){5-6}\cmidrule(lr){7-8}\cmidrule(lr){9-10}\cmidrule(lr){11-12}
      & & \metrichead{avg@32} & \metrichead{pass@32} & \metrichead{avg@32} & \metrichead{pass@32} & \metrichead{avg@32} & \metrichead{pass@32} & \metrichead{avg@32} & \metrichead{pass@32} & \metrichead{avg@32} & \metrichead{pass@32} \\
    \midrule
    \multirow{3}{*}{Offline} & SFT & \seedscore{33.17}{0.06} & \seedscore{\underline{38.00}}{1.83} & \seedscore{45.20}{0.01} & \seedscore{52.87}{0.16} & \seedscore{83.16}{0.04} & \seedscore{87.00}{1.12} & \seedscore{84.61}{0.02} & \seedscore{88.56}{0.09} & \seedscore{61.53}{0.02} & \seedscore{66.61}{0.65} \\
     & DPO & \seedscore{33.52}{0.05} & \seedscore{37.33}{1.49} & \seedscore{45.60}{0.01} & \seedscore{53.38}{0.16} & \seedscore{83.27}{0.03} & \seedscore{\underline{90.50}}{1.12} & \seedscore{85.11}{0.01} & \seedscore{89.76}{0.09} & \seedscore{61.87}{0.02} & \seedscore{67.74}{0.38} \\
     & \textbf{CBPO-Offline} & \seedscore{\underline{34.04}}{0.05} & \seedscore{\underline{38.00}}{1.83} & \seedscore{\underline{47.20}}{0.01} & \seedscore{\underline{55.29}}{0.20} & \seedscore{\underline{84.13}}{0.03} & \seedscore{89.50}{1.12} & \seedscore{\underline{85.61}}{0.02} & \seedscore{\underline{90.64}}{0.09} & \seedscore{\underline{62.74}}{0.02} & \seedscore{\underline{68.36}}{0.39} \\
    \midrule
    \multirow{5}{*}{On-policy} & PPO & \seedscore{41.27}{0.05} & \seedscore{\underline{47.33}}{1.49} & \seedscore{50.31}{0.01} & \seedscore{58.68}{0.20} & \seedscore{89.72}{0.04} & \seedscore{95.50}{1.12} & \seedscore{89.40}{0.01} & \seedscore{96.44}{0.09} & \seedscore{67.67}{0.01} & \seedscore{74.49}{0.43} \\
     & DAPO & \seedscore{39.40}{0.05} & \seedscore{42.67}{1.49} & \seedscore{49.08}{0.01} & \seedscore{58.16}{0.16} & \seedscore{88.53}{0.03} & \seedscore{97.00}{1.12} & \seedscore{89.60}{0.01} & \seedscore{96.76}{0.09} & \seedscore{66.65}{0.01} & \seedscore{73.65}{0.66} \\
     & GSPO & \seedscore{39.04}{0.05} & \seedscore{42.00}{1.83} & \seedscore{49.12}{0.01} & \seedscore{58.24}{0.20} & \seedscore{87.13}{0.03} & \seedscore{97.00}{1.12} & \seedscore{89.16}{0.01} & \seedscore{96.56}{0.09} & \seedscore{66.11}{0.02} & \seedscore{73.45}{0.61} \\
     & REINFORCE++ & \seedscore{35.96}{0.05} & \seedscore{40.67}{1.49} & \seedscore{48.76}{0.01} & \seedscore{57.28}{0.16} & \seedscore{83.53}{0.03} & \seedscore{95.50}{1.12} & \seedscore{88.93}{0.02} & \seedscore{95.04}{0.09} & \seedscore{64.29}{0.02} & \seedscore{72.12}{0.42} \\
     & \textbf{CBPO-OnPolicy} & \seedscore{\underline{42.27}}{0.05} & \seedscore{\underline{47.33}}{1.49} & \seedscore{\underline{50.95}}{0.01} & \seedscore{\underline{59.12}}{0.16} & \seedscore{\underline{90.69}}{0.03} & \seedscore{\underline{98.00}}{1.12} & \seedscore{\underline{89.90}}{0.01} & \seedscore{\underline{96.96}}{0.09} & \seedscore{\underline{68.45}}{0.01} & \seedscore{\underline{75.35}}{0.62} \\
    \midrule
    \multirow{2}{*}{Off-policy} & TBRM & \seedscore{43.25}{0.05} & \seedscore{\underline{47.33}}{1.49} & \seedscore{\underline{52.14}}{0.01} & \seedscore{\underline{60.22}}{0.16} & \seedscore{91.34}{0.03} & \seedscore{\underline{98.00}}{1.12} & \seedscore{\underline{90.19}}{0.01} & \seedscore{97.24}{0.09} & \seedscore{69.23}{0.02} & \seedscore{\underline{75.70}}{0.39} \\
     & \textbf{CBPO-OffPolicy} & \seedscore{\underline{43.52}}{0.05} & \seedscore{\underline{47.33}}{1.49} & \seedscore{51.65}{0.01} & \seedscore{60.00}{0.16} & \seedscore{\underline{91.73}}{0.03} & \seedscore{96.50}{1.37} & \seedscore{\underline{90.19}}{0.02} & \seedscore{\underline{97.36}}{0.09} & \seedscore{\underline{69.27}}{0.02} & \seedscore{75.30}{0.58} \\
    \midrule
    Joint SFT/RL\hspace{3pt} & BRIDGE & \seedscore{44.25}{0.05} & \seedscore{48.00}{1.83} & \seedscore{53.78}{0.01} & \seedscore{63.38}{0.20} & \seedscore{92.22}{0.04} & \seedscore{98.00}{1.12} & \seedscore{91.65}{0.01} & \seedscore{97.24}{0.09} & \seedscore{70.47}{0.02} & \seedscore{76.66}{0.40} \\
    \midrule
    Unified & \textbf{CBPO-Unified} & \seedscore{\textbf{46.52}}{0.06} & \seedscore{\textbf{52.67}}{1.49} & \seedscore{\textbf{56.65}}{0.01} & \seedscore{\textbf{66.84}}{0.16} & \seedscore{\textbf{93.17}}{0.04} & \seedscore{\textbf{99.50}}{1.12} & \seedscore{\textbf{94.89}}{0.02} & \seedscore{\textbf{97.88}}{0.11} & \seedscore{\textbf{72.81}}{0.02} & \seedscore{\textbf{79.22}}{0.41} \\
    \bottomrule
  \end{tabularx}
  \vspace{-15pt}
\end{table}

\subsection{Performance across Data Regimes}
\label{sec:core_cross_regime}

\paragraph{Restricted data regimes.}
Tables~\ref{tab:core_cross_regime} and~\ref{tab:core_cross_regime_8b} compare methods initialized from Qwen3-1.7B and Qwen3-8B, respectively.
With Qwen3-1.7B, CBPO-Offline improves SFT by 1.48/1.67 aggregate \texttt{avg@32/pass@32} points.
Offline DPO provides a preference-based comparison using both correct and incorrect records. CBPO-Offline exceeds DPO by 1.27/1.83 aggregate \texttt{avg@32/pass@32} points with Qwen3-1.7B and by 0.87/0.62 points with Qwen3-8B.
CBPO-OnPolicy reaches 41.84/63.02, exceeding the strongest on-policy baseline by 0.47/1.04 points.
CBPO-OffPolicy improves TBRM by 0.38 points on aggregate \texttt{avg@32} and improves or matches it on six of the eight benchmark-metric entries. Its aggregate \texttt{pass@32} is 0.51 points higher, with lower scores on AIME25 and MinervaMath and higher scores on AMC23 and MATH500.
With Qwen3-8B, the three restricted variants improve the strongest regime-matched baselines on aggregate \texttt{avg@32} by 0.87, 0.78, and 0.04 points, respectively.
These comparisons show that the shared regression target is effective when restricted to any one of the three data regimes.

\paragraph{Unified learning.}
CBPO-Unified achieves the best aggregate \texttt{avg@32/pass@32}, reaching 44.57/68.46 with Qwen3-1.7B and 72.81/79.22 with Qwen3-8B.
At the two scales, it exceeds BRIDGE by 1.47/3.90 and 2.34/2.56 points, and the strongest single-regime results by 1.88/3.84 and 3.54/3.52 points, respectively.
The \texttt{avg@32} gains over the strongest single-regime results range from 1.53 to 2.19 points across the 1.7B benchmarks. At 8B, CBPO-Unified leads all eight benchmark-metric entries.
Appendix~\ref{app:heterogeneous_sources} extends the BRIDGE comparison to three teacher sources and their mixture.

\paragraph{Learning dynamics and cost.}
Fig.~\ref{fig:core_performance_comparison} shows that CBPO-Unified improves rapidly early in training and leads both metrics on all four benchmarks at the final checkpoint.
The advantage extends beyond the final checkpoint. CBPO-Unified continues improving after the rapid early gains shared by the rollout methods, while its competitors plateau lower or fluctuate around weaker solutions.
This behavior is consistent across both accuracy and coverage metrics and is especially pronounced on AIME25 and AMC23.
With Qwen3-1.7B, CBPO-Unified uses 119.36 GPU hours, within $2\%$ of DAPO, REINFORCE++, GSPO, and TBRM.
PPO requires 232.84 GPU hours because it additionally trains a critic.
The 8B results show the same cost pattern. CBPO-Unified requires 437.44 GPU hours, below the non-PPO on-policy baselines and less than half the cost of PPO.

\begingroup
\setlength{\intextsep}{16pt}
\begin{figure}[t]
  \centering
  \includegraphics[width=\textwidth]{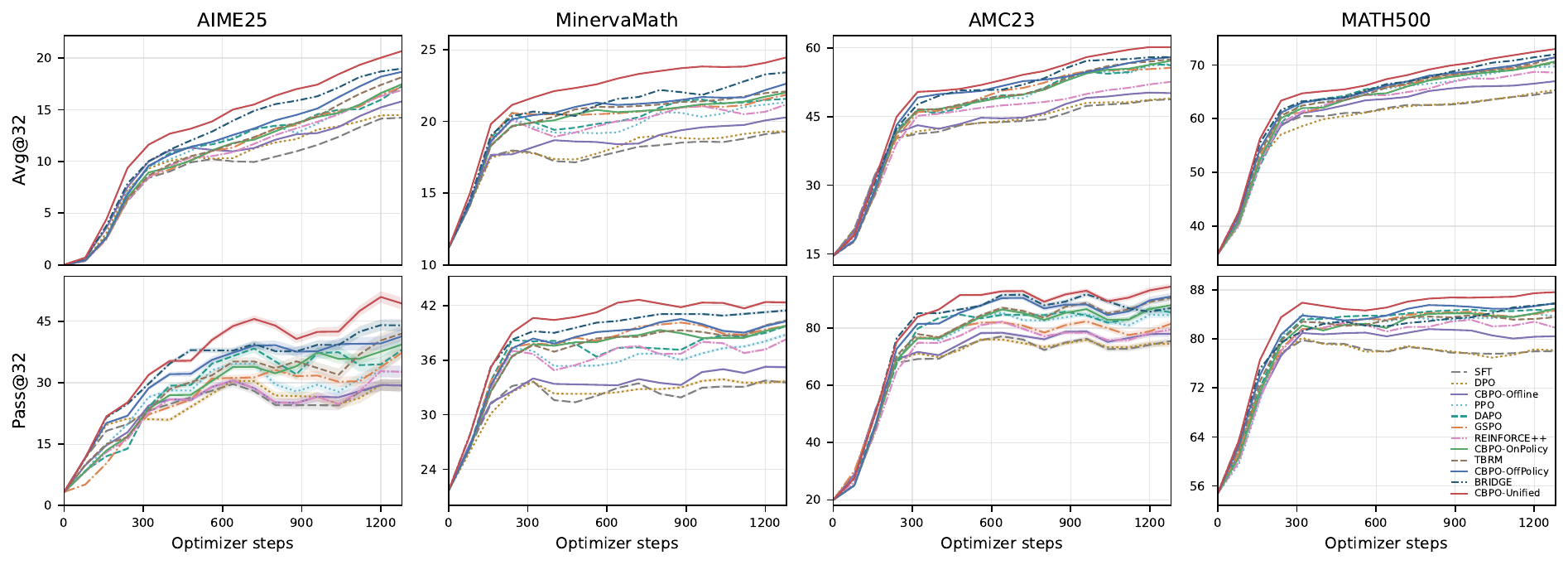}
  \vspace{-23pt}
  \caption{Qwen3-1.7B \texttt{avg@32}/\texttt{pass@32} trajectories for twelve methods on four benchmarks.}
  \label{fig:core_performance_comparison}
  \vspace{-18pt}
\end{figure}
\endgroup

\subsection{Shared-Target Optimization and Data Reuse}
\label{sec:one_stage_comparison}

\paragraph{End-to-end performance.}
Table~\ref{tab:one_vs_two_stage} compares SFT$\rightarrow$DAPO with CBPO under progressively expanded training-data configurations.
CBPO w/o History (PosCoT) applies the shared target to the same positive offline trajectories as SFT$\rightarrow$DAPO and to fresh rollouts under the same current-policy generation budget.
CBPO w/o History adds negative offline trajectories.
CBPO-Unified FIFO-1 additionally trains on reference responses and retains only the latest rollout round.

CBPO w/o History (PosCoT) scores 0.13 points higher on aggregate \texttt{avg@32} and 0.68 points lower on aggregate \texttt{pass@32} than SFT$\rightarrow$DAPO under matched positive offline trajectories and current-policy generation budgets.
CBPO w/o History exceeds CBPO w/o History (PosCoT) by 0.33/1.40 aggregate \texttt{avg@32/pass@32} points after adding negative offline trajectories, and CBPO-Unified FIFO-1 exceeds CBPO w/o History by 0.38/1.65 points after adding reference-response training.

CBPO-Unified FIFO-1 improves aggregate \texttt{avg@32/pass@32} by 0.84/2.37 points over SFT$\rightarrow$DAPO and raises \texttt{avg@32} on every benchmark.
Its \texttt{pass@32} improves on all four benchmarks, including a 0.64-point increase on MATH500.
Retaining earlier rollout rounds further improves aggregate performance by 1.49/2.07 points, so CBPO-Unified exceeds SFT$\rightarrow$DAPO by 2.33/4.44 points overall and leads the reported results on all eight benchmark-metric combinations.
Relative to the two-stage pipeline, the unified gains reach 2.36/6.66 on AIME25, 2.49/1.47 on MinervaMath, 2.89/7.50 on AMC23, and 1.61/2.12 on MATH500.

\begin{table}[H]
  \centering
  \scriptsize
  \setlength{\tabcolsep}{0.25pt}
  \renewcommand{\arraystretch}{1.0}
  \vspace{-5pt}
  \caption{Qwen3-1.7B comparison of SFT$\rightarrow$DAPO and CBPO variants over five training seeds.}
  \label{tab:one_vs_two_stage}
  \begin{tabularx}{\linewidth}{@{}>{\fontsize{6}{7}\selectfont}l *{10}{>{\centering\arraybackslash}X}@{}}
    \toprule
    \multirow[c]{2}{*}{\raisebox{-2pt}{\textbf{Methods}}}
      & \multicolumn{2}{c}{\textbf{AIME25}}
      & \multicolumn{2}{c}{\textbf{MinervaMath}}
      & \multicolumn{2}{c}{\textbf{AMC23}}
      & \multicolumn{2}{c}{\textbf{MATH500}}
      & \multicolumn{2}{c}{\textbf{Avg.}} \\
    \cmidrule(lr){2-3}\cmidrule(lr){4-5}\cmidrule(lr){6-7}\cmidrule(lr){8-9}\cmidrule(lr){10-11}
      & \metrichead{avg@32} & \metrichead{pass@32} & \metrichead{avg@32} & \metrichead{pass@32} & \metrichead{avg@32} & \metrichead{pass@32} & \metrichead{avg@32} & \metrichead{pass@32} & \metrichead{avg@32} & \metrichead{pass@32} \\
    \midrule
    SFT $\rightarrow$ DAPO & \seedscore{18.31}{0.05} & \seedscore{42.67}{1.49} & \seedscore{21.98}{0.01} & \seedscore{40.88}{0.16} & \seedscore{57.28}{0.03} & \seedscore{87.00}{1.12} & \seedscore{71.38}{0.02} & \seedscore{85.52}{0.11} & \seedscore{42.24}{0.02} & \seedscore{64.02}{0.36} \\
    CBPO w/o History (PosCoT) & \seedscore{18.17}{0.06} & \seedscore{42.67}{1.49} & \seedscore{22.33}{0.02} & \seedscore{40.29}{0.20} & \seedscore{57.33}{0.03} & \seedscore{84.50}{1.12} & \seedscore{71.67}{0.01} & \seedscore{85.88}{0.11} & \seedscore{42.37}{0.02} & \seedscore{63.34}{0.43} \\
    CBPO w/o History & \seedscore{18.77}{0.05} & \seedscore{42.67}{1.49} & \seedscore{22.52}{0.02} & \seedscore{41.25}{0.16} & \seedscore{57.98}{0.03} & \seedscore{89.00}{1.37} & \seedscore{71.54}{0.01} & \seedscore{86.04}{0.09} & \seedscore{42.70}{0.01} & \seedscore{64.74}{0.64} \\
    CBPO-Unified FIFO-1 & \seedscore{18.98}{0.05} & \seedscore{44.67}{1.83} & \seedscore{22.85}{0.01} & \seedscore{41.25}{0.16} & \seedscore{58.55}{0.04} & \seedscore{93.50}{1.37} & \seedscore{71.96}{0.01} & \seedscore{86.16}{0.09} & \seedscore{43.08}{0.02} & \seedscore{66.39}{0.61} \\
    \textbf{CBPO-Unified} & \seedscore{\textbf{20.67}}{0.06} & \seedscore{\textbf{49.33}}{1.49} & \seedscore{\textbf{24.47}}{0.01} & \seedscore{\textbf{42.35}}{0.16} & \seedscore{\textbf{60.17}}{0.03} & \seedscore{\textbf{94.50}}{1.12} & \seedscore{\textbf{72.99}}{0.01} & \seedscore{\textbf{87.64}}{0.09} & \seedscore{\textbf{44.57}}{0.02} & \seedscore{\textbf{68.46}}{0.59} \\
    \bottomrule
  \end{tabularx}
  \vspace{-14pt}
\end{table}

\paragraph{Replay sampling.}
Appendix~\ref{app:replay_sampling} specifies complete CBPO-Unified training runs with response-balanced, prompt-balanced, source-balanced, and age-balanced replay sampling.
The comparisons use the same initialization, training seed, generation budget, and optimizer-step budget.

\paragraph{Historical replay diagnostics.}
\label{sec:replay_horizon}

The replay-horizon comparison in Table~\ref{tab:replay_horizon} retains the same offline and reference data across the FIFO and full-replay variants.
For the finite history variants, the replay horizon is $K\in\{1,4,16\}$.
At round $t$, the retained rollout index set is $\mathcal I_t^{(K)}:=\{\max(0,t-K+1),\ldots,t\}$.
The corresponding training buffer is $\mathcal B_t^{(K)}:=\mathcal D_{\mathrm{off}}\cup\mathcal D_{\mathrm{ref}}\cup\bigl(\bigcup\nolimits_{j\in\mathcal I_t^{(K)}}\mathcal D_j^{\mathrm{roll}}\bigr)$.
Full replay uses all rollout rounds collected up to $t$.

\begin{table}[!htbp]
  \centering
  \scriptsize
  \setlength{\tabcolsep}{3pt}
  \vspace{-5pt}
  \caption{Qwen3-1.7B with no history, FIFO-$K$, or full replay.}
  \label{tab:replay_horizon}
  \begin{tabularx}{\linewidth}{@{}l *{10}{>{\centering\arraybackslash}X}@{}}
    \toprule
    \multirow[c]{2}{*}{\raisebox{-2pt}{\textbf{Methods}}}
      & \multicolumn{2}{c}{\textbf{AIME25}}
      & \multicolumn{2}{c}{\textbf{MinervaMath}}
      & \multicolumn{2}{c}{\textbf{AMC23}}
      & \multicolumn{2}{c}{\textbf{MATH500}}
      & \multicolumn{2}{c}{\textbf{Avg.}} \\
    \cmidrule(lr){2-3}\cmidrule(lr){4-5}\cmidrule(lr){6-7}\cmidrule(lr){8-9}\cmidrule(lr){10-11}
      & \texttt{avg@32} & \texttt{pass@32}
      & \texttt{avg@32} & \texttt{pass@32}
      & \texttt{avg@32} & \texttt{pass@32}
      & \texttt{avg@32} & \texttt{pass@32}
      & \texttt{avg@32} & \texttt{pass@32} \\
    \midrule
    CBPO-Offline
      & 15.83 & 30.00
      & 20.29 & 35.29
      & 50.08 & 77.50
      & 66.99 & 80.40
      & 38.30 & 55.80 \\
    CBPO w/o History
      & 18.75 & 43.33
      & 22.52 & 41.18
      & 57.97 & 90.00
      & 71.54 & 86.00
      & 42.70 & 65.13 \\
    CBPO-Unified FIFO-1
      & 18.96 & 43.33
      & 22.85 & 41.18
      & 58.52 & 92.50
      & 71.96 & 86.20
      & 43.07 & 65.80 \\
    CBPO-Unified FIFO-4
      & 19.27 & 46.67
      & 23.10 & 41.18
      & 58.91 & 92.50
      & 71.91 & 86.20
      & 43.30 & 66.64 \\
    CBPO-Unified FIFO-16
      & 19.69 & 46.67
      & 23.30 & 41.54
      & 59.14 & 92.50
      & 72.27 & 86.40
      & 43.60 & 66.78 \\
    \textbf{CBPO-Unified}
      & \textbf{20.63} & \textbf{50.00}
      & \textbf{24.47} & \textbf{42.28}
      & \textbf{60.16} & \textbf{95.00}
      & \textbf{72.99} & \textbf{87.60}
      & \textbf{44.56} & \textbf{68.72} \\
    \bottomrule
  \end{tabularx}
  \vspace{-10pt}
\end{table}

Extending the finite horizon from one to 16 rounds increases \texttt{avg@32} on all four benchmarks.
Full replay achieves the best \texttt{avg@32} and \texttt{pass@32} on every benchmark, supporting continued reuse of earlier rollouts under the same regression objective.

Appendix~\ref{app:replay_content} reports training-time diagnostics by explicit generating-policy age interval and current regression-residual level for replayed records.
For rollouts older than 16 policy versions, high-residual replay outperforms age-balanced replay on both metrics across all four benchmarks, whereas low-residual replay underperforms it.
Appendix Sections~\ref{app:objective-continuity-examples} and~\ref{app:historical-replay-examples} present matched-prompt response comparisons illustrating objective continuity and historical replay, respectively.

\vspace{-5pt}
\paragraph{Optimization trajectories.}
We project checkpoints from both runs into the shared two-dimensional parameter subspace in Fig.~\ref{fig:optimization_trajectories}.
For each projected path, $L$ is its cumulative length, $D$ is the distance between its endpoints, $L/D$ is the path ratio, and cumulative turning sums the directional changes between successive segments.
Lower path ratio and cumulative turning indicate a more direct optimization trajectory.

Fig.~\ref{fig:optimization_trajectories} and Table~\ref{tab:trajectory_geometry} show that SFT$\rightarrow$DAPO turns sharply at the stage boundary and ends in a lower-reward region ($L/D=1.2207$, cumulative turn $368.75^\circ$), whereas CBPO-Unified follows a more direct projected path toward a higher-reward region ($L/D=1.0205$, $58.20^\circ$).
The lower path ratio and cumulative turning persist with 5 and 10 PCA components (Appendix Figs.~\ref{fig:trajectory_pca5} and~\ref{fig:trajectory_pca10}), and layer normalization changes them only slightly (Appendix Table~\ref{tab:trajectory_layer_scale}).
Appendix Table~\ref{tab:objective_switch_control} reports the objective-switch comparison in projected turning and final \texttt{avg@32}/\texttt{pass@32}.

\vspace{-3pt}
\begin{table}[H]
  \centering
  \scriptsize
  \setlength{\tabcolsep}{5pt}
  \renewcommand{\arraystretch}{1.0}
  \vspace{-5pt}
  \caption{Geometry of the projected optimization trajectories for 2, 5, and 10 components.}
  \label{tab:trajectory_geometry}
  \begin{tabularx}{0.95\linewidth}{@{}>{\raggedright\arraybackslash}X*{4}{>{\centering\arraybackslash}X}@{}}
    \toprule
    \textbf{Methods} & $\boldsymbol{L}$ & $\boldsymbol{D}$ & $\boldsymbol{L/D}$
      & \textbf{Cumulative turn} \\
    \midrule
    \multicolumn{5}{@{}l}{\textit{2 components: 82.8744\%}} \\
    SFT$\rightarrow$DAPO & 2.1000 & 1.7203 & 1.2207 & $368.75^\circ$ \\
    \textbf{CBPO-Unified} & \textbf{1.7631} & \textbf{1.7278} & \textbf{1.0205} & $\boldsymbol{58.20^\circ}$ \\
    \midrule
    \multicolumn{5}{@{}l}{\textit{5 components: 91.6378\%}} \\
    SFT$\rightarrow$DAPO & 2.2498 & 1.8869 & 1.1923 & $295.67^\circ$ \\
    \textbf{CBPO-Unified} & \textbf{1.7688} & \textbf{1.7349} & \textbf{1.0196} & $\boldsymbol{69.76^\circ}$ \\
    \midrule
    \multicolumn{5}{@{}l}{\textit{10 components: 95.9950\%}} \\
    SFT$\rightarrow$DAPO & 2.2956 & 1.9223 & 1.1942 & $290.89^\circ$ \\
    \textbf{CBPO-Unified} & \textbf{1.7804} & \textbf{1.7497} & \textbf{1.0175} & $\boldsymbol{71.22^\circ}$ \\
    \bottomrule
  \end{tabularx}
  \vspace{-14pt}
\end{table}

\subsection{Ablation Studies}

\paragraph{Negative trajectories.}
\label{sec:negative_ratio}
For each DPSK-CoT problem, we retain three trajectories and set the number of negative trajectories to 0, 1, 2, or 3, corresponding to $q_{\mathrm{neg}}\in\{0\%,33.3\%,66.7\%,100\%\}$.
Within each required reward class, trajectories are selected in their original generation order.
As shown in Table~\ref{tab:negative_ratio}, increasing $q_{\mathrm{neg}}$ from $0\%$ to $33.3\%$ improves aggregate \texttt{avg@32/pass@32} by 1.10/1.89 points and produces the best result on every benchmark and metric.
The \texttt{avg@32} improvement is consistent across all four benchmarks, ranging from 0.91 to 1.50 points.
CBPO incorporates incorrect completions through their sequence-level verifier rewards. When a completion's log-probability exceeds its shared target, its regression loss favors reducing the probability of that entire completion.

\vspace{-6pt}
\begin{table}[H]
  \centering
  \scriptsize
  \setlength{\tabcolsep}{2pt}
  \renewcommand{\arraystretch}{1.0}
  \vspace{-5pt}
  \caption{CBPO-Unified $q_{\mathrm{neg}}$ sensitivity with three trajectories per prompt.}
  \label{tab:negative_ratio}
  \begin{tabularx}{\linewidth}{@{}l *{10}{>{\centering\arraybackslash}X}@{}}
    \toprule
    \multirow{2}{*}{$q_{\mathrm{neg}}$}
      & \multicolumn{2}{c}{\textbf{AIME25}}
      & \multicolumn{2}{c}{\textbf{MinervaMath}}
      & \multicolumn{2}{c}{\textbf{AMC23}}
      & \multicolumn{2}{c}{\textbf{MATH500}}
      & \multicolumn{2}{c}{\textbf{Avg.}} \\
    \cmidrule(lr){2-3}\cmidrule(lr){4-5}\cmidrule(lr){6-7}\cmidrule(lr){8-9}\cmidrule(lr){10-11}
      & \texttt{avg@32} & \texttt{pass@32}
      & \texttt{avg@32} & \texttt{pass@32}
      & \texttt{avg@32} & \texttt{pass@32}
      & \texttt{avg@32} & \texttt{pass@32}
      & \texttt{avg@32} & \texttt{pass@32} \\
    \midrule
    $0\%$ & 19.58 & 46.67 & 23.56 & 41.54 & 59.22 & 92.50 & 71.49 & 86.60 & 43.46 & 66.83 \\
    $33.3\%$ & \textbf{20.63} & \textbf{50.00} & \textbf{24.47} & \textbf{42.28} & \textbf{60.16} & \textbf{95.00} & \textbf{72.99} & \textbf{87.60} & \textbf{44.56} & \textbf{68.72} \\
    $66.7\%$ & 19.06 & 43.33 & 23.17 & 41.18 & 59.14 & 92.50 & 71.38 & 86.80 & 43.19 & 65.95 \\
    $100\%$ & 18.65 & 43.33 & 22.94 & 40.44 & 58.13 & 92.50 & 70.28 & 85.60 & 42.50 & 65.47 \\
    \bottomrule
  \end{tabularx}
  \vspace{-11pt}
\end{table}

The trend is non-monotonic.
Increasing $q_{\mathrm{neg}}$ from $33.3\%$ to $66.7\%$ lowers aggregate \texttt{avg@32/pass@32} by 1.37/2.77 points. Using only negative demonstrations lowers them by 2.06/3.25 points.
Negative trajectories are therefore useful as contrastive supervision, but they cannot replace the successful trajectories that anchor high-reward reasoning behavior.
Appendix~\ref{app:negative_controls} separately compares adding 0, 1, or 2 negatives to two fixed positives and varying negative-response reward and content.

\paragraph{Reference sample size.}
Table~\ref{tab:kref_controlled_main} (Left) varies $K_{\mathrm{ref}}\in\{8,16,24,32\}$ with $M=10$, $R_{\mathrm{roll}}=16$, and the optimizer-step budget held fixed.
We start at 8 reference responses, motivated by the MATH500 performance plateau near $N=8$ reported for A$^\star$-PO \citep{brantley2025astarpo}.
The reference responses used in the training loss and the optimization start time are identical across configurations.

\begin{table}[H]
  \centering
  \scriptsize
  \setlength{\tabcolsep}{10pt}
  \renewcommand{\arraystretch}{1.0}
  \vspace{-5pt}
  \caption{Left: $K_{\mathrm{ref}}$ sensitivity with fixed training records, update timing, $M$, and $R_{\mathrm{roll}}$. Right: Current-policy sample size with fixed reference statistics and update timing.}
  \label{tab:kref_controlled_main}
  \label{tab:rollout_fixed_reference}
  \begin{minipage}[t]{0.48\linewidth}
    \centering
    \begin{tabular*}{\linewidth}{@{\extracolsep{\fill}}ccc@{}}
      \toprule
      $\boldsymbol{K_{\mathrm{ref}}}$ & \texttt{avg@32} & \texttt{pass@32} \\
      \midrule
      8 & 42.36 & 66.52 \\
      16 & 44.15 & 69.09 \\
      24 & 44.56 & 68.72 \\
      32 & \textbf{45.27} & \textbf{70.15} \\
      \bottomrule
    \end{tabular*}
  \end{minipage}\hfill%
  \begin{minipage}[t]{0.48\linewidth}
    \centering
    \begin{tabular*}{\linewidth}{@{\extracolsep{\fill}}ccc@{}}
      \toprule
      $\boldsymbol{R_{\mathrm{roll}}}$ & \texttt{avg@32} & \texttt{pass@32} \\
      \midrule
      8  & 42.15 & 67.32 \\
      16 & 44.56 & 68.72 \\
      24 & 45.81 & 70.93 \\
      32 & \textbf{46.43} & \textbf{71.67} \\
      \bottomrule
    \end{tabular*}
  \end{minipage}
\end{table}
\vspace{-5pt}

Aggregate \texttt{avg@32} increases monotonically from 42.36 to 45.27 across the tested reference sample sizes, while \texttt{pass@32} rises overall from 66.52 to 70.15 with a decrease at $K_{\mathrm{ref}}=24$.
Under a fixed budget of 40 generated responses per prompt, the coupled schedule in Table~\ref{tab:kref_schedule_sensitivity} instead attains its highest scores at $(K_{\mathrm{ref}},M,R_{\mathrm{roll}})=(24,10,16)$.
Appendix~\ref{app:normalizer_diagnostics} reports prompt-level reference success and confidence bounds for normalizer estimation.

\paragraph{Current-policy sample size.}
All reference preprocessing is completed before the first optimizer step.
Each run uses 1,280 optimizer steps, $M=10$, and four current-policy collection passes over the same prompts, with matched optimizer and collection positions across configurations.
For the rollout-only comparison in Table~\ref{tab:rollout_fixed_reference} (Right), $K_{\mathrm{ref}}=24$ and the reference training records remain fixed.
We vary $R_{\mathrm{roll}}\in\{8,16,24,32\}$ by collecting $R_{\mathrm{roll}}/4$ responses per prompt at each of the four collection passes.

\begin{wraptable}{r}{0.52\textwidth}
  \vspace{-\intextsep}
  \centering
  \scriptsize
  \setlength{\tabcolsep}{3pt}
  \renewcommand{\arraystretch}{1.0}
  \setlength{\abovecaptionskip}{11pt}
  \setlength{\belowcaptionskip}{11pt}
  \vspace{-10pt}
  \caption{Budget-coupled $K_{\mathrm{ref}},M,R_{\mathrm{roll}}$ sensitivity.}
  \vspace{-10pt}
  \label{tab:kref_schedule_sensitivity}
  \begin{tabular*}{\linewidth}{@{\extracolsep{\fill}}ccccc@{}}
    \toprule
    $\boldsymbol{K_{\mathrm{ref}}}$ & $\boldsymbol{M}$
      & $\boldsymbol{R_{\mathrm{roll}}}$
      & \texttt{avg@32} & \texttt{pass@32} \\
    \midrule
    8 & 5 & 32 & 37.23 & 60.57 \\
    20 & 8 & 20 & 40.65 & 64.49 \\
    24 & 10 & 16 & \textbf{44.56} & \textbf{68.72} \\
    32 & 20 & 8 & 42.26 & 66.97 \\
    \bottomrule
  \end{tabular*}
  \vspace{-10pt}
\end{wraptable}

\paragraph{Reference--rollout budget allocation.}
In the epoch-coupled sweep in Table~\ref{tab:kref_schedule_sensitivity}, $R_{\mathrm{roll}}=EG-K_{\mathrm{ref}}$ with $E=10$ and $G=4$, giving a fixed budget of 40 generated responses per prompt.
Changing $K_{\mathrm{ref}}$ changes the onset of updates and the current-policy sample count, while $M$ is adjusted to maintain the total budget of 1,280 optimizer steps.
The configuration $(K_{\mathrm{ref}},M,R_{\mathrm{roll}})=(24,10,16)$ achieves the highest aggregate scores among these jointly varied configurations, reaching 44.56 and 68.72.
Allocating 32 responses to reference estimation leaves eight current-policy responses and yields 42.26/66.97, supporting the default allocation of 24 reference and 16 current-policy responses under this budget.

\subsection{Evaluations Beyond Mathematical Reasoning}
\label{sec:evaluations_beyond_mathematical_reasoning}

We evaluate general-domain post-training by initializing PPO and CBPO-Unified from Qwen3-1.7B and training on the General subset of HelpSteer3-Preference with Qwen3-Nemotron-32B-GenRM-Principle as a learned reward model. With GPT-6-Luna-Max as judge, CBPO-Unified reaches an 8.1\% Arena-Hard v2 win rate and a 6.91 MT-Bench mean score, compared with 7.5\% and 6.35 for PPO (Table~\ref{tab:general_domain_posttraining}).

We also evaluate coding, instruction-following, and scientific question answering after math-focused training by comparing each Qwen3 checkpoint with its CBPO-Unified counterpart. On LiveCodeBench v5 pass@1, IFEval prompt-level strict accuracy, and GPQA-Diamond accuracy, the changes for 1.7B are $-0.17$, $-0.37$, and $-0.51$ percentage points, respectively. For 8B, they are $-0.04$, $-0.18$, and $0.00$ points (Table~\ref{tab:cross_domain_evaluation}).

%% file: sections/6_conclusion.tex
\section{Conclusion}
We introduced Cross-Behavior Policy Optimization, which makes Offline Demonstrations and On- and Off-Policy Rollouts compatible with one regression loss through a common record representation. Verifier rewards, fixed-reference log-probabilities, and frozen prompt normalizers provide the target fields for both successful and unsuccessful completions. This representation supports their continued reuse without behavior-policy likelihoods, importance ratios, a critic, or an objective switch.

Across two model scales, CBPO improves each restricted data regime, while their combination performs best. Compared with SFT$\rightarrow$RLVR, CBPO follows a substantially more direct path toward a higher-reward policy, and historical replay provides further gains. Balanced negative demonstrations and accurate reference normalization further strengthen the unified method. These findings establish CBPO as an effective alternative to stage-wise reasoning post-training.

%% file: sections/7_limitations.tex
\section{Limitations}
CBPO also incurs an upfront generation cost to estimate the frozen prompt normalizer, and its behavior under offline distribution shifts across domains or model families remains untested.

%% file: appendix.tex
\addtocontents{toc}{\protect\setcounter{tocdepth}{2}}
\begingroup
  \hypersetup{linktoc=all}
  \normalfont
  \linespread{1.18}\selectfont
  \setlength{\parindent}{0pt}
  \setlength{\parskip}{0pt}
  \makeatletter
  \renewcommand{\@pnumwidth}{2em}
  \renewcommand{\@tocrmarg}{3em}
  \renewcommand{\l@section}[2]{%
    \ifnum\c@tocdepth>\z@
      \addpenalty{\@secpenalty}%
      \addvspace{1.2em}%
      \begingroup
        \parindent\z@
        \leftskip2em
        \rightskip\@pnumwidth
        \parfillskip-\@pnumwidth
        \@tempdima1.5em
        \leavevmode\bfseries
        \advance\leftskip\@tempdima
        \hskip-\@tempdima
        #1\nobreak\hfil\nobreak
        \hb@xt@\@pnumwidth{\hss #2}\par
      \endgroup
    \fi}
  \renewcommand{\l@subsection}[2]{%
    \ifnum\c@tocdepth>\@ne
      \addvspace{2pt}%
      \@dottedtocline{2}{3.5em}{2.5em}{#1}{#2}%
    \fi}
  \pdfbookmark[0]{Table of Contents}{appendix.contents}
  {\Large\bfseries Table of Contents\par}
  \vspace{0.25em}
  \hrule height0.4pt
  \vspace{0.25em}
  \@starttoc{toc}
  \vfill
  \hrule height0.4pt
  \makeatother
\endgroup
\clearpage

\section*{Appendix}
This appendix provides the theoretical analysis, complete algorithmic details, data and verifier specifications, training configurations, additional results, diagnostic protocols, and qualitative examples for the main paper.

\setcounter{secnumdepth}{2}
\counterwithin{equation}{section}
\setcounter{algorithm}{0}
\counterwithin{figure}{section}
\counterwithin{table}{section}
\renewcommand{\thealgorithm}{S\arabic{algorithm}}
\renewcommand{\theHequation}{appendix.\thesection.\arabic{equation}}
\renewcommand{\theHalgorithm}{appendix.\arabic{algorithm}}
\renewcommand{\theHfigure}{appendix.\thesection.\arabic{figure}}
\renewcommand{\theHtable}{appendix.\thesection.\arabic{table}}

\section{Theoretical Analysis}
\label{app:mc-logz-bias}

\subsection{Behavior-source invariance of the CBPO target}
\label{app:behavior_source_invariance}
For a stored record, let $\mu$ denote the behavior policy that produced its completion.
After verifier scoring and reference preprocessing, the record contains
\[
    \bigl(x,y,r(x,y),\log\pi_{\mathrm{ref}}(y\!\mid\!x),
    \log\widehat Z_{\mathrm{ref}}(x)\bigr).
\]
Its CBPO target and loss are
\[
    \widehat t(x,y)
    =
    \log\pi_{\mathrm{ref}}(y\!\mid\!x)
    +\frac{r(x,y)}{\beta}
    -\log\widehat Z_{\mathrm{ref}}(x).
\]
\[
    \ell_{\mathrm{CBPO}}(x,y;\theta)
    =
    \left(\log\pi_\theta(y\!\mid\!x)-\widehat t(x,y)\right)^2.
\]
Neither expression contains $\mu$.
Consequently, two records with identical logged fields induce the same per-sample loss and gradient even if they were collected by different behavior policies.
This is the precise sense in which CBPO uses one shared target across data sources.
The buffer-sampling distribution determines the weight of each record in the aggregate objective.

\subsection{Finite-sample bias of the frozen prompt normalizer}
This section analyzes the finite-sample bias introduced by replacing the
exact reference-policy partition function with its Monte-Carlo estimate in
the CBPO residual. The analysis clarifies that the direct estimator of
\(Z_{\mathrm{ref}}(x)\) is unbiased, whereas the log-partition term
\(\log \widehat Z_{\mathrm{ref}}(x)\) is biased at finite sample size. Under
the bounded-reward and bounded-length conditions used in our implementation,
this bias is bounded and vanishes as the number of reference-policy samples
used for partition estimation increases.

For a fixed prompt \(x\), define
\[
    W(x,y)
    =
    e^{\frac{r(x,y)}{\beta}},
    \qquad
    Z_{\mathrm{ref}}(x)
    =
    \mathbb{E}_{y\sim \pi_{\mathrm{ref}}(\cdot\mid x)}
    \left[W(x,y)\right].
\]
Let \(y_1,\ldots,y_N\overset{\mathrm{i.i.d.}}{\sim}
\pi_{\mathrm{ref}}(\cdot\!\mid\!x)\), where \(N\) denotes the number of
reference-policy samples used to estimate the prompt-level partition
function. We denote this sample count by \(K_{\mathrm{ref}}\), so
\(N=K_{\mathrm{ref}}\) in all experimental descriptions. In the
implementation-coupled runs, the same \(K_{\mathrm{ref}}\) also sets the
update onset through \(eG>K_{\mathrm{ref}}\). The Monte-Carlo estimator is
\[
    \widehat Z_{\mathrm{ref},N}(x)
    =
    \frac{1}{N}
    \sum_{i=1}^{N}
    W(x,y_i)
    =
    \frac{1}{N}
    \sum_{i=1}^{N}
    e^{\frac{r(x,y_i)}{\beta}}.
\]
Then
\[
    \mathbb{E}\left[
        \widehat Z_{\mathrm{ref},N}(x)
    \right]
    =
    Z_{\mathrm{ref}}(x),
\]
so the direct estimator of \(Z_{\mathrm{ref}}(x)\) is unbiased. However,
CBPO uses the log-partition estimate
\(\log \widehat Z_{\mathrm{ref},N}(x)\) inside the regression target.
Since the logarithm is concave, Jensen's inequality gives
\[
    \mathbb{E}
    \left[
        \log \widehat Z_{\mathrm{ref},N}(x)
    \right]
    \le
    \log
    \mathbb{E}
    \left[
        \widehat Z_{\mathrm{ref},N}(x)
    \right]
    =
    \log Z_{\mathrm{ref}}(x).
\]
Thus the log-partition estimate is generally downward biased at finite
\(N\).

\paragraph{Bias bound.}
Assume that, for the fixed prompt \(x\), there exists a constant
\(a_x>0\) such that
\[
    W(x,y)=e^{\frac{r(x,y)}{\beta}}\ge a_x,
    \quad y\sim \pi_{\mathrm{ref}}(\cdot\!\mid\!x)\quad \text{a.s.}
\]
Also assume
\[
    \sigma_x^2
    =
    \operatorname{Var}_{y\sim\pi_{\mathrm{ref}}(\cdot\mid x)}
    \left[
        W(x,y)
    \right].
\]
Then
\[
    \operatorname{Var}
    \left[
        \widehat Z_{\mathrm{ref},N}(x)
    \right]
    =
    \frac{\sigma_x^2}{N}.
\]

We now bound the bias of
\(\log \widehat Z_{\mathrm{ref},N}(x)\). For compactness, write
\[
    Z = Z_{\mathrm{ref}}(x),
    \qquad
    \widehat Z_N = \widehat Z_{\mathrm{ref},N}(x).
\]
By Taylor's theorem applied to \(f(z)=\log z\) around \(Z\),
\[
    \log \widehat Z_N
    =
    \log Z
    +
    \frac{\widehat Z_N-Z}{Z}
    -
    \frac{(\widehat Z_N-Z)^2}{2\xi_N^2},
\]
where \(\xi_N\) lies between \(Z\) and \(\widehat Z_N\). Since
\(W(x,y)\ge a_x\), we have \(\widehat Z_N\ge a_x\) and
\(Z\ge a_x\), hence \(\xi_N\ge a_x\). Taking expectation and using
\(\mathbb{E}[\widehat Z_N-Z]=0\), we obtain
\[
    \log Z
    -
    \mathbb{E}
    \left[
        \log \widehat Z_N
    \right]
    =
    \mathbb{E}
    \left[
        \frac{(\widehat Z_N-Z)^2}{2\xi_N^2}
    \right].
\]
Therefore,
\[
    0
    \le
    \log Z
    -
    \mathbb{E}
    \left[
        \log \widehat Z_N
    \right]
    \le
    \frac{1}{2a_x^2}
    \mathbb{E}
    \left[
        (\widehat Z_N-Z)^2
    \right].
\]
Since \(\widehat Z_N\) is unbiased,
\[
    \mathbb{E}
    \left[
        (\widehat Z_N-Z)^2
    \right]
    =
    \operatorname{Var}
    \left[
        \widehat Z_N
    \right]
    =
    \frac{\sigma_x^2}{N}.
\]
Thus
\begin{equation}
    0
    \le
    \log Z_{\mathrm{ref}}(x)
    -
    \mathbb{E}
    \left[
        \log \widehat Z_{\mathrm{ref},N}(x)
    \right]
    \le
    \frac{\sigma_x^2}{2Na_x^2}.
    \label{eq:logz-bias-bound}
\end{equation}
Equivalently,
\[
    \left|
    \mathbb{E}
    \left[
        \log \widehat Z_{\mathrm{ref},N}(x)
    \right]
    -
    \log Z_{\mathrm{ref}}(x)
    \right|
    \le
    \frac{\sigma_x^2}{2Na_x^2}.
\]
Hence the log-partition bias is \(O(1/N)\) and converges to zero as
\(N\to\infty\).

\paragraph{Gradient bias.}
We next translate the log-partition bias into a gradient-bias bound for the
CBPO objective at a fixed parameter value $\theta$, holding the collection distribution fixed.
Assume that the gradient expectations are integrable and differentiation may be interchanged with expectation.
Define the exact CBPO residual
\[
    \Delta_\theta(x,y)
    =
    \log
    \frac{\pi_\theta(y\!\mid\!x)}{\pi_{\mathrm{ref}}(y\!\mid\!x)}
    -
    \frac{r(x,y)}{\beta}
    +
    \log Z_{\mathrm{ref}}(x),
\]
and the Monte-Carlo residual
\[
    \widehat\Delta_{\theta,N}(x,y)
    =
    \log
    \frac{\pi_\theta(y\!\mid\!x)}{\pi_{\mathrm{ref}}(y\!\mid\!x)}
    -
    \frac{r(x,y)}{\beta}
    +
    \log \widehat Z_{\mathrm{ref},N}(x).
\]
Let \(q_t(y\!\mid\!x)\) denote the regime-specific training-record distribution
over completions for prompt \(x\) at training iteration \(t\).
Its support can include the applicable offline, reference, current, and historical records.
The exact and Monte-Carlo prompt-conditional losses are
\[
    \mathcal{L}_x(\theta)
    =\mathbb{E}_{y\sim q_t(\cdot\mid x)}
        \left[\Delta_\theta(x,y)^2\right],\qquad
    \widehat{\mathcal{L}}_{x,N}(\theta)
    =\mathbb{E}_{y\sim q_t(\cdot\mid x)}
        \left[\widehat\Delta_{\theta,N}(x,y)^2\right].
\]
Because \(\pi_{\mathrm{ref}}\) and
\(\widehat Z_{\mathrm{ref},N}(x)\) are fixed during policy optimization,
\[
    \nabla_\theta
    \log
    \frac{\pi_\theta(y\!\mid\!x)}{\pi_{\mathrm{ref}}(y\!\mid\!x)}
    =
    \nabla_\theta \log \pi_\theta(y\!\mid\!x).
\]
Define the score-gradient term
\[
    A_\theta(x,y)
    =
    \nabla_\theta
    \log
    \frac{\pi_\theta(y\!\mid\!x)}{\pi_{\mathrm{ref}}(y\!\mid\!x)}.
\]
Assume its first moment under the training-record distribution is finite, and define
\[
    M_x(\theta)
    =
    \mathbb{E}_{y\sim q_t(\cdot\mid x)}
    \left[
        \left\|
        A_\theta(x,y)
        \right\|
    \right].
\]
Assume also that the partition-estimation samples used to construct
\(\widehat Z_{\mathrm{ref},N}(x)\) are independent of the training record
\(y\) used in the loss. Then
\[
    \nabla_\theta \mathcal{L}_x(\theta)
    =
    2
    \mathbb{E}_{y\sim q_t(\cdot\mid x)}
    \left[
        \Delta_\theta(x,y) A_\theta(x,y)
    \right],
\]
and
\[
    \nabla_\theta \widehat{\mathcal{L}}_{x,N}(\theta)
    =
    2
    \mathbb{E}_{y\sim q_t(\cdot\mid x)}
    \left[
        \widehat\Delta_{\theta,N}(x,y) A_\theta(x,y)
    \right].
\]
Let
\[
    \epsilon_N(x)
    =
    \log \widehat Z_{\mathrm{ref},N}(x)
    -
    \log Z_{\mathrm{ref}}(x).
\]
Then
\[
    \widehat\Delta_{\theta,N}(x,y)
    =
    \Delta_\theta(x,y)
    +
    \epsilon_N(x).
\]
Taking expectation over the partition-estimation randomness gives
\[
    \mathbb{E}
    \left[
        \nabla_\theta
        \widehat{\mathcal{L}}_{x,N}(\theta)
    \right]
    -
    \nabla_\theta
    \mathcal{L}_x(\theta)
    =
    2
    \mathbb{E}
    \left[
        \epsilon_N(x)
    \right]
    \mathbb{E}_{y\sim q_t(\cdot\mid x)}
    \left[
        A_\theta(x,y)
    \right].
\]
Therefore,
\[
    \left\|
    \mathbb{E}
    \left[
        \nabla_\theta
        \widehat{\mathcal{L}}_{x,N}(\theta)
    \right]
    -
    \nabla_\theta
    \mathcal{L}_x(\theta)
    \right\|
    \le
    2
    \left|
    \mathbb{E}
    \left[
        \epsilon_N(x)
    \right]
    \right|
    \mathbb{E}_{y\sim q_t(\cdot\mid x)}
    \left[
        \left\|
        A_\theta(x,y)
        \right\|
    \right].
\]
Using Eq.~\eqref{eq:logz-bias-bound},
\begin{equation}
    \left\|
    \mathbb{E}
    \left[
        \nabla_\theta
        \widehat{\mathcal{L}}_{x,N}(\theta)
    \right]
    -
    \nabla_\theta
    \mathcal{L}_x(\theta)
    \right\|
    \le
    \frac{\sigma_x^2}{Na_x^2}
    M_x(\theta).
    \label{eq:gradient-bias-bound}
\end{equation}
The gradient bias induced by the Monte-Carlo log-partition estimate is
bounded at finite \(N\) and vanishes as
\(N\to\infty\) under the stated first-moment assumption.

Reusing reference-policy samples for both normalizer estimation and training
introduces a covariance term. A leave-one-out partition estimate removes
this dependence. When the replay
sample is drawn from the same \(N\) reference samples used to estimate
\(\widehat Z_{\mathrm{ref},N}(x)\), and if
\(\|A_\theta(x,y)\|\le \overline M_x\), Taylor expansion yields the bound
\[
\begin{aligned}
    B_x
    &:=
    \mathbb{E}\!\left[
        \|A_\theta(x,Y)\|
        |W(x,Y)-Z_{\mathrm{ref}}(x)|
    \right], \\
    &\left\|\mathbb{E}[\widehat g_N]-g\right\|
    \le
    \frac{2B_x}{NZ_{\mathrm{ref}}(x)}
    +
    \frac{\overline M_x\sigma_x^2}{Na_x^2}.
\end{aligned}
\]
where \(g\) is the exact reference-sampling gradient and
\(\widehat g_N\) is the empirical gradient using
\(\log \widehat Z_{\mathrm{ref},N}(x)\). Under bounded rewards and bounded
score gradients, this dependence-induced term is also \(O(1/N)\). Therefore,
the same asymptotic conclusion holds.

For completeness, fix $\theta$ and write
$A_i=A_\theta(x,Y_i)$, $W_i=W(x,Y_i)$, and
$Z=Z_{\mathrm{ref}}(x)$ for independent reference draws $Y_1,\ldots,Y_N$.
Let
$\widehat g_N=2N^{-1}\sum_i(\Delta_\theta(x,Y_i)+\epsilon_N(x))A_i$
and $g=2\mathbb E[\Delta_\theta(x,Y_1)A_1]$.
Exchangeability gives
$\mathbb E[\widehat g_N]-g=2\mathbb E[\epsilon_N(x)A_1]$.
The same Taylor expansion as above yields
\[
    \epsilon_N(x)
    =\frac{1}{NZ}\sum_{i=1}^N(W_i-Z)
     -\frac{(\widehat Z_N-Z)^2}{2\xi_N^2},
    \qquad \xi_N\ge a_x.
\]
For $i\ne1$, independence gives
$\mathbb E[(W_i-Z)A_1]=0$.
The remaining linear term has norm at most $2B_x/(NZ)$.
The remainder has norm at most
$\overline M_x\,\mathbb E[(\widehat Z_N-Z)^2]/a_x^2
=\overline M_x\sigma_x^2/(Na_x^2)$, proving the stated bound.
For $N>1$, omitting $Y_1$ from its own normalizer estimate makes that estimate independent of $A_1$ at fixed $\theta$, which removes the covariance term and leaves the log-normalizer bias for $N-1$ samples.

\paragraph{Satisfaction of the assumptions in our setting.}
Our math RLVR verifier assigns rewards in \(\{0,0.1,1.0\}\subset[0,1]\).
For any fixed \(\beta>0\),
\[
    1
    \le
    e^{\frac{r(x,y)}{\beta}}
    \le
    e^{\frac{1}{\beta}}.
\]
Thus \(a_x=1\), and the variance satisfies
\[
    \sigma_x^2
    \le
    \frac{
        \left(
            e^{1/\beta}-1
        \right)^2
    }{4}.
\]
For the default \(\beta=0.2\), the weight \(e^{r/\beta}\) lies in \([1,e^5]\).
At fixed \(\theta\), finite per-completion score gradients and finite replay support ensure the first-moment condition on \(M_x(\theta)\) required by Eq.~\eqref{eq:gradient-bias-bound}.

\subsection{Population consistency and regime-dependent approximation}
\label{app:regime_error}

\paragraph{Setting and notation.}
Fix the reference policy, reward function, and $\beta>0$ from Eq.~\eqref{eq:cbpo_opt_policy}.
Let $\rho$ be the prompt distribution in the KL-regularized objective.
For clarity, prompts and completions are discrete, and
$\mathcal Y_x=\{y:\pi_{\mathrm{ref}}(y\mid x)>0\}$ is the completion support for prompt $x$.
Assume bounded rewards and normalized candidate policies that are positive on $\mathcal Y_x$ and zero outside it.
These conditions make $\pi^\star$ positive on the same support.
All collection laws and archives are supported on pairs with $y\in\mathcal Y_x$.
All expectations below are assumed finite.
The policy class $\Pi$ contains the shared-parameter policies being fitted, so a single $\pi\in\Pi$ specifies a distribution for every prompt.
Define
\begin{equation}
\begin{aligned}
    J_\rho(\pi)
    &=
    \mathbb E_{x\sim\rho}\!\left[
        \mathbb E_{y\sim\pi(\cdot\mid x)}r(x,y)
        -\beta D_{\mathrm{KL}}\!\left(
            \pi(\cdot\mid x)\,\|\,\pi_{\mathrm{ref}}(\cdot\mid x)
        \right)
    \right],\\
    e_\pi(x,y)
    &=\log\pi(y\mid x)-\log\pi^\star(y\mid x),\qquad
    \|f\|_q
    =\sqrt{\mathbb E_{(x,y)\sim q}[f(x,y)^2]}.
\end{aligned}
\label{eq:regime_notation}
\end{equation}
For $\pi=\pi_\theta$, $e_\pi$ equals the exact residual $\Delta_\theta$ defined above.

A regime $g$ induces a joint population training-record distribution $q_g(x,y)$.
It includes both prompt frequencies and completion frequencies, and its prompt marginal can differ from $\rho$.
For a finite archive of $n_g$ indexed records, the empirical distribution and exact population risk are
\begin{equation}
    \widehat q_g(x,y)
    =\frac{1}{n_g}\sum_{i=1}^{n_g}
        \mathbf 1\{(x_i,y_i)=(x,y)\},\qquad
    L_g(\pi)=\|e_\pi\|_{q_g}^2.
\label{eq:regime_distributions}
\end{equation}
Duplicate completions remain distinct records in the sum.
$\widehat q_g$ is the joint distribution induced by the uniform record-index sampler in Section~\ref{sec:pr_schedule}.
The distribution $q_g$ describes collection at the population level, whereas $\widehat q_g$ describes the records actually available for optimization.
For adaptive collection, $q_g$ will be the average conditional collection law defined in Eq.~\eqref{eq:adaptive_collection_law}.

\paragraph{KL objective-gap identity.}
The reward-tilted policy in Eq.~\eqref{eq:cbpo_opt_policy} satisfies
\begin{equation}
    J_\rho(\pi^\star)-J_\rho(\pi)
    =
    \beta\,\mathbb E_{x\sim\rho}
    D_{\mathrm{KL}}\!\left(
        \pi(\cdot\mid x)\,\|\,\pi^\star(\cdot\mid x)
    \right).
    \label{eq:regime_kl_gap}
\end{equation}
To verify the identity, substitute
$r(x,y)=\beta\log(\pi^\star(y\mid x)/\pi_{\mathrm{ref}}(y\mid x))
+\beta\log Z_{\mathrm{ref}}(x)$ into $J_\rho$ to obtain
\[
    J_\rho(\pi)
    =\beta\mathbb E_\rho\log Z_{\mathrm{ref}}(x)
    -\beta\mathbb E_\rho
        D_{\mathrm{KL}}(\pi(\cdot\mid x)\,\|\,\pi^\star(\cdot\mid x)).
\]
At $\pi=\pi^\star$, the KL term is zero.
Subtracting gives Eq.~\eqref{eq:regime_kl_gap} and establishes the optimality of $\pi^\star$ over all supported policies.
This is the standard fixed-reference KL-regularized policy identity \citep{rafailov2023dpo}.

\paragraph{Proposition 1 (Population consistency).}
Suppose the normalizer is exact, $\pi^\star\in\Pi$, and
\begin{equation}
    q_g(x,y)>0
    \quad\text{whenever}\quad
    \rho(x)\pi^\star(y\mid x)>0.
    \label{eq:regime_qualitative_coverage}
\end{equation}
Then $L_g$ has minimum zero, and every global minimizer equals $\pi^\star$ for $\rho$-almost every prompt.
Consequently, all regimes satisfying these assumptions share the same optimal policy on the objective prompt distribution.
The statement also holds for a source mixture whenever its combined support satisfies Eq.~\eqref{eq:regime_qualitative_coverage}.

\paragraph{Proof.}
Since $e_{\pi^\star}=0$, realizability gives $L_g(\pi^\star)=0$.
Nonnegativity of the squared loss implies that every minimizer $\pi_g$ has $e_{\pi_g}=0$ $q_g$-almost surely.
Condition~\eqref{eq:regime_qualitative_coverage} transfers this equality to every pair with positive $\rho(x)\pi^\star(y\mid x)$.
Thus $\pi_g(y\mid x)=\pi^\star(y\mid x)$ throughout $\mathcal Y_x$ for $\rho$-almost every $x$.
Both policies assign zero mass outside $\mathcal Y_x$.
For a mixture $q=\sum_s\alpha_s q_s$ with nonnegative weights summing to one, its support is the union of the supports of sources with $\alpha_s>0$.
The same argument therefore applies to the mixture.

\paragraph{Approximation under a restricted policy class.}
Without realizability, exact population regression selects a best $q_g$-weighted fit,
$\inf_{\pi\in\Pi}\|\log\pi-\log\pi^\star\|_{q_g}$.
Changing $q_g$ changes these weights and can change the best fit.
The following results quantify its effect on $J_\rho$, together with finite records, estimated targets, and incomplete optimization.

\subsection{From training-record error to the KL-regularized objective}

\paragraph{Lemma 1 (Distribution-shift transfer).}
For any fitted policy $\widehat\pi_g$, define its objective joint distribution
$p_g(x,y)=\rho(x)\widehat\pi_g(y\mid x)$.
Assume $|e_{\widehat\pi_g}(x,y)|\le B$ on the relevant support for some $B<\infty$.
For any $C\ge1$, let
\begin{equation}
    S_g(C)=\{(x,y):p_g(x,y)\le Cq_g(x,y)\},\qquad
    \kappa_g(C)=\sum_{(x,y)\notin S_g(C)}p_g(x,y).
\label{eq:regime_coverage_tail}
\end{equation}
Then
\begin{equation}
    J_\rho(\pi^\star)-J_\rho(\widehat\pi_g)
    \le
    \beta\sqrt C\,\|e_{\widehat\pi_g}\|_{q_g}
    +\beta B\,\kappa_g(C).
    \label{eq:regime_transfer}
\end{equation}
Here $\kappa_g(C)$ includes both uncovered pairs with $q_g=0<p_g$ and pairs whose objective-to-training density ratio exceeds $C$.

\paragraph{Proof.}
By Eq.~\eqref{eq:regime_kl_gap}, the objective gap divided by $\beta$ equals $\mathbb E_{p_g}e_{\widehat\pi_g}$.
Split this expectation over $S_g(C)$ and its complement, and bound it by the corresponding absolute expectations.
Cauchy--Schwarz and $p_g\le Cq_g$ on $S_g(C)$ give
\[
    \sum_{S_g(C)}p_g|e_{\widehat\pi_g}|
    \le
    \sqrt{
        \left(\sum_{S_g(C)}p_g\right)
        \left(\sum_{S_g(C)}p_g e_{\widehat\pi_g}^2\right)
    }
    \le\sqrt C\,\|e_{\widehat\pi_g}\|_{q_g}.
\]
The complementary sum is at most $B\kappa_g(C)$.
Multiplying by $\beta$ proves the claim.
When $p_g\le Cq_g$ everywhere, $\kappa_g(C)=0$.
Thus quantitative coverage controls error transfer, while support coverage alone gives the identifiability result in Proposition 1.

\paragraph{Error definitions.}
Write
$\epsilon_Z(x)=\log\widehat Z_{\mathrm{ref}}(x)-\log Z_{\mathrm{ref}}(x)$,
which is $\epsilon_N(x)$ above with $N=K_{\mathrm{ref}}$.
The fitted empirical risk is
\begin{equation}
    \widehat L_g^Z(\pi)
    =\|e_\pi+\epsilon_Z\|_{\widehat q_g}^2.
    \label{eq:regime_empirical_risk}
\end{equation}
Define four errors in root-mean-square units:
\begin{equation}
\begin{aligned}
    \varepsilon_{\mathrm{app},g}
    &=\inf_{\pi\in\Pi}\|e_\pi\|_{q_g},\qquad
    \varepsilon_{Z,g}
    =\|\epsilon_Z\|_{\widehat q_g},\\
    \varepsilon_{\mathrm{sam},g}
    &=\sup_{\pi\in\Pi}
        \left|\|e_\pi\|_{q_g}-\|e_\pi\|_{\widehat q_g}\right|,\\
    \varepsilon_{\mathrm{opt},g}
    &=\|e_{\widehat\pi_g}+\epsilon_Z\|_{\widehat q_g}
      -\inf_{\pi\in\Pi}\|e_\pi+\epsilon_Z\|_{\widehat q_g}.
\end{aligned}
\label{eq:regime_error_terms}
\end{equation}
The first term measures the best approximation in the policy class under the regime's training distribution.
The second measures the frozen target error on the archive.
The third measures population-to-archive sampling error uniformly over candidate policies.
The fourth is the remaining empirical optimization error and is nonnegative for $\widehat\pi_g\in\Pi$.
If the achieved squared empirical loss is within $\eta_{\mathrm{opt}}$ of its infimum, then
$\varepsilon_{\mathrm{opt},g}\le\sqrt{\eta_{\mathrm{opt}}}$.

\paragraph{Lemma 2 (Regression error decomposition).}
For any finite archive and any fitted $\widehat\pi_g\in\Pi$,
\begin{equation}
    \|e_{\widehat\pi_g}\|_{q_g}
    \le
    \varepsilon_{\mathrm{app},g}
    +2\varepsilon_{Z,g}
    +2\varepsilon_{\mathrm{sam},g}
    +\varepsilon_{\mathrm{opt},g}.
    \label{eq:regime_regression_bound}
\end{equation}
This is a deterministic statement for the realized data, frozen estimates, and fitted policy.

\paragraph{Proof.}
For any $\eta>0$, choose $\pi_\eta\in\Pi$ with
$\|e_{\pi_\eta}\|_{q_g}\le\varepsilon_{\mathrm{app},g}+\eta$.
The definitions and the triangle inequality imply
\[
\begin{aligned}
    \|e_{\widehat\pi_g}\|_{q_g}
    &\le \|e_{\widehat\pi_g}\|_{\widehat q_g}
          +\varepsilon_{\mathrm{sam},g}\\
    &\le \|e_{\widehat\pi_g}+\epsilon_Z\|_{\widehat q_g}
          +\varepsilon_{Z,g}+\varepsilon_{\mathrm{sam},g}\\
    &\le \|e_{\pi_\eta}+\epsilon_Z\|_{\widehat q_g}
          +\varepsilon_{\mathrm{opt},g}
          +\varepsilon_{Z,g}+\varepsilon_{\mathrm{sam},g}\\
    &\le \|e_{\pi_\eta}\|_{\widehat q_g}
          +2\varepsilon_{Z,g}
          +\varepsilon_{\mathrm{sam},g}+\varepsilon_{\mathrm{opt},g}\\
    &\le \varepsilon_{\mathrm{app},g}+\eta
          +2\varepsilon_{Z,g}
          +2\varepsilon_{\mathrm{sam},g}+\varepsilon_{\mathrm{opt},g}.
\end{aligned}
\]
Letting $\eta$ decrease to zero proves the result.
The inequality uses no independence between the reference samples, the archive, and $\widehat\pi_g$, so it also covers reuse of reference responses for training.
The bound on $\varepsilon_{\mathrm{opt},g}$ follows from
$|\sqrt u-\sqrt v|\le\sqrt{|u-v|}$ for $u,v\ge0$.

\paragraph{Theorem 1 (Regime-dependent objective gap).}
Under the assumptions of Lemma 1 and the definitions in Eq.~\eqref{eq:regime_error_terms}, for every $C\ge1$,
\begin{equation}
    J_\rho(\pi^\star)-J_\rho(\widehat\pi_g)
    \le
    \beta\sqrt C\,
    \bigl(
        \varepsilon_{\mathrm{app},g}
        +2\varepsilon_{Z,g}
        +2\varepsilon_{\mathrm{sam},g}
        +\varepsilon_{\mathrm{opt},g}
    \bigr)
    +\beta B\,\kappa_g(C).
\label{eq:regime_objective_bound}
\end{equation}

\paragraph{Proof.}
Apply Lemma 1 to $\widehat\pi_g$ and substitute Lemma 2.
All quantities are evaluated for the same realized archive and regime distribution, so the substitution is valid even when collection is adaptive.

Equation~\eqref{eq:regime_objective_bound} separates normalizer estimation, finite-record sampling, function approximation, and optimization.
Distribution shift enters through the transfer factor $C$ and residual coverage mass $\kappa_g(C)$.
The bound applies to the objective under the specified $\rho$.
Under exact targets, exact population optimization, and realizability, Proposition 1 gives equality with $\pi^\star$ whenever its support condition holds.
For finite data and a restricted class, Eq.~\eqref{eq:regime_objective_bound} instead quantifies how the regime-specific fit can differ from this optimum.

\subsection{Finite-reference target error and finite-record bounds}
\label{app:regime_finite_sample}

\paragraph{Proposition 2 (Normalizer mean-square and uniform bounds).}
For each fixed prompt, let $W=e^{r/\beta}$ and suppose
$0<a\le W\le b<\infty$.
Construct $\widehat Z_{\mathrm{ref}}(x)$ from $K=K_{\mathrm{ref}}$ independent draws from the reference distribution defining $Z_{\mathrm{ref}}(x)$.
Then
\begin{equation}
    \mathbb E[\epsilon_Z(x)^2]
    \le \frac{\sigma_x^2}{Ka^2}.
    \label{eq:regime_normalizer_mse}
\end{equation}
For a fixed set $\mathcal X_0$ of $m$ training prompts and any $\delta_Z\in(0,1)$, define
\begin{equation}
    h_Z
    =(b-a)\sqrt{\frac{\log(2m/\delta_Z)}{2K}},\qquad
    u_Z
    =\min\!\left\{
        \log\frac ba,\,
        \log\!\left(1+\frac{h_Z}{a}\right)
    \right\}.
\label{eq:regime_normalizer_radius}
\end{equation}
With probability at least $1-\delta_Z$,
$\max_{x\in\mathcal X_0}|\epsilon_Z(x)|\le u_Z$.
Consequently $\varepsilon_{Z,g}\le u_Z$ for any archive supported on these prompts, including an archive whose composition depends on the reference samples.

\paragraph{Proof.}
If $a=b$, both normalizers equal $a$, and the mean-square and uniform bounds hold with zero error.
Assume $a<b$.
Both $Z_{\mathrm{ref}}(x)$ and $\widehat Z_{\mathrm{ref}}(x)$ lie in $[a,b]$.
The mean-value theorem gives
\[
    |\epsilon_Z(x)|
    \le \frac{|\widehat Z_{\mathrm{ref}}(x)-Z_{\mathrm{ref}}(x)|}{a}.
\]
Taking squared expectations and using the variance of an independent sample mean proves Eq.~\eqref{eq:regime_normalizer_mse}.
For the probability bound, the bounded-sample inequality \citep{hoeffding1963probability} gives
\[
    \Pr\!\left(
        |\widehat Z_{\mathrm{ref}}(x)-Z_{\mathrm{ref}}(x)|>t
    \right)
    \le 2e^{-\frac{2Kt^2}{(b-a)^2}}.
\]
Set $t=h_Z$, so the per-prompt failure probability is at most $\delta_Z/m$.
A union bound gives
$|\widehat Z_{\mathrm{ref}}(x)-Z_{\mathrm{ref}}(x)|\le h_Z$
simultaneously for all $x\in\mathcal X_0$ with probability at least $1-\delta_Z$.
For any positive $u,v$, their logarithmic difference satisfies
\begin{equation}
    |\log u-\log v|
    =\log\!\left(1+\frac{|u-v|}{\min(u,v)}\right).
    \label{eq:normalizer_log_transfer}
\end{equation}
Applying this identity with $u=\widehat Z_{\mathrm{ref}}(x)$ and $v=Z_{\mathrm{ref}}(x)$, and using $\min(u,v)\ge a$, yields
\[
    |\epsilon_Z(x)|
    \le\log\!\left(1+\frac{h_Z}{a}\right)
    \qquad\text{for all }x\in\mathcal X_0.
\]
The deterministic inequality $|\epsilon_Z(x)|\le\log(b/a)$ supplies the other term in $u_Z$.
Since $\log(1+s)\le s$ for $s\ge0$, this logarithmic radius is no larger than
$\min\{\log(b/a),h_Z/a\}$ and is strictly smaller when $0<h_Z<b-a$.
On the simultaneous event, every empirical weighting of these prompts has root-mean-square error at most $u_Z$.
No independence across prompts or between the normalizer and replay archive is needed for this last implication.
For the rewards used here, one may take $a=1$ and $b=e^{1/\beta}$.

Equation~\eqref{eq:logz-bias-bound} bounds the signed expectation of the log-normalizer error at order $K^{-1}$.
Equation~\eqref{eq:regime_normalizer_mse} controls its root-mean-square magnitude at order $1/\sqrt{K}$.
The latter magnitude, rather than the signed expectation alone, is the target-error quantity entering Theorem 1.

\paragraph{Adaptive collection and policy-class complexity.}
Let $Z_i=(X_i,Y_i)$ denote the $i$th newly collected training record and $\mathcal F_i$ the collection history through that record.
The conditional law
$Q_i(\cdot)=\Pr(Z_i\in\cdot\mid\mathcal F_{i-1})$
may depend on preceding policy updates, selected prompts, and earlier samples.
For a fixed number $n$ of collected records, define
\begin{equation}
    \overline q_n=\frac1n\sum_{i=1}^n Q_i,
    \qquad
    \widehat q_n=\frac1n\sum_{i=1}^n\delta_{Z_i}.
    \label{eq:adaptive_collection_law}
\end{equation}
Here $\delta_z$ is the point mass at $z$.
$\overline q_n$ is generally random and describes the average conditional collection law.
For independent samples from a fixed distribution $q$, it reduces to $q$.
Any collection-time acceptance rule is included in $Q_i$.

Assume a fixed policy class with $|e_\pi(x,y)|\le B$ for every $\pi\in\Pi$ and every pair in the declared domain.
Let $\mathcal E=\{e_\pi:\pi\in\Pi\}$.
For $\eta>0$, suppose $\mathcal E$ has a finite deterministic sup-norm $\eta$-net with centers in $\mathcal E$ and size $\mathcal N_\infty(\eta,\mathcal E)$.
Thus every $e_\pi$ is within $\eta$ of one center at every pair in the domain.
These assumptions can be realized by a compact parameter set with continuous positive sequence probabilities on a finite prompt set and bounded-length completion space.

\paragraph{Proposition 3 (Uniform error for adaptive records).}
For $\delta_{\mathrm{sam}}\in(0,1)$, define
\begin{equation}
    v_n
    =
    B^2\sqrt{
        \frac{\log\!\left(2\mathcal N_\infty(\eta,\mathcal E)/
                    \delta_{\mathrm{sam}}\right)}{2n}
    }
    +4B\eta.
    \label{eq:regime_sampling_radius}
\end{equation}
With probability at least $1-\delta_{\mathrm{sam}}$,
\begin{equation}
    \sup_{\pi\in\Pi}
    \left|
        \|e_\pi\|_{\overline q_n}^2
        -\|e_\pi\|_{\widehat q_n}^2
    \right|
    \le v_n,\qquad
    \sup_{\pi\in\Pi}
    \left|
        \|e_\pi\|_{\overline q_n}
        -\|e_\pi\|_{\widehat q_n}
    \right|
    \le \sqrt{v_n}.
\label{eq:regime_sampling_uniform}
\end{equation}
For a finite policy class, one may instead take an exact net of size $|\Pi|$ and set $\eta=0$.

\paragraph{Proof.}
Fix a deterministic net center $e_j$ and write
\[
    U_{i,j}=e_j(Z_i)^2,\qquad
    D_{i,j}=U_{i,j}-\mathbb E[U_{i,j}\mid\mathcal F_{i-1}].
\]
Then $0\le U_{i,j}\le B^2$, and $D_{i,j}$ has conditional mean zero.
The conditional exponential-moment bound is
\[
    \mathbb E[e^{\lambda D_{i,j}}\mid\mathcal F_{i-1}]
    \le e^{\lambda^2 B^4/8}.
\]
To see this, for any conditionally bounded variable $U\in[0,B^2]$,
let $h(\lambda)=\log\mathbb E[e^{\lambda U}\mid\mathcal F_{i-1}]$.
Its second derivative is the variance of $U$ under the exponentially tilted conditional law.
That variance is at most $B^4/4$, since any variable in $[0,B^2]$ has variance at most one quarter of the squared range.
Integrating this bound twice from zero gives
$h(\lambda)-\lambda h'(0)\le\lambda^2 B^4/8$.

Iterated conditional expectation now yields
\[
    \mathbb E\!\left[e^{\lambda\sum_{i=1}^nD_{i,j}}\right]
    \le e^{n\lambda^2 B^4/8}.
\]
Markov's inequality, optimized at $\lambda=4t/B^4$, bounds the upper tail of
$n^{-1}\sum_iD_{i,j}$ by $e^{-2nt^2/B^4}$.
Applying the same argument to the lower tail gives
\[
    \Pr\!\left(
        \left|\frac1n\sum_{i=1}^nD_{i,j}\right|>t
    \right)
    \le 2e^{-2nt^2/B^4}.
\]
Moreover,
$n^{-1}\sum_iD_{i,j}
=\|e_j\|_{\widehat q_n}^2-\|e_j\|_{\overline q_n}^2$.
A union bound over the net gives the first term in $v_n$.

For any $e_\pi$ and a net center $e_j$ with
$\|e_\pi-e_j\|_\infty\le\eta$,
\[
    |e_\pi(x,y)^2-e_j(x,y)^2|
    \le |e_\pi-e_j|\,|e_\pi+e_j|
    \le 2B\eta.
\]
Replacing the center by $e_\pi$ in the empirical and conditional-law expectations therefore contributes at most $4B\eta$.
This proves the squared-risk bound.
Applying $|\sqrt u-\sqrt v|\le\sqrt{|u-v|}$ proves the root-mean-square bound.
For $B=0$, both bounds hold trivially.
The concentration is uniform over a fixed class, so the fitted policy may depend on all the collected records.

\paragraph{Corollary 1 (Finite-sample objective bound).}
Assume the conditions of Propositions 2 and 3, with all collected prompts in $\mathcal X_0$.
Take $q_g=\overline q_n$, $\widehat q_g=\widehat q_n$, and let $\widehat\pi_g\in\Pi$ be any fitted policy.
With probability at least $1-\delta_Z-\delta_{\mathrm{sam}}$, for every $C\ge1$,
\begin{equation}
    J_\rho(\pi^\star)-J_\rho(\widehat\pi_g)
    \le
    \beta\sqrt C\,
    \bigl(
        \varepsilon_{\mathrm{app},g}
        +2u_Z+2\sqrt{v_n}
        +\varepsilon_{\mathrm{opt},g}
    \bigr)
    +\beta B\,\kappa_g(C).
\label{eq:regime_finite_sample_bound}
\end{equation}

\paragraph{Proof.}
Proposition 2 bounds $\varepsilon_{Z,g}$ by the logarithmic radius $u_Z$ in Eq.~\eqref{eq:regime_normalizer_radius}, uniformly over archive prompts.
Proposition 3 bounds $\varepsilon_{\mathrm{sam},g}$ uniformly over $\Pi$.
The two events hold jointly with the stated probability by a union bound, irrespective of dependence between them.
Substitute both bounds into Theorem 1.
In particular, using reference responses both to estimate the normalizer and as training records does not require sample splitting in this argument.

\subsection{Source mixtures, fixed archives, and replay}
\label{app:regime_mixtures}

\paragraph{Regime distributions.}
At a chosen training checkpoint, partition eligible records into disjoint offline, reference, current-rollout, and historical-rollout groups.
Denote their empirical distributions by $\widehat q_s$ and their record counts by $n_s$.
For a regime with eligible source set $\mathcal I_g$, uniform record replay induces
\begin{equation}
    \widehat q_g
    =\sum_{s\in\mathcal I_g}\alpha_{g,s}\widehat q_s,
    \qquad
    \alpha_{g,s}
    =\frac{n_s}{\sum_{u\in\mathcal I_g}n_u}.
    \label{eq:regime_source_mixture}
\end{equation}
Zero-count groups are omitted.
For CBPO-Offline, $\mathcal I_g$ consists of offline and reference records.
CBPO-OnPolicy uses only the latest rollout group.
CBPO-OffPolicy uses reference, current, and historical rollout groups.
CBPO-Unified uses all four groups.
The reference samples used only for target construction in CBPO-OnPolicy are not part of its training mixture.

With corresponding population or conditional collection laws $q_s$, define
$q_g=\sum_s\alpha_{g,s}q_s$.
The target is unchanged across these mixtures, but their approximation errors, finite-record errors, and coverage quantities in Theorem 1 can differ.
For example, if one component satisfies $p_g\le C_s q_s$ and has weight $\alpha_{g,s}>0$, then
$q_g\ge\alpha_{g,s}q_s$ implies
$p_g\le(C_s/\alpha_{g,s})q_g$.
Thus a source can supply coverage, while its mixture weight controls the corresponding transfer factor.
More explicitly, suppose a reference component has population law
$q_{\mathrm{ref}}(x,y)=\rho(x)\pi_{\mathrm{ref}}(y\mid x)$ and mixture weight $\alpha_{g,\mathrm{ref}}>0$.
Under $a\le e^{r/\beta}\le b$ and $|e_{\widehat\pi_g}|\le B$,
\[
    \frac{\widehat\pi_g(y\mid x)}{\pi_{\mathrm{ref}}(y\mid x)}
    =
    e^{e_{\widehat\pi_g}(x,y)}
    \frac{e^{r(x,y)/\beta}}{Z_{\mathrm{ref}}(x)}
    \le e^B\frac ba.
\]
Hence $p_g\le Cq_g$ with
$C=e^B b/(a\alpha_{g,\mathrm{ref}})$ and $\kappa_g(C)=0$.
This statement concerns the population reference component, with its finite archive handled by the sampling term.
It applies to any regime retaining that component under the stated prompt-distribution condition.
The relative performance of regimes depends on the complete bound, including approximation and optimization.

\paragraph{Combining fresh and fixed data.}
Let $\mathcal I_{\mathrm{fresh}}\subseteq\mathcal I_g$ index the fresh-data groups.
Suppose each such group obeys the collection conditions of Proposition 3 at a fixed group size $n_s$.
Fix $\delta_{\mathrm{sam}}\in(0,1)$ and choose deterministic failure levels $\delta_s\in(0,1)$ before collecting the fresh records such that
\begin{equation}
    \sum_{s\in\mathcal I_{\mathrm{fresh}}}\delta_s
    \le\delta_{\mathrm{sam}}.
    \label{eq:regime_source_failure_budget}
\end{equation}
Let $v_s$ be the squared-risk bound for source $s$ at failure probability $\delta_s$.
For an offline archive treated as fixed, condition on that archive and set
$q_s=\widehat q_s$ for this component, so its conditional squared-risk discrepancy is zero.
Its finite support and reward composition still enter $\varepsilon_{\mathrm{app},g}$ and $\kappa_g(C)$.
Then, with probability at least $1-\sum_{s\in\mathcal I_{\mathrm{fresh}}}\delta_s$,
\begin{equation}
    \varepsilon_{\mathrm{sam},g}
    \le
    \sqrt{
        \sum_{s\in\mathcal I_{\mathrm{fresh}}}\alpha_{g,s}v_s
    }.
    \label{eq:regime_stratified_bound}
\end{equation}

\paragraph{Proof.}
On the simultaneous per-source events, every $\pi\in\Pi$ satisfies
\[
    \left|\|e_\pi\|_{q_g}^2-\|e_\pi\|_{\widehat q_g}^2\right|
    =
    \left|\sum_s\alpha_{g,s}
       \bigl(\|e_\pi\|_{q_s}^2-\|e_\pi\|_{\widehat q_s}^2\bigr)\right|
    \le\sum_{s\in\mathcal I_{\mathrm{fresh}}}\alpha_{g,s}v_s.
\]
The fixed-archive terms vanish by construction.
Take square roots using the inequality from Lemma 2 and then the supremum over $\Pi$.
A union bound gives simultaneous probability at least
$1-\sum_{s\in\mathcal I_{\mathrm{fresh}}}\delta_s\ge1-\delta_{\mathrm{sam}}$.
The fixed archive introduces no additional concentration event.
Substituting Eq.~\eqref{eq:regime_stratified_bound} for the sampling term in Theorem 1 gives the corresponding objective bound for a fixed offline archive combined with fresh reference and rollout records.
Under the uniform normalizer conditions of Corollary 1, combining this event with the normalizer event gives joint probability at least $1-\delta_Z-\delta_{\mathrm{sam}}$, without requiring independence.

\paragraph{Collection counts and optimizer reuse.}
The sample counts in these concentration bounds count collected records, not repeated accesses to a replay file.
A uniform minibatch drawn without replacement has expected loss equal to the archive mean because each record has the same inclusion probability.
Drawing another minibatch therefore optimizes the same empirical risk.
Repeated replay can reduce $\varepsilon_{\mathrm{opt},g}$, whereas collecting new responses changes $\widehat q_g$, the sampling term, and coverage.
The adaptive argument applies at a fixed collection horizon with the stated conditional laws.
For an archive formed by later data-dependent filtering, Theorem 1 still applies with its actual weights, and a probabilistic sampling bound must account for that selection rule.
This separates the effect of collection and retention from the effect of optimizing a fixed archive.

\subsection{Extension to truncated reference sampling}
\label{app:reference_decoding_bias}

\paragraph{Setting and error decomposition.}
Retain the original KL-regularized objective, reference log-probabilities, and optimal policy $\pi^\star$ from Eq.~\eqref{eq:cbpo_opt_policy}.
This extension considers reference collection with temperature $1.0$ and top-$p<1$.
Let $\nu_{\mathrm{ref}}(\cdot\mid x)$ denote the completion distribution induced by the frozen reference model and its decoding configuration.
Such truncation can induce a sampling distribution different from the original $\pi_{\mathrm{ref}}$ used for likelihood scoring.
For a fixed prompt, assume $0<a_x\le W(x,y)=e^{r(x,y)/\beta}\le b_x<\infty$ on both supports, and define
\begin{equation}
    Z_\nu(x)=\mathbb E_{Y\sim\nu_{\mathrm{ref}}(\cdot\mid x)}[W(x,Y)],\qquad
    d_{\mathrm{dec}}(x)=\log Z_\nu(x)-\log Z_{\mathrm{ref}}(x).
\label{eq:decoder_normalizer_definitions}
\end{equation}
In this extension, the frozen estimate is the same unweighted sample mean as in Eq.~\eqref{eq:cbpo_z_hat}, but its $K=K_{\mathrm{ref}}$ independent responses are drawn from $\nu_{\mathrm{ref}}$.
Define its finite-sample error around the decoded sampling law by
$\epsilon_{\mathrm{MC},K}(x)=\log\widehat Z_{\mathrm{ref},K}(x)-\log Z_\nu(x)$.
The total normalizer error from Eq.~\eqref{eq:regime_error_terms} then satisfies the exact identity
\begin{equation}
    \epsilon_Z(x)
    =\log\widehat Z_{\mathrm{ref},K}(x)-\log Z_{\mathrm{ref}}(x)
    =\epsilon_{\mathrm{MC},K}(x)+d_{\mathrm{dec}}(x).
\label{eq:decoder_error_split}
\end{equation}
The first term is finite-sample error, while the second is the difference between the normalizers of the decoded and original reference laws.

\paragraph{Bias and limit.}
Let $\sigma_{\nu,x}^2=\operatorname{Var}_{\nu_{\mathrm{ref}}}(W(x,Y))$.
Then
\begin{equation}
    d_{\mathrm{dec}}(x)-\frac{\sigma_{\nu,x}^2}{2Ka_x^2}
    \le \mathbb E[\epsilon_Z(x)]
    \le d_{\mathrm{dec}}(x).
    \label{eq:decoder_signed_bias}
\end{equation}
Moreover, $\epsilon_Z(x)\to d_{\mathrm{dec}}(x)$ almost surely and in expectation as $K\to\infty$.

\paragraph{Proof.}
The sample mean has expectation $Z_\nu(x)$ and variance $\sigma_{\nu,x}^2/K$.
Suppress $x$ and write $\widehat Z=\widehat Z_{\mathrm{ref},K}(x)$.
Taylor's theorem at $Z_\nu$ gives
\[
    \epsilon_{\mathrm{MC},K}
    =\frac{\widehat Z-Z_\nu}{Z_\nu}
      -\frac{(\widehat Z-Z_\nu)^2}{2\xi_K^2},
    \qquad \xi_K\ge a_x.
\]
The linear term has zero mean, and the expected remainder lies between
$-\sigma_{\nu,x}^2/(2Ka_x^2)$ and zero.
Adding $d_{\mathrm{dec}}$ proves Eq.~\eqref{eq:decoder_signed_bias}.
The strong law of large numbers and continuity of $\log$ on $[a_x,b_x]$ give the almost-sure limit.
Since $|\epsilon_Z|\le\log(b_x/a_x)$, bounded convergence gives the expectation limit.
Thus the signed total bias need not be negative, and increasing $K$ removes the finite-sample component.

\paragraph{A distributional bound on decoding bias.}
Define the completion-level total variation distance
\[
    \tau_x
    =\frac12\sum_y
        |\nu_{\mathrm{ref}}(y\mid x)-\pi_{\mathrm{ref}}(y\mid x)|.
\]
Then
\begin{equation}
    |d_{\mathrm{dec}}(x)|
    \le D_{\mathrm{dec}}(x)
    :=\log\!\left(1+\frac{b_x-a_x}{a_x}\tau_x\right)
    \le\log\frac{b_x}{a_x}.
    \label{eq:decoder_tv_bound}
\end{equation}

\paragraph{Proof.}
For $b_x>a_x$, write $W=a_x+(b_x-a_x)f$ with $0\le f\le1$.
The positive and negative parts of the signed measure
$\nu_{\mathrm{ref}}-\pi_{\mathrm{ref}}$ each have total mass $\tau_x$, hence
$|\mathbb E_{\nu_{\mathrm{ref}}}f-\mathbb E_{\pi_{\mathrm{ref}}}f|\le\tau_x$.
Therefore $|Z_\nu-Z_{\mathrm{ref}}|\le(b_x-a_x)\tau_x$.
For positive $u,v$,
\[
    |\log u-\log v|
    =\log\!\left(1+\frac{|u-v|}{\min(u,v)}\right).
\]
Apply this identity with $u=Z_\nu$, $v=Z_{\mathrm{ref}}$, and
$\min(Z_\nu,Z_{\mathrm{ref}})\ge a_x$.
The last inequality follows from $\tau_x\le1$.
If $a_x=b_x$, both normalizers are identical and the bound is zero.
Here $\tau_x$ compares complete-response distributions.
A token-level top-$p$ threshold does not set $\tau_x$ equal to $1-p$.

\paragraph{Mean-square and uniform bounds.}
The finite-sample and total errors satisfy
\begin{equation}
    \mathbb E[\epsilon_{\mathrm{MC},K}(x)^2]
        \le \frac{\sigma_{\nu,x}^2}{Ka_x^2},\qquad
    \sqrt{\mathbb E[\epsilon_Z(x)^2]}
        \le |d_{\mathrm{dec}}(x)|
            +\frac{\sigma_{\nu,x}}{a_x\sqrt K}.
\label{eq:decoder_normalizer_mse}
\end{equation}
For uniform bounds $0<a\le W\le b$ and a fixed set $\mathcal X_0$ of $m$ prompts, define $h_Z$ as in Eq.~\eqref{eq:regime_normalizer_radius} and set
$u_{\mathrm{MC}}=\min\{\log(b/a),\log(1+h_Z/a)\}$,
with $K$ independent decoded draws per prompt and failure probability $\delta_Z$.
With probability at least $1-\delta_Z$,
$\max_{x\in\mathcal X_0}|\epsilon_{\mathrm{MC},K}(x)|\le u_{\mathrm{MC}}$.
For any archive supported on these prompts, this gives
\begin{equation}
    \varepsilon_{\mathrm{dec},g}
        :=\|d_{\mathrm{dec}}\|_{\widehat q_g},\qquad
    \varepsilon_{Z,g}
        \le u_{\mathrm{MC}}+\varepsilon_{\mathrm{dec},g},\qquad
    \varepsilon_{\mathrm{dec},g}
        \le\|D_{\mathrm{dec}}\|_{\widehat q_g}.
\label{eq:decoder_target_bound}
\end{equation}
The deterministic cap $\varepsilon_{Z,g}\le\log(b/a)$ also holds.

\paragraph{Proof.}
The mean-value theorem on $[a_x,b_x]$ gives
$|\epsilon_{\mathrm{MC},K}(x)|\le|\widehat Z-Z_\nu|/a_x$.
Taking squared expectations proves the first inequality of Eq.~\eqref{eq:decoder_normalizer_mse}.
The second follows from Eq.~\eqref{eq:decoder_error_split} and the triangle inequality in $L^2$ over the reference draws.
For the uniform bound, $a=b$ makes both error components zero.
For $a<b$, independent decoded draws obey
\[
    \Pr\!\left(|\widehat Z_{\mathrm{ref},K}(x)-Z_\nu(x)|>t\right)
    \le2e^{-\frac{2Kt^2}{(b-a)^2}}.
\]
Set $t=h_Z$ and take a union bound with per-prompt failure probability $\delta_Z/m$.
Apply Eq.~\eqref{eq:normalizer_log_transfer} to
$\widehat Z_{\mathrm{ref},K}(x)$ and $Z_\nu(x)$, both bounded below by $a$, to obtain
\[
    |\epsilon_{\mathrm{MC},K}(x)|
    \le\log\!\left(1+\frac{h_Z}{a}\right)
    \qquad\text{for all }x\in\mathcal X_0.
\]
Both normalizers also lie in $[a,b]$, giving the deterministic cap and hence the stated $u_{\mathrm{MC}}$.
On the resulting simultaneous event, every empirical weighting has
$\|\epsilon_{\mathrm{MC},K}\|_{\widehat q_g}\le u_{\mathrm{MC}}$.
The triangle inequality and Eq.~\eqref{eq:decoder_tv_bound} prove Eq.~\eqref{eq:decoder_target_bound}.
Since $\widehat Z_{\mathrm{ref},K}$ and $Z_{\mathrm{ref}}$ also lie in $[a,b]$, the cap applies to the total error.
These implications require no independence across prompts or between the reference estimates and the archive.
Thus the $K^{-1}$ log-bias and $1/\sqrt{K}$ root-mean-square rates apply to the finite-sample component around $Z_\nu$, while $d_{\mathrm{dec}}$ is separate.

\paragraph{Gradient effect at fixed parameters.}
Fix $\theta$ and a training-record law $q_t(\cdot\mid x)$ independent of the reference draws, with the integrability and differentiation conditions used in the preceding exact-reference analysis.
Retain $\Delta_\theta$, $A_\theta$, and $M_x(\theta)$ from that analysis.
Differentiating the squared residual and using Eq.~\eqref{eq:decoder_error_split} gives
\[
    \mathbb E[\nabla_\theta\widehat{\mathcal L}_{x,K}]
        -\nabla_\theta\mathcal L_x
    =2\bigl(d_{\mathrm{dec}}+\mathbb E[\epsilon_{\mathrm{MC},K}]\bigr)
        \mathbb E_{q_t}[A_\theta].
\]
The factorization follows from independence.
Since $\|\mathbb E_{q_t}A_\theta\|\le M_x(\theta)$, the Taylor bound yields
\begin{equation}
    \left\|\mathbb E[\nabla_\theta\widehat{\mathcal L}_{x,K}]
        -\nabla_\theta\mathcal L_x\right\|
    \le 2M_x(\theta)|d_{\mathrm{dec}}|
        +\frac{M_x(\theta)\sigma_{\nu,x}^2}{Ka_x^2}.
\label{eq:decoder_gradient_bias}
\end{equation}

For reused reference responses $Y_1,\ldots,Y_K\sim\nu_{\mathrm{ref}}$, let
$A_i=A_\theta(x,Y_i)$, $W_i=W(x,Y_i)$, and assume
$\|A_i\|\le\overline M_x$.
Define
\[
\resizebox{\linewidth}{!}{$\displaystyle
    g_\nu=2\mathbb E_{\nu_{\mathrm{ref}}}[\Delta_\theta A_\theta],\quad
    \widehat g_{\nu,K}
        =\frac2K\sum_{i=1}^K
            \bigl(\Delta_\theta(x,Y_i)+\epsilon_Z(x)\bigr)A_i,\quad
    B_{\nu,x}
        =\mathbb E_{\nu_{\mathrm{ref}}}
            [\|A_\theta\|\,|W-Z_\nu|].
$}
\]
Exchangeability gives
\[
    \mathbb E[\widehat g_{\nu,K}]-g_\nu
    =2d_{\mathrm{dec}}\mathbb E_{\nu_{\mathrm{ref}}}[A_\theta]
      +2\mathbb E[\epsilon_{\mathrm{MC},K}A_1].
\]
In the Taylor expansion above, the linear terms satisfy
$\mathbb E[(W_i-Z_\nu)A_1]=0$ for $i\ne1$.
The $i=1$ contribution to this gradient difference has norm at most $2B_{\nu,x}/(KZ_\nu)$.
The quadratic remainder is bounded using $\overline M_x$ and
$\mathbb E[(\widehat Z-Z_\nu)^2]=\sigma_{\nu,x}^2/K$.
Therefore
\begin{equation}
    \left\|\mathbb E[\widehat g_{\nu,K}]-g_\nu
        -2d_{\mathrm{dec}}\mathbb E_{\nu_{\mathrm{ref}}}[A_\theta]\right\|
    \le
        \frac{2B_{\nu,x}}{KZ_\nu(x)}
        +\frac{\overline M_x\sigma_{\nu,x}^2}{Ka_x^2}.
\label{eq:decoder_reuse_gradient}
\end{equation}
For $K>1$, a leave-one-out normalizer removes the reuse covariance at fixed $\theta$, while its expectation still contains $d_{\mathrm{dec}}$.

\paragraph{Corollary 2 (Objective gap with reference decoding).}
Assume the uniform normalizer conditions above and the conditions of Lemma 1 and Proposition 3.
Use the actual conditional collection laws, with
$q_g=\overline q_n$ and $\widehat q_g=\widehat q_n$, and suppose all archive prompts lie in $\mathcal X_0$.
For any fitted $\widehat\pi_g\in\Pi$, with probability at least
$1-\delta_Z-\delta_{\mathrm{sam}}$, for every $C\ge1$,
\begin{equation}
    J_\rho(\pi^\star)-J_\rho(\widehat\pi_g)
    \le\beta\sqrt C\,
        \bigl(\varepsilon_{\mathrm{app},g}
              +2u_{\mathrm{MC}}+2\varepsilon_{\mathrm{dec},g}
         +2\sqrt{v_n}+\varepsilon_{\mathrm{opt},g}\bigr)
        +\beta B\,\kappa_g(C).
\label{eq:decoder_objective_bound}
\end{equation}

\paragraph{Proof.}
Theorem 1 is a pathwise bound for any frozen normalizer error and the stated collection law.
On the simultaneous normalizer event,
Eq.~\eqref{eq:decoder_target_bound} bounds
$\varepsilon_{Z,g}$ by $u_{\mathrm{MC}}+\varepsilon_{\mathrm{dec},g}$.
Proposition 3 bounds $\varepsilon_{\mathrm{sam},g}$ by $\sqrt{v_n}$.
A union bound combines these events without requiring independence.
Substitution in Theorem 1 gives Eq.~\eqref{eq:decoder_objective_bound}.
The additional decoding contribution is at most
$2\beta\sqrt C\,\|D_{\mathrm{dec}}\|_{\widehat q_g}$.
For source mixtures and fixed offline archives satisfying the conditions in Appendix~\ref{app:regime_mixtures}, replace $\sqrt{v_n}$ in Eq.~\eqref{eq:decoder_objective_bound} by
\[
    \sqrt{
        \sum_{s\in\mathcal I_{\mathrm{fresh}}}\alpha_{g,s}v_s
    }.
\]
With the failure levels allocated as in Eq.~\eqref{eq:regime_source_failure_budget}, a union bound gives joint probability at least
\[
    1-\delta_Z-\sum_{s\in\mathcal I_{\mathrm{fresh}}}\delta_s
    \ge 1-\delta_Z-\delta_{\mathrm{sam}}.
\]

When the reference prompts have marginal $\rho$, the decoded reference component has law
$q_{\mathrm{ref}}(x,y)=\rho(x)\nu_{\mathrm{ref}}(y\mid x)$.
Accordingly, the transfer factor and uncovered mass are evaluated for this actual component and the other eligible sources.
The exact-reference coverage example in Appendix~\ref{app:regime_mixtures} uses $\rho\pi_{\mathrm{ref}}$.
For the decoded mixture, any zero-tail coverage claim requires the corresponding domination condition on its actual law.
Decoding therefore enters the target through $d_{\mathrm{dec}}$ and can also affect coverage through $q_g$.
Setting $\nu_{\mathrm{ref}}=\pi_{\mathrm{ref}}$ makes $d_{\mathrm{dec}}=0$ and recovers the ideal reference-sampling bounds.

\section{Additional Experimental Setup}
\label{app:prompt_reward}

\begingroup
\setlength{\emergencystretch}{2em}
\paragraph{Artifact sources and offline-record preparation.}
RLVR-Math-16k draws from hiyouga/math12k and NuminaMath-1.5-RL-Verifiable \citep{hiyouga_math12k,numinamath15_rlverifiable}.
The fixed DPSK-CoT trajectories were generated by DeepSeek-R1-Distill-\allowbreak{}Qwen-32B \citep{deepseekai2025r1}.
The teacher generates 32 candidates per prompt, and every retained prompt has at least three reward-$1$ candidates.
The same maximum response length of 2,048 tokens was used for both DPSK-CoT trajectory generation and responses generated during training.
The held-out evaluation artifacts named in the main paper follow the cited releases \citep{lightman2023letsverify,mathai_math500,lewkowycz2022solving,mathai_minervamath,mathai_amc23,mathai_aime25}.
Offline trajectories follow the preprocessing in Sec.~\ref{sec:cbpo_unified}.
Each record retains its source and collection round for replay-age and source-composition diagnostics, but these fields do not enter the CBPO loss.
\par
\endgroup

\begingroup
\setlength{\emergencystretch}{2em}
\paragraph{Prompt construction.}
The RLVR-Math-16k rows store a common system message followed by a row-specific user problem.
The collection and evaluation code renders this message list with the Qwen tokenizer chat template using \texttt{add\_\allowbreak generation\_\allowbreak prompt=True} and passes \texttt{enable\_\allowbreak thinking=True} when supported by the tokenizer.
Fig.~\ref{fig:app_prompt_system} shows the full two-message prompt delivered to the model.
\par
\endgroup

\raggedbottom
\begin{figure*}[t]
\begin{tcolorbox}[enhanced jigsaw, title=Delivered prompt, colback=white, colframe=black!35, boxrule=0.4pt, arc=1mm, left=1mm, right=1mm, top=1mm, bottom=1mm]
\footnotesize
\textbf{Message 1 -- system}\\
You are a math problem solver. For each problem, think through it step by step within \texttt{\textless think\textgreater} \texttt{\textless/think\textgreater} tags, then provide your final answer using \texttt{\textbackslash boxed\{\}}.\\[3pt]
Requirements:\\
\hspace*{1em}$\bullet$~Show your complete reasoning process inside \texttt{\textless think\textgreater} tags.\\
\hspace*{1em}$\bullet$~Provide your final answer inside \texttt{\textbackslash boxed\{\}}.\\[3pt]
Example:\\
\hspace*{1em}User: If $3x + 7 = 22$, what is $x$?\\
\hspace*{1em}Assistant: \texttt{\textless think\textgreater}\ $3x + 7 = 22$;\ \ $3x = 22 - 7 = 15$;\ \ $x = 15 / 3 = 5$\ \texttt{\textless/think\textgreater}\ The answer is \texttt{\textbackslash boxed\{5\}}.\\[3pt]
- - -\\[3pt]
\textbf{Message 2 -- user (row-specific math problem)}\\
\hspace*{1em}What is the least positive integer greater than 1 that leaves a remainder of 1 when divided by each of 2, 3, 4, 5, 6, 7, 8 and 9?\\[3pt]
\hspace*{1em}\textit{(At runtime the row-specific problem is substituted here. See Table~\ref{tab:prompt_instances} for representative instances.)}
\end{tcolorbox}
\caption{Two-message chat prompt with the system instruction and a row-specific math problem.}
\label{fig:app_prompt_system}
\end{figure*}

\paragraph{Concrete prompt instances.}
Table~\ref{tab:prompt_instances} shows representative online-collection user messages and verifier targets from the RLVR-Math-16k training split.
Math12k rows come from \texttt{hiyouga/math12k}.
AMC/AIME and Olympiads rows come from \texttt{nlile/NuminaMath-1.5-RL-Verifiable}.
At runtime, each user message is appended after the system message above, then rendered through the tokenizer chat template before generation.

\begin{table}[!htbp]
  \centering
  \small
  \setlength{\tabcolsep}{4pt}
  \renewcommand{\arraystretch}{1.18}
  \vspace{-5pt}
  \caption{Representative RLVR-Math-16k prompts with source provenance and verifier targets.}
  \label{tab:prompt_instances}
  \begin{tabular*}{\linewidth}{@{\extracolsep{\fill}}>{\raggedright\arraybackslash}p{0.12\columnwidth}>{\raggedright\arraybackslash}p{0.70\columnwidth}>{\raggedright\arraybackslash}p{0.10\columnwidth}@{}}
    \toprule
    \textbf{Problem provenance} & \textbf{User message} & \textbf{Ground truth} \\
    \midrule
    Math12k & What is the least positive integer greater than 1 that leaves a remainder of 1 when divided by each of 2, 3, 4, 5, 6, 7, 8 and 9? & \texttt{2521} \\
    Math12k & Find the remainder when $6x^4-14x^3-4x^2+2x-26$ is divided by $2x - 6.$ & \texttt{52} \\
    AMC/AIME & The workers in a factory produce widgets and whoosits. For each product, production time is constant and identical for all workers, but not necessarily equal for the two products. In one hour, $100$ workers can produce $300$ widgets and $200$ whoosits. In two hours, $60$ workers can produce $240$ widgets and $300$ whoosits. In three hours, $50$ workers can produce $150$ widgets and $m$ whoosits. Find $m$. & \texttt{450} \\
    AMC/AIME & How many whole numbers between 1 and 1000 do not contain the digit 1? (A) 512, (B) 648, (C) 720, (D) 728, or (E) 800. & \texttt{728} \\
    Olympiads & It is known that the difference between the variable quantities $z$ and $y$ is proportional to the quantity $x$, and the difference between the quantities $x$ and $z$ is proportional to the quantity $y$. The coefficient of proportionality is the same and is a positive integer $k$. A certain value of the quantity $z$ is $5 / 3$ times the difference of the corresponding values of $x$ and $y$. Find the numerical value of the coefficient $k$. & \texttt{3} \\
    Olympiads & Mr. and Mrs. Zeta want to name baby Zeta so that the first letters of the baby's first, middle, and last names are in alphabetical order and the letters are distinct. The number of possible names is (A) 276, (B) 300, (C) 552, (D) 600, or (E) 15600. & \texttt{300} \\
    \bottomrule
  \end{tabular*}
  \vspace{-10pt}
\end{table}

\paragraph{Reward implementation.}
The intermediate $0.1$ reward summarized in the main paper is a format component activated when a final \texttt{\textbackslash boxed\{...\}} answer can be extracted.
Verifier-confirmed equivalence to the normalized ground truth contributes the remaining $0.9$.
The format component is used only to encourage a parseable final answer.

\paragraph{Baseline and reproducibility details.}
\label{app:baseline_details}
The baseline groups and shared training controls follow Sec.~\ref{sec:experimental_setup}.
SFT uses the same positive offline trajectories as the corresponding CBPO configuration.
For the off-policy baseline, TBRM uses $n=4$ \citep{yuan2025tbrm}.
The implementations of DPO, PPO, DAPO, REINFORCE++, GSPO, and BRIDGE use verl 0.7.1.
The shared actor learning rate, decoding settings, hardware, and training seeds are specified in Table~\ref{tab:rollout_only_parameters}.

\noindent\textbf{Offline DPO.}
The initial and reference policies, prompt set, optimizer-step budget, and evaluation protocol are matched to CBPO-Offline.
We use the standard sigmoid DPO loss with coefficient $\beta_{\mathrm{DPO}}=0.1$, full-parameter optimization, and a frozen initial reference policy.
Sequence log probabilities sum over response tokens, including the end-of-sequence token, with prompt tokens excluded.
Pairing cycles through the smaller group until every response in the larger group has been paired.
Prompts without both labels produce no preference pairs.
Pairs that exceed the prompt or response length limit are excluded as complete pairs.
Each update uses 256 preference pairs.
AdamW uses weight decay $0.1$, a 10-step learning-rate warmup, and a gradient-norm limit of $1.0$.
Record reuse and token exposure are recorded separately from the optimizer-step budget.

\noindent\textbf{On-policy baselines.}
PPO uses generalized advantage estimation with $\lambda=0.999$ and a critic learning rate of $10^{-5}$.
DAPO uses lower and upper clipping widths of $0.2$ and $0.28$, a dual-clipping coefficient of $10$, and token-mean loss aggregation.
Its soft overlength penalty uses a 1,024-token buffer and penalty factor $1.0$.
REINFORCE++ uses discounted returns with $\gamma=1$ and batch-level normalization over valid response tokens, without a learned critic.
Its token-level policy ratios use a symmetric clipping width of $0.2$.
GSPO uses sequence-level importance ratios with lower and upper clipping widths of $3\times10^{-4}$ and $4\times10^{-4}$.
Advantages are standardized within each prompt group, and the loss averages over tokens within a response and then over responses.
These four baselines disable both reward-side and actor-loss KL penalties.
Their actor optimizers use weight decay $0.1$ and a 10-step learning-rate warmup.

\noindent\textbf{Cooperative SFT/RL.}
BRIDGE \citep{chen2026bridge} shares the initial student, training prompts, positive offline trajectories, current-policy generation budget, and student optimizer-step budget with CBPO-Unified at each model size.
Our implementation follows its joint student, auxiliary RL model, and shared LoRA-teacher updates.
The student and auxiliary model reuse the same student-generated rollouts.
The supervised/RL mixing weight is $0.5$.
The student and auxiliary learning rates are $10^{-6}$, and the LoRA-teacher learning rate is $5\times10^{-5}$.
LoRA uses rank 16 and scaling parameter 16 on the query, key, value, output, gate, up, and down projections.
The RL loss uses symmetric clipping width $0.2$, KL coefficient $10^{-4}$, and entropy coefficient $10^{-5}$.
The supervised loss is averaged over response tokens.
AdamW uses zero weight decay and no warmup, with a gradient-norm limit of $1.0$.
CBPO-Unified additionally uses negative offline trajectories, reference-response training, and historical replay.
The Qwen3-1.7B CBPO w/o History (PosCoT) configuration provides a comparison under matched data-source restrictions.
Cost accounting includes reference preprocessing and scoring, rollout generation, and all parameter updates, including BRIDGE's auxiliary RL model and LoRA updates.
Table~\ref{tab:training_cost} reports the per-run GPU hours for both model sizes.

\begin{table}[!htbp]
  \centering
  \small
  \setlength{\tabcolsep}{8pt}
  \caption{Per-run training cost in GPU hours on two NVIDIA B200 GPUs.}
  \label{tab:training_cost}
  \begin{tabular*}{\linewidth}{@{\extracolsep{\fill}}llcc@{}}
    \toprule
    \textbf{Regime} & \textbf{Methods} & \textbf{Qwen3-1.7B} & \textbf{Qwen3-8B} \\
    \midrule
    \multirow{3}{*}{Offline} & SFT & 102.23 & 413.92 \\
     & DPO & 106.69 & 430.28 \\
     & \textbf{CBPO-Offline} & 105.94 & 421.76 \\
    \midrule
    \multirow{5}{*}{On-policy} & PPO & 232.84 & 901.36 \\
     & DAPO & 118.43 & 473.72 \\
     & GSPO & 118.65 & 475.28 \\
     & REINFORCE++ & 119.73 & 478.92 \\
     & \textbf{CBPO-OnPolicy} & 113.20 & 432.80 \\
    \midrule
    \multirow{2}{*}{Off-policy} & TBRM & 117.05 & 448.20 \\
     & \textbf{CBPO-OffPolicy} & 117.51 & 466.04 \\
    \midrule
    Joint SFT/RL & BRIDGE & 143.24 & 514.16 \\
    \midrule
    Unified & \textbf{CBPO-Unified} & 119.36 & 437.44 \\
    \bottomrule
  \end{tabular*}
\end{table}

\subsection{Training--Evaluation Decontamination}
\label{app:decontamination}

Both the online prompt pool and the question text associated with offline trajectories were screened against all four evaluation benchmarks.
Confirmed overlapping training questions were excluded before reference sampling and current-policy rollout collection.
The main results in Tables~\ref{tab:core_cross_regime}, \ref{tab:core_cross_regime_8b}, and~\ref{tab:one_vs_two_stage} use the resulting training data.

\paragraph{Data scope.}
The decontamination checks compare training questions before filtering with questions from the four evaluation benchmarks.
The unfiltered training prompt pool contains 16,268 distinct questions.
The evaluation set contains 842 distinct questions, with 30 from AIME25, 272 from MinervaMath, 40 from AMC23, and 500 from MATH500.
Screening uses question text and excludes shared system instructions and examples.

\paragraph{Matching rules.}
Exact matching compares the extracted text after stripping only leading and trailing whitespace.
Mathematical-expression normalization applies Unicode NFKC normalization, removes invisible characters and whitespace, and standardizes equivalent LaTeX fraction commands, display delimiters, style wrappers, and inequality notation.
Numerical values, variable case, and operator distinctions are retained.
This step normalizes notation rather than testing arbitrary algebraic equivalence.
Approximate matching uses character 3--5-gram TF-IDF over normalized question texts, with case preserved, sublinear term frequency, and a vocabulary capped at 250,000 features.
Every evaluation question is compared with every training-question text, and pairs with cosine similarity at least 0.80 enter the candidate set.

Semantic retrieval uses the Sentence-BERT framework \citep{reimers2019sentencebert} with the all-MiniLM-L6-v2 encoder to map question text to 384-dimensional embeddings.
Long questions are split into chunks of at most 254 content tokens without truncating the remaining text.
Attention-masked mean pooling and L2 normalization produce chunk embeddings, whose content-length-weighted average is normalized to obtain the question embedding.
The semantic candidate threshold is a cosine similarity of 0.85.
Both similarity thresholds are screening rules, with candidate equivalence assessed from the mathematical conditions and requested quantity.
Codex-assisted content comparison distinguishes the same problem from a changed answer request, a shared setting with a different subquestion, and a different problem.

\paragraph{Screening results.}
Table~\ref{tab:decontamination_counts} reports candidate pairs and confirmed same-problem matches.
Each pair consists of one evaluation question and one distinct training-question text.
The four checks produce 151 distinct candidate pairs involving 97 evaluation questions and 149 training-question texts.
Content comparison identifies 14 same-problem pairs involving 13 evaluation questions and 14 training-question texts.
One evaluation question matches two differently worded training versions.

\begin{table}[!htbp]
  \centering
  \small
  \setlength{\tabcolsep}{4pt}
  \caption{Training--evaluation overlap screening before decontamination. Counts are deduplicated within each row. Checks can retrieve the same pair or question, so their counts are not additive.}
  \label{tab:decontamination_counts}
  \begin{tabular*}{\linewidth}{@{\extracolsep{\fill}}lcccc@{}}
    \toprule
    \textbf{Check} & \shortstack{\textbf{Candidate}\\\textbf{pairs}} & \shortstack{\textbf{Candidate}\\\textbf{test questions}} & \shortstack{\textbf{Same-problem}\\\textbf{pairs}} & \shortstack{\textbf{Matched}\\\textbf{test questions}} \\
    \midrule
    Exact-string matching & 0 & 0 & 0 & 0 \\
    Mathematical-expression normalization & 1 & 1 & 1 & 1 \\
    Approximate-text matching & 90 & 52 & 6 & 5 \\
    Semantic-similarity retrieval & 106 & 77 & 13 & 12 \\
    \midrule
    Deduplicated union & 151 & 97 & 14 & 13 \\
    \bottomrule
  \end{tabular*}
\end{table}

Two additional pairs use the same problem instance but change the requested answer or precision, involving two evaluation questions and two training-question texts.
A further 15 pairs share concrete data or a diagram but ask different subquestions, involving eight evaluation questions and 15 training-question texts.
The remaining 120 candidate pairs are different problems.
Table~\ref{tab:decontamination_benchmarks} separates these categories by evaluation group.
All candidates belong to the MATH500 group, with no candidates retrieved for the other three groups under the stated rules and thresholds.

\begin{table}[!htbp]
  \centering
  \small
  \setlength{\tabcolsep}{4pt}
  \caption{Evaluation questions matched to the unfiltered training prompt pool. Each category counts distinct evaluation questions. A question may occur in more than one category.}
  \label{tab:decontamination_benchmarks}
  \begin{tabular*}{\linewidth}{@{\extracolsep{\fill}}lcccc@{}}
    \toprule
    \textbf{Benchmark} & \shortstack{\textbf{Evaluation}\\\textbf{questions}} & \shortstack{\textbf{Same}\\\textbf{problem}} & \shortstack{\textbf{Changed}\\\textbf{answer request}} & \shortstack{\textbf{Shared setting,}\\\textbf{different question}} \\
    \midrule
    AIME25 & 30 & 0 & 0 & 0 \\
    MinervaMath & 272 & 0 & 0 & 0 \\
    AMC23 & 40 & 0 & 0 & 0 \\
    MATH500 & 500 & 13 & 2 & 8 \\
    \bottomrule
  \end{tabular*}
\end{table}

\paragraph{Data versions and retained counts.}
The 14 same-problem training questions are excluded, leaving 16,254 questions in the training prompt pool.
The two changed-answer cases and the shared-setting cases are reported separately from this removal count.
The complete evaluation set remains at 842 questions.
Table~\ref{tab:decontamination_sizes} summarizes the numbers of training and evaluation questions before and after filtering.
The screening results include full candidate texts, comparison decisions, original prompt identifiers, and file line numbers.

\begin{table}[!htbp]
  \centering
  \small
  \setlength{\tabcolsep}{6pt}
  \caption{Training and evaluation question counts before and after excluding same-problem training questions.}
  \label{tab:decontamination_sizes}
  \begin{tabular*}{\linewidth}{@{\extracolsep{\fill}}lccc@{}}
    \toprule
    \textbf{Unit} & \textbf{Before filtering} & \textbf{Removed} & \textbf{Retained} \\
    \midrule
    Training-question texts & 16,268 & 14 & 16,254 \\
    Evaluation questions & 842 & 0 & 842 \\
    \bottomrule
  \end{tabular*}
\end{table}

\section{Additional Domain Evaluations}
\label{app:additional_domain_evaluations}

\subsection{General-Domain Post-Training with a Learned Reward Model}
\label{app:general_domain_posttraining}

We evaluate general-domain post-training with a learned reward model. PPO and CBPO-Unified are initialized from Qwen3-1.7B~\citep{yang2025qwen3} and trained on the General subset of HelpSteer3-Preference~\citep{wang2025helpsteer3}. Both methods use Qwen3-Nemotron-32B-GenRM-Principle~\citep{nvidia2025genrm} with the same helpfulness principle. The difference between its Yes and No scores is mapped to a reward in $[0,1]$. We evaluate the initial checkpoint and both trained policies on Arena-Hard v2~\citep{li2024arenahard,lmarena2025arenahardv2} and MT-Bench~\citep{zheng2023mtbench}, using GPT-6-Luna-Max as the judge. CBPO-Unified improves over PPO by 0.6 percentage points in Arena-Hard win rate and 0.56 points in MT-Bench mean score.

\begin{table}[H]
  
  \centering
  \small
  \setlength{\tabcolsep}{7pt}
  \caption{General-domain post-training with a learned reward model.}
  \label{tab:general_domain_posttraining}
  \begin{tabularx}{0.9\linewidth}{@{}l*{2}{>{\centering\arraybackslash}X}@{}}
    \toprule
    \textbf{Model} & \shortstack{Arena-Hard v2\\win rate (\%)} & \shortstack{MT-Bench mean\\judge score (1--10)} \\
    \midrule
    Qwen3-1.7B & 2.8 & 4.66 \\
    PPO & 7.5 & 6.35 \\
    CBPO-Unified & 8.1 & 6.91 \\
    \bottomrule
  \end{tabularx}
\end{table}

\subsection{Cross-Domain Evaluation after Math-Focused Training}
\label{app:cross_domain_evaluation}

To assess performance beyond the domains targeted by math-focused post-training, we compare each base checkpoint with its corresponding CBPO-Unified checkpoint on held-out coding, instruction-following, and scientific question-answering evaluations. We report LiveCodeBench v5 pass@1~\citep{jain2024livecodebench}, IFEval prompt-level strict accuracy~\citep{zhou2023ifeval}, and GPQA-Diamond accuracy~\citep{rein2023gpqa}. Across the six model-size and benchmark comparisons, scores change by $-0.51$ to $0.00$ percentage points. Five comparisons show a decrease and one is unchanged.

\begin{table}[H]
  
  \centering
  \small
  \setlength{\tabcolsep}{3pt}
  \caption{Cross-domain evaluation before and after math-focused post-training. Scores are percentages.}
  \label{tab:cross_domain_evaluation}
  \begin{tabularx}{\linewidth}{@{}>{\centering\arraybackslash}p{0.82in}>{\centering\arraybackslash}p{1.10in}*{3}{>{\centering\arraybackslash}X}@{}}
    \toprule
    \textbf{Model size} & \textbf{Checkpoint} & \shortstack{LiveCodeBench v5\\pass@1} & \shortstack{IFEval\\prompt-level strict\\accuracy} & \shortstack{GPQA-Diamond\\accuracy} \\
    \midrule
    1.7B & Base Policy & 15.02 & 34.75 & 5.56 \\
    1.7B & CBPO-Unified & 14.85 & 34.38 & 5.05 \\
    8B & Base Policy & 33.20 & 45.10 & 12.12 \\
    8B & CBPO-Unified & 33.16 & 44.92 & 12.12 \\
    \bottomrule
  \end{tabularx}
\end{table}

\subsection{Single-Run Comparisons on Mathematical Benchmarks}
\label{app:single_run_math_comparisons}

Table~\ref{tab:single_run_math_comparisons} reports additional single-run avg@32/pass@32 results on AIME25, MinervaMath, AMC23, and MATH500 using the same metric layout as Table~\ref{tab:core_cross_regime}. 

The Qwen3-1.7B and Qwen3-8B checkpoint rows use the same evaluation protocol and generation budget. The A$^\star$-PO comparison uses Qwen/Qwen3-1.7B initialization and a shared evaluation protocol. A$^\star$-PO uses $\beta_1=0.5$ for value estimation, $\beta_2=10^{-3}$ for policy regression, and $N=8$ reference responses per prompt. AGRO and SPO start from Qwen/Qwen3-1.7B and use the same data-source pool and training budget as the CBPO-Unified comparison, with the same evaluation protocol. CBPO-Offline (PosCoT) uses only the positive demonstrations used by SFT, with the same Qwen3-1.7B initialization and data-source pool.

\begin{table}[H]
  
  \centering
  \scriptsize
  \setlength{\tabcolsep}{0.25pt}
  \renewcommand{\arraystretch}{1.0}
  \caption{Additional single-run results on four mathematical benchmarks. Avg. is the unweighted mean across the four benchmarks.}
  \label{tab:single_run_math_comparisons}
  \begin{tabularx}{\linewidth}{@{}>{\fontsize{6}{7}\selectfont}l>{\fontsize{6}{7}\selectfont}l *{10}{>{\centering\arraybackslash}X}@{}}
    \toprule
    \multirow[c]{2}{*}{\raisebox{-2pt}{\textbf{Regime}}} & \multirow[c]{2}{*}{\raisebox{-2pt}{\textbf{Methods}}}
      & \multicolumn{2}{c}{\textbf{AIME25}}
      & \multicolumn{2}{c}{\textbf{MinervaMath}}
      & \multicolumn{2}{c}{\textbf{AMC23}}
      & \multicolumn{2}{c}{\textbf{MATH500}}
      & \multicolumn{2}{c}{\textbf{Avg.}} \\
    \cmidrule(lr){3-4}\cmidrule(lr){5-6}\cmidrule(lr){7-8}\cmidrule(lr){9-10}\cmidrule(lr){11-12}
      & & \metrichead{avg@32} & \metrichead{pass@32} & \metrichead{avg@32} & \metrichead{pass@32} & \metrichead{avg@32} & \metrichead{pass@32} & \metrichead{avg@32} & \metrichead{pass@32} & \metrichead{avg@32} & \metrichead{pass@32} \\
    \midrule
    \multirow{2}{*}{Base} & Qwen3-1.7B & 1.88 & 10.00 & 18.53 & 31.62 & 27.66 & 37.50 & 53.09 & 70.80 & 25.29 & 37.48 \\
     & Qwen3-8B & 10.21 & 13.33 & 29.15 & 42.28 & 63.91 & 67.50 & 68.68 & 78.20 & 42.99 & 50.33 \\
    \midrule
    \multirow{2}{*}{Offline} & SFT & 14.17 & 26.67 & 19.27 & 33.46 & 48.83 & 77.50 & 64.99 & 78.00 & 36.81 & 53.91 \\
     & CBPO-Offline (PosCoT) & 14.38 & 26.67 & 19.30 & 33.46 & 48.91 & 75.00 & 65.22 & 78.00 & 36.95 & 53.28 \\
    \midrule
    \multirow{2}{*}{On-policy} & A$^\star$-PO & 17.08 & 33.33 & 21.54 & 39.34 & 56.33 & 85.00 & 70.41 & 84.60 & 41.34 & 60.57 \\
     & CBPO-OnPolicy & 17.40 & 36.67 & 22.00 & 39.71 & 57.19 & 87.50 & 70.71 & 85.00 & 41.82 & 62.22 \\
    \midrule
    \multirow{3}{*}{Unified} & AGRO & 19.17 & 43.33 & 23.56 & 41.54 & 58.20 & 90.00 & 71.89 & 86.00 & 43.21 & 65.22 \\
     & SPO & 19.27 & 43.33 & 23.76 & 41.91 & 58.44 & 90.00 & 72.32 & 86.20 & 43.45 & 65.36 \\
     & CBPO-Unified & 20.63 & 46.67 & 24.48 & 42.65 & 60.16 & 92.50 & 73.01 & 87.80 & 44.57 & 67.40 \\
    \bottomrule
  \end{tabularx}
\end{table}

\subsection{Extended-Generation Evaluation on Mathematical Benchmarks}
\label{app:extended_generation_evaluation}

We evaluate the original Qwen3-1.7B and Qwen3-8B checkpoints and their CBPO-Unified counterparts in thinking mode with evaluation output limits of 8,192 and 16,384 tokens. For each model and limit, the prompts, decoding settings, stopping criteria, answer extraction, and verification are held fixed, with 32 responses sampled per problem. Only the evaluation output limit changes; training remains unchanged, including its 2,048-token response limit. The five benchmarks contain 872 problems in total, giving 27,904 responses per model and output limit.

Tables~\ref{tab:extended_generation_8192} and~\ref{tab:extended_generation_16384} report standard avg@32/pass@32 over all sampled responses. Each benchmark score is shown as avg@32 and pass@32, and Avg. is the unweighted mean across the five benchmarks. Overall truncation is the percentage of responses whose finish reason is length and whose generated length reaches the applicable output limit, out of 27,904 responses.

In Tables~\ref{tab:extended_generation_8192}--\ref{tab:extended_filtered_16384}, bold marks the higher value for each performance metric within each matched model-size pair (1.7B or 8B). For truncation rates, it marks the lower value. Ties are bolded in both rows.

At both output limits, CBPO-Unified has higher avg@32 on every benchmark and equal or higher pass@32 than the corresponding original checkpoint. At 16,384 tokens, its average avg@32 exceeds Qwen3-1.7B by 17.30 points and Qwen3-8B by 22.96 points.

\begin{table}[H]
  
  \centering
  \scriptsize
  \setlength{\tabcolsep}{3pt}
  \caption{Standard results at an 8,192-token evaluation output limit. Each benchmark entry reports avg@32 / pass@32 (\%); Avg. is the unweighted mean across the five benchmarks.}
  \label{tab:extended_generation_8192}
  \begin{tabular*}{\linewidth}{@{\extracolsep{\fill}}l*{7}{c}@{}}
    \toprule
    \textbf{Model} & \textbf{MATH500} & \textbf{MinervaMath} & \textbf{AMC23} & \textbf{AIME25} & \textbf{AIME26} & \textbf{Avg.} & \textbf{Overall trunc.} \\
    \midrule
    Qwen3-1.7B & 76.99\,/\,88.80 & 26.29\,/\,40.44 & 63.20\,/\,87.50 & 22.92\,/\,43.33 & 22.81\,/\,46.67 & 42.44\,/\,61.35 & 18.21 \\
    CBPO-Unified-1.7B & \textbf{87.47}\,/\,\textbf{97.80} & \textbf{37.34}\,/\,\textbf{50.74} & \textbf{76.33}\,/\,\textbf{100.00} & \textbf{41.88}\,/\,\textbf{66.67} & \textbf{42.08}\,/\,\textbf{63.33} & \textbf{57.02}\,/\,\textbf{75.71} & \textbf{8.59} \\
    Qwen3-8B & 81.74\,/\,89.20 & 33.08\,/\,45.59 & 73.59\,/\,95.00 & 28.54\,/\,53.33 & 31.56\,/\,50.00 & 49.70\,/\,66.62 & 18.64 \\
    CBPO-Unified-8B & \textbf{98.13}\,/\,\textbf{100.00} & \textbf{61.87}\,/\,\textbf{72.06} & \textbf{97.19}\,/\,\textbf{100.00} & \textbf{68.33}\,/\,\textbf{83.33} & \textbf{69.69}\,/\,\textbf{83.33} & \textbf{79.04}\,/\,\textbf{87.75} & \textbf{4.00} \\
    \bottomrule
  \end{tabular*}
\end{table}

\begin{table}[H]
  
  \centering
  \scriptsize
  \setlength{\tabcolsep}{3pt}
  \caption{Standard results at a 16,384-token evaluation output limit. Each benchmark entry reports avg@32 / pass@32 (\%); Avg. is the unweighted mean across the five benchmarks.}
  \label{tab:extended_generation_16384}
  \begin{tabular*}{\linewidth}{@{\extracolsep{\fill}}l*{7}{c}@{}}
    \toprule
    \textbf{Model} & \textbf{MATH500} & \textbf{MinervaMath} & \textbf{AMC23} & \textbf{AIME25} & \textbf{AIME26} & \textbf{Avg.} & \textbf{Overall trunc.} \\
    \midrule
    Qwen3-1.7B & 80.73\,/\,90.40 & 27.26\,/\,41.91 & 76.09\,/\,95.00 & 31.35\,/\,63.33 & 31.15\,/\,70.00 & 49.32\,/\,72.13 & 5.42 \\
    CBPO-Unified-1.7B & \textbf{95.02}\,/\,\textbf{99.20} & \textbf{41.64}\,/\,\textbf{55.51} & \textbf{89.69}\,/\,\textbf{100.00} & \textbf{52.29}\,/\,\textbf{83.33} & \textbf{54.48}\,/\,\textbf{83.33} & \textbf{66.62}\,/\,\textbf{84.28} & \textbf{2.07} \\
    Qwen3-8B & 86.92\,/\,91.60 & 34.58\,/\,46.32 & 90.00\,/\,\textbf{100.00} & 53.54\,/\,80.00 & 57.19\,/\,83.33 & 64.45\,/\,80.25 & 5.09 \\
    CBPO-Unified-8B & \textbf{99.64}\,/\,\textbf{100.00} & \textbf{65.95}\,/\,\textbf{77.57} & \textbf{100.00}\,/\,\textbf{100.00} & \textbf{85.21}\,/\,\textbf{96.67} & \textbf{86.25}\,/\,\textbf{96.67} & \textbf{87.41}\,/\,\textbf{94.18} & \textbf{0.72} \\
    \bottomrule
  \end{tabular*}
\end{table}

We exclude a response from the conditional metrics only when its finish reason is length and its generated length reaches the applicable output limit. If all 32 responses for a problem meet this criterion, that problem is excluded from both metrics. Filtered avg is the unweighted mean, over retained problems, of each problem's fraction of remaining responses that are correct; filtered pass is the fraction of retained problems with at least one correct remaining response. Avg. is the unweighted mean across benchmarks. Because the retained problem sets and response counts vary, these filtered metrics are descriptive diagnostics rather than standard avg@32/pass@32 or a controlled comparison isolating the effect of truncation. Tables~\ref{tab:extended_filtered_8192} and~\ref{tab:extended_filtered_16384} report these metrics, and Table~\ref{tab:extended_fully_truncated} gives the number of problems removed from both metrics because all 32 responses were truncated.

\begin{table}[H]
  
  \centering
  \scriptsize
  \setlength{\tabcolsep}{1.2pt}
  \caption{Conditional metrics after excluding cap-hit responses at an 8,192-token output limit. Each benchmark entry reports filtered avg / filtered pass (\%).}
  \label{tab:extended_filtered_8192}
  \begin{tabular*}{\linewidth}{@{\extracolsep{\fill}}l*{6}{c}@{}}
    \toprule
    \textbf{Model} & \textbf{MATH500} & \textbf{MinervaMath} & \textbf{AMC23} & \textbf{AIME25} & \textbf{AIME26} & \textbf{Avg.} \\
    \midrule
    Qwen3-1.7B & 83.91\,/\,89.96 & 27.22\,/\,40.74 & 88.52\,/\,94.44 & 47.66\,/\,60.00 & 62.06\,/\,75.00 & 61.87\,/\,72.03 \\
    CBPO-Unified-1.7B & \textbf{91.91}\,/\,\textbf{97.80} & \textbf{43.23}\,/\,\textbf{50.74} & \textbf{90.49}\,/\,\textbf{100.00} & \textbf{63.25}\,/\,\textbf{71.43} & 60.07\,/\,70.37 & \textbf{69.79}\,/\,\textbf{78.07} \\
    Qwen3-8B & 89.12\,/\,91.75 & 35.62\,/\,46.79 & 96.61\,/\,97.30 & 71.90\,/\,73.33 & \textbf{83.33}\,/\,\textbf{86.67} & 75.32\,/\,79.17 \\
    CBPO-Unified-8B & \textbf{98.13}\,/\,\textbf{100.00} & \textbf{65.37}\,/\,\textbf{72.06} & \textbf{100.00}\,/\,\textbf{100.00} & \textbf{81.63}\,/\,\textbf{86.21} & 80.62\,/\,86.21 & \textbf{85.15}\,/\,\textbf{88.89} \\
    \bottomrule
  \end{tabular*}
\end{table}

\begin{table}[H]
  
  \centering
  \scriptsize
  \setlength{\tabcolsep}{1.2pt}
  \caption{Conditional metrics after excluding cap-hit responses at a 16,384-token output limit. Each benchmark entry reports filtered avg / filtered pass (\%).}
  \label{tab:extended_filtered_16384}
  \begin{tabular*}{\linewidth}{@{\extracolsep{\fill}}l*{6}{c}@{}}
    \toprule
    \textbf{Model} & \textbf{MATH500} & \textbf{MinervaMath} & \textbf{AMC23} & \textbf{AIME25} & \textbf{AIME26} & \textbf{Avg.} \\
    \midrule
    Qwen3-1.7B & 81.88\,/\,90.40 & 27.29\,/\,41.91 & 82.37\,/\,95.00 & 39.64\,/\,62.07 & 40.29\,/\,72.41 & 54.29\,/\,72.36 \\
    CBPO-Unified-1.7B & \textbf{95.25}\,/\,\textbf{99.20} & \textbf{41.87}\,/\,\textbf{55.51} & \textbf{93.64}\,/\,\textbf{100.00} & \textbf{61.77}\,/\,\textbf{83.33} & \textbf{62.74}\,/\,\textbf{83.33} & \textbf{71.05}\,/\,\textbf{84.28} \\
    Qwen3-8B & 88.30\,/\,91.58 & 34.66\,/\,46.32 & 96.77\,/\,\textbf{100.00} & 73.76\,/\,85.19 & 75.88\,/\,85.71 & 73.87\,/\,81.76 \\
    CBPO-Unified-8B & \textbf{99.64}\,/\,\textbf{100.00} & \textbf{66.06}\,/\,\textbf{77.57} & \textbf{100.00}\,/\,\textbf{100.00} & \textbf{93.17}\,/\,\textbf{96.67} & \textbf{93.25}\,/\,\textbf{96.67} & \textbf{90.42}\,/\,\textbf{94.18} \\
    \bottomrule
  \end{tabular*}
\end{table}

\begin{table}[H]
  
  \centering
  \small
  \setlength{\tabcolsep}{4pt}
  \caption{Problems excluded from both conditional metrics because all 32 responses reached the output limit. Counts are shown as benchmark problems / total problems.}
  \label{tab:extended_fully_truncated}
  \begin{tabularx}{\linewidth}{@{}>{\raggedright\arraybackslash}p{1.35in}c *{5}{>{\centering\arraybackslash}X}@{}}
    \toprule
    \textbf{Model} & \shortstack{Output limit\\(tokens)}
      & \shortstack{MATH500\\(500)} & \shortstack{MinervaMath\\(272)}
      & \shortstack{AMC23\\(40)} & \shortstack{AIME25\\(30)} & \shortstack{AIME26\\(30)} \\
    \midrule
    Qwen3-1.7B & 8,192 & 12 & 2 & 4 & 10 & 14 \\
    CBPO-Unified-1.7B & 8,192 & 0 & 0 & 0 & 2 & 3 \\
    Qwen3-8B & 8,192 & 15 & 7 & 3 & 15 & 15 \\
    CBPO-Unified-8B & 8,192 & 0 & 0 & 0 & 1 & 1 \\
    \midrule
    Qwen3-1.7B & 16,384 & 0 & 0 & 0 & 1 & 1 \\
    CBPO-Unified-1.7B & 16,384 & 0 & 0 & 0 & 0 & 0 \\
    Qwen3-8B & 16,384 & 1 & 0 & 0 & 3 & 2 \\
    CBPO-Unified-8B & 16,384 & 0 & 0 & 0 & 0 & 0 \\
    \bottomrule
  \end{tabularx}
\end{table}

\section{Hyperparameters and Training-System Settings}
\label{app:training_system_settings}
Table~\ref{tab:rollout_only_parameters} records the core Qwen3-1.7B CBPO configuration.
The data-source restrictions that define CBPO-Offline, CBPO-OnPolicy, CBPO-OffPolicy, and CBPO-Unified follow the special forms specified in Sec.~\ref{sec:special_forms}. The shared target, generation settings, and optimization budget remain aligned across the corresponding comparisons.

The response budgets use 16,254 retained prompts and four responses per prompt per epoch, with six reference-collection epochs and four current-policy epochs.
This gives 24 reference responses and 16 current-policy responses per prompt, for a total of 40.

\begin{table}[H]
  \centering
  \small
  \setlength{\tabcolsep}{3pt}
  \vspace{-5pt}
  \caption{Core Qwen3-1.7B CBPO settings for data, generation, optimization, and hardware.}
  \label{tab:rollout_only_parameters}
  \begin{tabular*}{\linewidth}{@{\extracolsep{\fill}}>{\raggedright\arraybackslash}p{0.39\columnwidth}>{\raggedright\arraybackslash}p{0.55\columnwidth}@{}}
    \toprule
    \textbf{Setting} & \textbf{Value} \\
    \midrule
    Method scope & Core Qwen3-1.7B CBPO configuration \\
    Base and reference model & Qwen3-1.7B \\
    Online prompt pool & Decontaminated RLVR-Math-16k, 16,254 prompts \\
    Held-out evaluation & 842 prompts across AIME25, MinervaMath, AMC23, and MATH500 \\
    Evaluation samples & 32 responses per prompt \\
    Target prompt batch size & $P=512$ \\
    Responses per prompt & $G=4$ \\
    Replay minibatch size & $Q=2048$ responses \\
    Sequence batch size & 16 responses \\
    Gradient-accumulation steps & 128 \\
    Reference normalizer samples per prompt & $K_{\mathrm{ref}}=24$ \\
    Reward-temperature coefficient & $\beta=0.2$ \\
    Actor learning rate & $1\times10^{-6}$ \\
    Target optimizer-step budget & $U_{\mathrm{target}}=1280$ \\
    Target prompt batches per epoch & $B_{\mathrm{epoch}}=32$ \\
    Total epochs & $E=10$ \\
    Replay updates per trainable iteration & $M=10$ \\
    Total response budget & 650,160 \\
    Reference-response budget & 390,096 \\
    Current-policy response budget & 260,064 \\
    Reference sampling temperature & 1.0 \\
    Reference sampling top-$p$ & 1.0 \\
    Training temperature & 1.0 \\
    Training top-$p$ & 0.95 \\
    Maximum prompt length & 1024 \\
    Maximum response length & 2048 \\
    Hardware & Two NVIDIA B200 GPUs \\
    Training seeds for Tables~\ref{tab:core_cross_regime}--\ref{tab:one_vs_two_stage} & Five independent seeds \\
    Training seed for other ablations and diagnostics & 42 \\
    \bottomrule
  \end{tabular*}
  \vspace{-10pt}
\end{table}

\flushbottom

\section{Additional Diagnostic Protocols}
\label{app:diagnostic_protocols}

\subsection{Heterogeneous Offline Sources}
\label{app:heterogeneous_sources}

We compare CBPO-Unified and BRIDGE on Qwen3-1.7B under four offline-source conditions to examine sensitivity to the behavior that generated the demonstrations.
The three individual sources are DeepSeek-R1-Distill-Qwen-1.5B, DeepSeek-R1-Distill-Qwen-7B, and DeepSeek-R1-Distill-Qwen-32B \citep{deepseekai2025r1}.
Table~\ref{tab:heterogeneous_sources} abbreviates them as R1-Qwen-1.5B, R1-Qwen-7B, and R1-Qwen-32B.
Each teacher generates at most 256 candidate trajectories per prompt at temperature $1.0$ and top-$p=0.95$, with the response-length limit reported in Table~\ref{tab:rollout_only_parameters}.
We retain the common prompt set for which R1-Qwen-1.5B produces at least two positive trajectories within this budget excluding other prompts which retains 8,724 prompts for all conditions.
For every retained prompt, CBPO-Unified uses two positive and one negative offline trajectory, whereas BRIDGE uses the same two positive trajectories.
The mixed condition uses the union of the three teacher-specific datasets for each method.

\begin{table}[!htbp]
  \centering
  \small
  \setlength{\tabcolsep}{1.5pt}
  \renewcommand{\arraystretch}{1.05}
  \caption{Qwen3-1.7B performance under three individual teacher sources and the union of all three sources on the common retained prompt set. The mixed condition combines the complete teacher-specific datasets for each method.}
  \label{tab:heterogeneous_sources}
  \begin{tabularx}{\linewidth}{@{}ll*{10}{>{\centering\arraybackslash}X}@{}}
    \toprule
    \multirow[c]{2}{*}{\raisebox{-2pt}{\textbf{Source}}}
      & \multirow[c]{2}{*}{\raisebox{-2pt}{\textbf{Methods}}}
      & \multicolumn{2}{c}{\textbf{AIME25}}
      & \multicolumn{2}{c}{\textbf{MinervaMath}}
      & \multicolumn{2}{c}{\textbf{AMC23}}
      & \multicolumn{2}{c}{\textbf{MATH500}}
      & \multicolumn{2}{c}{\textbf{Avg.}} \\
    \cmidrule(lr){3-4}\cmidrule(lr){5-6}\cmidrule(lr){7-8}\cmidrule(lr){9-10}\cmidrule(lr){11-12}
     &  & \metrichead{avg@32} & \metrichead{pass@32} & \metrichead{avg@32} & \metrichead{pass@32} & \metrichead{avg@32} & \metrichead{pass@32} & \metrichead{avg@32} & \metrichead{pass@32} & \metrichead{avg@32} & \metrichead{pass@32} \\
    \midrule
    \multirow{2}{*}{R1-Qwen-1.5B} 
            & BRIDGE & 15.21 & 26.67 & 21.76 & 40.07 & 54.30 & 75.00 & 70.23 & 82.00 & 40.37 & 55.94 \\
      & CBPO-Unified & 18.33 & 33.33 & 23.43 & 41.18 & 58.13 & 82.50 & 71.31 & 85.20 & 42.80 & 60.55 \\
    \midrule
    \multirow{2}{*}{R1-Qwen-7B} 
            & BRIDGE & 17.60 & 33.33 & 22.55 & 40.81 & 56.41 & 82.50 & 71.01 & 82.60 & 41.89 & 59.81 \\
      & CBPO-Unified & 19.90 & 43.33 & 23.95 & 41.54 & 58.98 & 87.50 & 72.05 & 86.20 & 43.72 & 64.64 \\
    \midrule
    \multirow{2}{*}{R1-Qwen-32B} 
            & BRIDGE & 18.33 & 43.33 & 23.01 & 41.18 & 57.66 & 85.00 & 71.31 & 83.00 & 42.58 & 63.13 \\
      & CBPO-Unified & 20.42 & 50.00 & 24.21 & 41.91 & 59.84 & 92.50 & 72.79 & 87.40 & 44.31 & 67.95 \\
    \midrule
    \multirow{2}{*}{Mixed} 
            & BRIDGE & 18.96 & 43.33 & 23.21 & 41.54 & 57.89 & 87.50 & 71.65 & 83.20 & 42.93 & 63.89 \\
      & CBPO-Unified & 21.98 & 53.33 & 24.54 & 42.65 & 60.39 & 95.00 & 73.01 & 87.80 & 44.98 & 69.70 \\
    \bottomrule
  \end{tabularx}
\end{table}

Aggregate performance increases for both methods from R1-Qwen-1.5B through R1-Qwen-7B to R1-Qwen-32B, and the mixed condition gives the highest aggregate scores for each method.
For CBPO-Unified, mixing the three sources improves \texttt{avg@32/pass@32} by 2.18/9.15 points over R1-Qwen-1.5B and by 0.67/1.75 points over R1-Qwen-32B.
CBPO-Unified outperforms BRIDGE under all four source conditions, with aggregate margins of 1.73--2.43 points in \texttt{avg@32} and 4.61--5.81 points in \texttt{pass@32}.
Under the mixed condition, these margins are 2.05/5.81 points.

\subsection{Prompt-level Reference Normalizer Diagnostics}
\label{app:normalizer_diagnostics}

The diagnostic examines how the frozen normalizer varies with the reference policy's empirical success on the 16,254 training prompts retained after decontamination.
For a prompt $x$, let $c_x$ be the number of reward-$1$ completions among the 24 responses used to construct its frozen normalizer.
We group prompts by $c_x\in\{0\text{--}1,2\text{--}5,6\text{--}11,12\text{--}23,24\}$, combining reference banks with at most one correct response in the first group.
Table~\ref{tab:normalizer_difficulty} reports the number and proportion of prompts in each group, together with the frequencies of reward-$0$ and reward-$0.1$ reference responses.

\begin{table}[!htbp]
  \centering
  \small
  \setlength{\tabcolsep}{3pt}
  \caption{Reference-bank composition by reference success count. Reward frequencies are computed over the recorded reference responses in each group.}
  \label{tab:normalizer_difficulty}
  \begin{tabularx}{\linewidth}{@{}c *{4}{>{\centering\arraybackslash}X}@{}}
    \toprule
    \shortstack{Correct responses} & \shortstack{Prompts} & \shortstack{Proportion (\%)}
      & \shortstack{Reward 0 (\%)} & \shortstack{Reward 0.1 (\%)} \\
    \midrule
    0--1 & 5,963 & 36.69 & 77.30 & 20.33 \\
    2--5 & 3,157 & 19.42 & 63.56 & 22.98 \\
    6--11 & 937 & 5.76 & 52.19 & 13.06 \\
    12--23 & 3,110 & 19.13 & 17.38 & 1.94 \\
    24 & 3,087 & 18.99 & 0.00 & 0.00 \\
    \bottomrule
  \end{tabularx}
\end{table}

The reference bank contains at most one correct response for 36.69\% of the retained prompts, whereas 18.99\% have 24 correct responses.

\paragraph{Confidence bounds for the reference normalizer.}
For this analysis, assume independent reference responses $Y_i\sim\pi_{\mathrm{ref}}(\cdot\mid x)$ for a fixed prompt $x$, and let $W_i=e^{r(x,Y_i)/\beta}$.
With $\beta=0.2$ and $r\in\{0,0.1,1\}$, the weights take the three values $\mathbf w=(1,\sqrt{e},e^5)$.
We use their exact trinomial sampling distribution to construct confidence limits by test inversion \citep{miratrix2009trinomial}.
Let $\mathbf p\in\Delta_2$ denote the three reward probabilities, where $\Delta_2=\{\mathbf p\in[0,1]^3:\sum_jp_j=1\}$.
For $\mathbf N\sim\operatorname{Multinomial}(K,\mathbf p)$, define $T_K=\mathbf w^\top\mathbf N/K$ and $\mu(\mathbf p)=\mathbf w^\top\mathbf p$.
The observed value of $T_K$ is $\widehat Z_K(x)$.
Allocating $\alpha=0.025$ to each tail gives
\begin{equation}
  \begin{aligned}
    L_K(x)&=\inf\{\mu(\mathbf p):\mathbf p\in\Delta_2,\ \Pr_{\mathbf p}(T_K\ge\widehat Z_K(x))>\alpha\},\\
    U_K(x)&=\sup\{\mu(\mathbf p):\mathbf p\in\Delta_2,\ \Pr_{\mathbf p}(T_K\le\widehat Z_K(x))>\alpha\}.
  \end{aligned}
  \label{eq:normalizer_exact_limits}
\end{equation}
Each probability is evaluated from the finite multinomial distribution, and the endpoints optimize over all reward-probability vectors consistent with the corresponding tail test.
These are Buehler limits under sample-mean ordering \citep{lloyd2010buehler}.
For this fixed ordering and tail allocation, the upper limit is the smallest nondecreasing valid one-sided upper limit, and the lower limit is the largest nondecreasing valid one-sided lower limit.
The numerical endpoints are enclosed outward with absolute error at most $2\times10^{-6}$.
The resulting absolute log-error bound is
\begin{equation}
  u_K(x)=\max\left\{
    \log\frac{\widehat Z_K(x)}{L_K(x)},
    \log\frac{U_K(x)}{\widehat Z_K(x)}
  \right\},
  \qquad
  \Pr\!\left(
    |\log\widehat Z_K(x)-\log Z_{\mathrm{ref}}(x)|\le u_K(x)
  \right)\ge0.95.
  \label{eq:normalizer_exact_log_bound}
\end{equation}
To see the coverage guarantee, each inclusive tail probability, evaluated under the true $\mathbf p$ at the observed statistic, is at most $\alpha$ with probability at most $\alpha$.
By a union bound, with probability at least $1-2\alpha$, $\mu(\mathbf p)=Z_{\mathrm{ref}}(x)$ lies between the limits in Eq.~\eqref{eq:normalizer_exact_limits}.
Monotonicity of the logarithm then gives the stated bound.
This guarantee is pointwise in $x$ and $K$ before grouping by observed reference success.
The bound controls finite-sample error in estimating the reference normalizer $Z_{\mathrm{ref}}(x)$.
Using nested reference responses in collection order, Table~\ref{tab:normalizer_confidence} summarizes $u_K$ across all 16,254 training prompts retained after decontamination, using the same reference-success groups as Table~\ref{tab:normalizer_difficulty}.
The bucket means and 90th percentiles summarize the computed bounds across prompts within each group.
The overall mean weights the group means by the prompt counts in Table~\ref{tab:normalizer_difficulty}.

\begin{table}[!htbp]
  \centering
  \small
  \setlength{\tabcolsep}{3pt}
  \caption{Pointwise 95\% exact trinomial bounds on absolute log-normalizer estimation error by reference-sample budget $K$, under independent sampling from $\pi_{\mathrm{ref}}$.}
  \label{tab:normalizer_confidence}
  \begin{tabular*}{\linewidth}{@{\extracolsep{\fill}}ccccccccc@{}}
    \toprule
    & \multicolumn{4}{c}{Mean $u_K$} & \multicolumn{4}{c}{90th percentile $u_K$} \\
    \cmidrule(lr){2-5}\cmidrule(lr){6-9}
    \shortstack{Correct reference\\responses} & $K=8$ & $K=16$ & $K=24$ & $K=32$ & $K=8$ & $K=16$ & $K=24$ & $K=32$ \\
    \midrule
    0--1 & 3.9088 & 3.3017 & 2.9156 & 2.6821 & 4.0156 & 3.4453 & 3.0912 & 2.8242 \\
    2--5 & 2.6713 & 1.7165 & 1.3281 & 1.0938 & 4.0156 & 2.1142 & 1.6655 & 1.3406 \\
    6--11 & 1.6336 & 0.9610 & 0.7252 & 0.5992 & 2.5843 & 1.1732 & 0.8989 & 0.7624 \\
    12--23 & 0.7080 & 0.4044 & 0.3010 & 0.2564 & 1.1295 & 0.6222 & 0.4872 & 0.3940 \\
    24 & 0.4572 & 0.2288 & 0.1526 & 0.1282 & 0.4572 & 0.2288 & 0.1526 & 0.1282 \\
    \midrule
    Overall & 2.2693 & 1.7209 & 1.4560 & 1.3044 & 4.0156 & 3.4453 & 3.0912 & 2.8242 \\
    \bottomrule
  \end{tabular*}
\end{table}

The overall mean bound decreases from 2.2693 at $K=8$ to 1.7209 at $K=16$, 1.4560 at $K=24$, and 1.3044 at $K=32$.
At $K=24$, the mean bound is 0.1526 for the all-correct reference group and 2.9156 for the group with at most one correct response, indicating substantially different estimation uncertainty across reference-success groups.

\begin{figure}[!htbp]
  \centering
  \includegraphics[width=\linewidth]{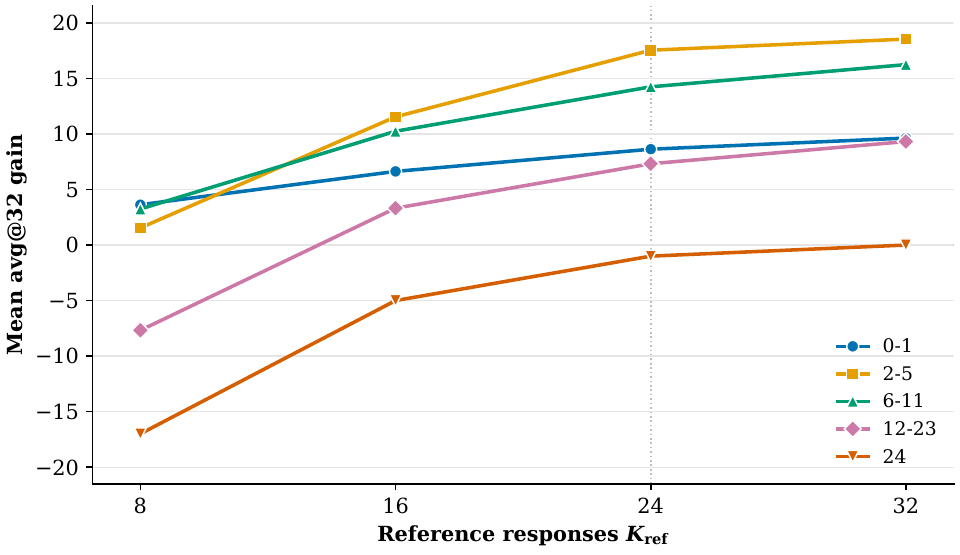}
  \caption{Mean \texttt{avg@32} gain over the reference-policy baseline at each reference-sample budget. The five curves correspond to difficulty groups defined by the number of correct responses in the 24-response reference bank.}
  \label{fig:normalizer_gain_diagnostic}
\end{figure}

\paragraph{Policy improvement across reference-sample budgets.}
We relate the estimation bounds to learning gains for the four Qwen3-1.7B configurations with $K_{\mathrm{ref}}\in\{8,16,24,32\}$ in Table~\ref{tab:kref_controlled_main} (Left).
For each reference-success group $b$, let $\bar a_{\mathrm{ref},b}$ denote the empirical correctness of the frozen reference policy and $\bar a_{\mathrm{final},b}(K)$ the final checkpoint's mean \texttt{avg@32} score, both expressed on the $[0,1]$ scale.
The group-level gain is $G_b(K)=100[\bar a_{\mathrm{final},b}(K)-\bar a_{\mathrm{ref},b}]$ percentage points.
The five groups are defined by the 24-response reference bank, including prompts with no correct reference response.
All four budgets use the same reference-policy baseline for each group, and the training-prompt diagnostic is reported separately from the four held-out benchmarks.
Fig.~\ref{fig:normalizer_gain_diagnostic} displays the five $G_b(K)$ curves on one set of axes.
Comparing these curves with the corresponding rows of Table~\ref{tab:normalizer_confidence} relates changes in estimation uncertainty to gains within each fixed difficulty group.

Across all five groups, $G_b(K)$ increases over the four evaluated budgets, but the marginal improvement diminishes as $K$ grows.
Increasing $K$ from 24 to 32 requires generating one third more reference responses per prompt, but yields only 0--1 additional \texttt{avg@32} points in each group.
These limited gains do not justify the additional generation cost in our setting, so we use $K_{\mathrm{ref}}=24$.
This choice is consistent with the empirical saturation observed in A$^\star$-PO, which uses $N=8$ reference responses per prompt and reports a MATH500 performance plateau around this budget \citep{brantley2025astarpo}.

\subsection{Negative-Trajectory Count and Type}
\label{app:negative_controls}

\paragraph{Adding negatives to fixed positive demonstrations.}
The count comparison retains the same two positive offline trajectories per prompt and adds 0, 1, or 2 negative offline trajectories from nested subsets in generation order.
Unlike the fixed-total composition sweep in Table~\ref{tab:negative_ratio}, this comparison keeps the positive demonstrations unchanged.
All configurations follow the CBPO-Unified protocol.

Eligible records are sampled uniformly, as in Sec.~\ref{sec:pr_schedule}.
Table~\ref{tab:negative_count_control} reports the four held-out benchmarks and their unweighted average.
Moving from 0 to 1 negative improves aggregate \texttt{avg@32/pass@32} by 1.40/2.03 points, whereas moving from 1 to 2 negatives lowers them by 0.45/1.70 points.

\begin{table}[!htbp]
  \centering
  \small
  \setlength{\tabcolsep}{2pt}
  \caption{Negative-count comparison within CBPO-Unified with two fixed positive offline trajectories per prompt. All other data sources and training settings are held fixed.}
  \label{tab:negative_count_control}
  \begin{tabularx}{\linewidth}{@{}l *{10}{>{\centering\arraybackslash}X}@{}}
    \toprule
    \multirow{2}{*}{\textbf{Negatives}}
      & \multicolumn{2}{c}{\textbf{AIME25}}
      & \multicolumn{2}{c}{\textbf{MinervaMath}}
      & \multicolumn{2}{c}{\textbf{AMC23}}
      & \multicolumn{2}{c}{\textbf{MATH500}}
      & \multicolumn{2}{c}{\textbf{Avg.}} \\
    \cmidrule(lr){2-3}\cmidrule(lr){4-5}\cmidrule(lr){6-7}\cmidrule(lr){8-9}\cmidrule(lr){10-11}
      & \metrichead{avg@32} & \metrichead{pass@32}
      & \metrichead{avg@32} & \metrichead{pass@32}
      & \metrichead{avg@32} & \metrichead{pass@32}
      & \metrichead{avg@32} & \metrichead{pass@32}
      & \metrichead{avg@32} & \metrichead{pass@32} \\
    \midrule
    0 & 19.38 & 46.67 & 23.23 & 41.18 & 58.91 & 92.50 & 71.11 & 86.40 & 43.16 & 66.69 \\
    1 & 20.63 & 50.00 & 24.47 & 42.28 & 60.16 & 95.00 & 72.99 & 87.60 & 44.56 & 68.72 \\
    2 & 20.10 & 46.67 & 24.21 & 41.91 & 59.69 & 92.50 & 72.45 & 87.00 & 44.11 & 67.02 \\
    \bottomrule
  \end{tabularx}
\end{table}

\paragraph{Negative reward and response content.}
For every retained prompt, the candidate bank generated by the DeepSeek-R1-Distill-Qwen-32B teacher contains the two selected positive offline trajectories, sufficient negative offline trajectories with rewards $0$ and $0.1$, and negative responses from both content categories considered below.
The no-negative reference uses only the same two positive offline trajectories as in Table~\ref{tab:core_cross_regime}.
Both comparisons add the first negative candidate in generation order that satisfies the corresponding negative-type criterion.
The reward comparison separates unparseable completions with reward $0$ from parseable but incorrect completions with reward $0.1$.
The content comparison distinguishes near-correct derivations from clearly invalid reasoning by manual review.
A near-correct response follows a largely valid derivation but contains a localized mathematical or final-answer error.
A clearly invalid response contains unsupported or incoherent derivations that do not yield a valid solution.
The within-comparison changes in aggregate performance quantify the effect of adding each negative type to the same positive data.

\begin{table}[!htbp]
  \centering
  \small
  \setlength{\tabcolsep}{2pt}
  \caption{Negative-type comparisons with training seed 42 and fixed positive demonstrations on the same 16,254 prompts. Both groups share the no-negative reference from Table~\ref{tab:negative_count_control}, and each negative condition uses one response per prompt.}
  \label{tab:negative_type_control}
  \begin{tabularx}{\linewidth}{@{}l *{10}{>{\centering\arraybackslash}X}@{}}
    \toprule
    \multirow{2}{*}{\textbf{Negative type}}
      & \multicolumn{2}{c}{\textbf{AIME25}}
      & \multicolumn{2}{c}{\textbf{MinervaMath}}
      & \multicolumn{2}{c}{\textbf{AMC23}}
      & \multicolumn{2}{c}{\textbf{MATH500}}
      & \multicolumn{2}{c}{\textbf{Avg.}} \\
    \cmidrule(lr){2-3}\cmidrule(lr){4-5}\cmidrule(lr){6-7}\cmidrule(lr){8-9}\cmidrule(lr){10-11}
      & \metrichead{avg@32} & \metrichead{pass@32}
      & \metrichead{avg@32} & \metrichead{pass@32}
      & \metrichead{avg@32} & \metrichead{pass@32}
      & \metrichead{avg@32} & \metrichead{pass@32}
      & \metrichead{avg@32} & \metrichead{pass@32} \\
    \midrule
    No negative & 19.38 & 46.67 & 23.23 & 41.18 & 58.91 & 92.50 & 71.11 & 86.40 & 43.16 & 66.69 \\
    \midrule
    Reward 0 & 19.48 & 46.67 & 23.10 & 40.81 & 59.14 & 92.50 & 71.03 & 86.40 & 43.19 & 66.59 \\
    Reward 0.1 & 20.94 & 50.00 & 24.76 & 42.65 & 60.39 & 95.00 & 73.11 & 88.00 & 44.80 & 68.91 \\
    \midrule
    Clearly invalid & 18.85 & 40.00 & 22.87 & 40.44 & 58.52 & 85.00 & 70.33 & 85.20 & 42.64 & 62.66 \\
    Near-correct & 21.35 & 53.33 & 24.89 & 43.01 & 60.86 & 97.50 & 73.62 & 88.60 & 45.18 & 70.61 \\
    \bottomrule
  \end{tabularx}
\end{table}

Relative to the shared no-negative reference, reward-$0.1$ negatives raise aggregate \texttt{avg@32/pass@32} by 1.64/2.22 points, while reward-$0$ negatives change them by 0.03/-0.10 points.
Near-correct negatives yield a 2.02/3.92-point gain, whereas clearly invalid negatives lower the two metrics by 0.52/4.03 points.
The near-correct condition improves both metrics on all four benchmarks, and the clearly invalid condition reduces both on all four.
The benefit of adding a negative trajectory therefore depends on the selected response rather than negative count alone.

\paragraph{Composition of the training negatives.}
The single negative offline trajectory per prompt used by the CBPO-Unified configuration in Table~\ref{tab:core_cross_regime} is jointly categorized by verifier reward and response content.
Table~\ref{tab:table1_negative_type_composition} reports each joint category as a percentage of the 16,254 selected negative trajectories.

\begin{table}[!htbp]
  \centering
  \small
  \setlength{\tabcolsep}{8pt}
  \caption{Proportion (\%) of selected negative offline trajectories by verifier reward and response-content category over the five training seeds of Table~\ref{tab:core_cross_regime}. Each seed contributes 16,254 negative trajectories.}
  \label{tab:table1_negative_type_composition}
  \begin{tabular}{@{}lcc@{}}
    \toprule
    \multirow{2}{*}{\textbf{Verifier reward}}
      & \multicolumn{2}{c}{\textbf{Response content}} \\
    \cmidrule(lr){2-3}
      & \textbf{Clearly invalid} & \textbf{Near-correct} \\
    \midrule
    0   & $40.32\pm0.07$ & $19.35\pm0.04$ \\
    0.1 & $24.19\pm0.05$ & $16.13\pm0.03$ \\
    \bottomrule
  \end{tabular}
\end{table}

Across the five training seeds, the selected negatives include 35.48\% near-correct responses and 40.32\% reward-$0.1$ responses, with 16.13\% in both categories.
This mixed composition is consistent with the default one-negative configuration outperforming the no-negative reference while remaining below the all-near-correct condition in the seed-42 comparison.

\subsection{Replay-Sampling Comparisons}
\label{app:replay_sampling}

\paragraph{Matched full-training protocol.}
The four Qwen3-1.7B CBPO-Unified runs share the same initialization and use the settings in Table~\ref{tab:rollout_only_parameters}, varying only the replay-sampling rule.
Current-policy rollouts generated during each run enter that run's evolving replay buffer under the common collection schedule.
Record draws are without replacement within each minibatch, and each update draws again from the current buffer.
Evaluation follows Sec.~\ref{sec:experimental_setup}, with results reported in Table~\ref{tab:replay_sampling_comparison}.

\paragraph{Age-balanced sampling.}
This condition follows response-balanced sampling at each update, but allocates the rollout component as evenly as possible among age~1 (current-policy samples), ages 2--4, 5--16, and more than 16 policy-version intervals.
Records are sampled uniformly within each age group.
Age-group membership is recomputed from the current policy version at every update.

\paragraph{Source-balanced sampling.}
At each update, the minibatch is divided as evenly as possible among the nonempty groups comprising offline trajectories, reference responses, the latest rollout batch, and earlier rollouts.
The group quotas differ by at most one record, and sampling is uniform within each group.

\paragraph{Prompt-balanced sampling.}
Each update first selects 2,048 distinct prompts uniformly, then draws one response uniformly from the archived records of each selected prompt.
This gives every prompt the same expected training share.

\paragraph{Response-balanced sampling.}
The response-balanced baseline draws records uniformly from the current buffer, following Sec.~\ref{sec:pr_schedule}.
A prompt's expected training share is proportional to its number of eligible responses at that update.

\begin{table}[!htbp]
  \centering
  \small
  \setlength{\tabcolsep}{2pt}
  \caption{Qwen3-1.7B replay-sampling comparisons across complete training runs with matched generation and optimizer-step budgets.}
  \label{tab:replay_sampling_comparison}
  \begin{tabularx}{\linewidth}{@{}l *{10}{>{\centering\arraybackslash}X}@{}}
    \toprule
    \multirow[c]{2}{*}{\raisebox{-2pt}{\textbf{Sampling rule}}}
      & \multicolumn{2}{c}{\textbf{AIME25}}
      & \multicolumn{2}{c}{\textbf{MinervaMath}}
      & \multicolumn{2}{c}{\textbf{AMC23}}
      & \multicolumn{2}{c}{\textbf{MATH500}}
      & \multicolumn{2}{c}{\textbf{Avg.}} \\
    \cmidrule(lr){2-3}\cmidrule(lr){4-5}\cmidrule(lr){6-7}\cmidrule(lr){8-9}\cmidrule(lr){10-11}
      & {\scalebox{0.85}{\texttt{avg@32}}} & {\scalebox{0.85}{\texttt{pass@32}}}
      & {\scalebox{0.85}{\texttt{avg@32}}} & {\scalebox{0.85}{\texttt{pass@32}}}
      & {\scalebox{0.85}{\texttt{avg@32}}} & {\scalebox{0.85}{\texttt{pass@32}}}
      & {\scalebox{0.85}{\texttt{avg@32}}} & {\scalebox{0.85}{\texttt{pass@32}}}
      & {\scalebox{0.85}{\texttt{avg@32}}} & {\scalebox{0.85}{\texttt{pass@32}}} \\
    \midrule
    Age-balanced
      & 20.52 & 50.00 & 24.31 & 42.28 & 60.23 & 95.00 & 72.91 & 87.40 & 44.49 & 68.67 \\
    Source-balanced
      & 20.31 & 46.67 & 24.57 & 42.65 & 60.08 & 92.50 & 73.12 & 88.00 & 44.52 & 67.45 \\
    Prompt-balanced
      & 20.42 & 50.00 & 24.51 & 42.28 & 60.16 & 95.00 & 72.94 & 87.60 & 44.51 & 68.72 \\
    Response-balanced
      & 20.63 & 50.00 & 24.47 & 42.28 & 60.16 & 95.00 & 72.99 & 87.60 & 44.56 & 68.72 \\
    \bottomrule
  \end{tabularx}
\end{table}

\subsection{Which Historical Rollouts Remain Useful?}
\label{app:replay_content}

\paragraph{Age and residual diagnostics.}
For each sampling condition, we log the historical rollout records selected at each optimizer step before applying the parameter update.
For every replay record, we record the sampling-policy version, reward, current sequence log-probability and frozen target.
After training, we aggregate these records using their age and residual level at the time of replay.
The policy version advances once per optimizer step. Age~1 denotes current-policy samples. Historical replay comprises ages~2 and above.
The regression residual is the current sequence log-probability minus the frozen target in Eq.~\eqref{eq:cbpo_target}.
At each optimizer step, the median absolute pre-update residual among the historical records actually replayed in that update, pooled across all ages, defines the Low/High boundary.
Low and High denote the lower and upper halves after sorting by absolute residual.
Historical rollouts are grouped at ages 2--4, 5--16, and more than 16 policy-version intervals.
This study changes the sampling population of each age group by separately selecting low or high residual records,
and compares the performance gains with the age-balanced replay setting of CBPO-Unified.
Table~\ref{tab:replay_age_residual} reports the sampled records' mean absolute residual, reward-$1$ proportion, and mean sequence log-probability together with benchmark performance for each sampling condition.
The reward-$1$ proportion is computed within each listed age--residual group.

\begin{table}[!htbp]
  \centering
  \scriptsize
  \setlength{\tabcolsep}{3pt}
  \caption{Replay statistics aggregated over training and benchmark performance by policy age and residual level. Age is measured in policy-version intervals at the time of replay.}
  \label{tab:replay_age_residual}
  \begin{tabularx}{\linewidth}{@{}l *{12}{>{\centering\arraybackslash}X}@{}}
    \toprule
    \multirow{2}{*}{\textbf{Age}}
      & \multirow{2}{*}{\shortstack{\textbf{Residual}\\\textbf{level}}}
      & \multirow{2}{*}{\shortstack{\textbf{Residual}\\\textbf{Abs Avg.}}}
      & \multirow{2}{*}{\shortstack{\textbf{Log prob}\\\textbf{Avg.}}}
      & \multirow[c]{2}{=}{\centering\shortstack[c]{\textbf{Reward-1}\\\textbf{prop.(\%)}}}
      & \multicolumn{2}{c}{\textbf{AIME25}}
      & \multicolumn{2}{c}{\textbf{MinervaMath}}
      & \multicolumn{2}{c}{\textbf{AMC23}}
      & \multicolumn{2}{c}{\textbf{MATH500}} \\
    \cmidrule(lr){6-7}\cmidrule(lr){8-9}\cmidrule(lr){10-11}\cmidrule(lr){12-13}
      &  &  &  &  & \metrichead{avg@32} & \metrichead{pass@32}
      & \metrichead{avg@32} & \metrichead{pass@32}
      & \metrichead{avg@32} & \metrichead{pass@32}
      & \metrichead{avg@32} & \metrichead{pass@32}
      \\
    \midrule
    2--4  & Low  & 1.53 & -181.75 & 86.57 & 20.42 & 50.00 & 24.15 & 41.91 & 60.16 & 95.00 & 72.75 & 87.00 \\
    2--4  & High & 7.06 & -266.46 & 20.19 & 20.63 & 50.00 & 24.41 & 42.28 & 60.16 & 95.00 & 73.00 & 87.60 \\
    5--16 & Low  & 2.18 & -200.82 & 62.81 & 20.31 & 46.67 & 24.00 & 41.91 & 59.92 & 92.50 & 72.60 & 86.40 \\
    5--16 & High & 8.43 & -223.48 & 51.22 & 20.83 & 53.33 & 24.60 & 42.65 & 60.39 & 95.00 & 73.30 & 88.00 \\
    $>16$ & Low  & 4.24 & -239.89 & 38.23 & 20.10 & 46.67 & 23.49 & 41.54 & 59.38 & 90.00 & 72.00 & 86.00 \\
    $>16$ & High & 10.50 & -203.64 & 74.64 & 21.04 & 53.33 & 24.90 & 43.01 & 61.02 & 97.50 & 73.70 & 88.60 \\
    \bottomrule
  \end{tabularx}
\end{table}

Across all four benchmarks, \texttt{avg@32} increases with replay age in High and decreases in Low, while \texttt{pass@32} follows the same directions or remains unchanged.
For rollouts older than 16 policy versions, High exceeds the age-balanced CBPO-Unified baseline in Table~\ref{tab:replay_sampling_comparison} on both metrics for every benchmark, reaching unweighted four-benchmark means of 45.16/70.61 versus 44.49/68.67 (+0.67/+1.94 percentage points).
Low falls below the baseline throughout, reaching 43.74/66.05 (-0.75/-2.62 points).

As residuals grow with age, the reward-$1$ share rises from 20.19\% to 74.64\% in High but falls from 86.57\% to 38.23\% in Low, consistent with greater forgetting of earlier correct responses.
Under the squared loss in Eq.~\eqref{eq:cbpo_loss}, larger absolute residuals produce stronger gradients with respect to sequence log-probability, helping explain why the youngest High subset slightly outperforms Low despite its smaller reward-$1$ share.
The oldest High subset's less negative mean log-probability is partly explained by its larger share of short correct responses whose targets are also closer to zero, so proximity to zero does not imply a smaller target mismatch.
These comparisons support using residuals to identify older responses that still provide useful corrective signals for countering forgetting.

\subsection{Checkpoint-wise Turning Along the Optimization Paths}
\label{sec:checkpoint_turning}

The main paper summarizes the geometry of two distinct training trajectories.
The SFT$\rightarrow$DAPO path is formed by checkpoints saved in training order during the two-stage run, whereas the CBPO-Unified path is formed by the ordered checkpoints from the unified run.
Thus, the two curves below describe the local geometry of two training methods rather than two measurements of a single trajectory.

\begin{figure}[H]
  \centering
  \includegraphics[width=0.94\textwidth]{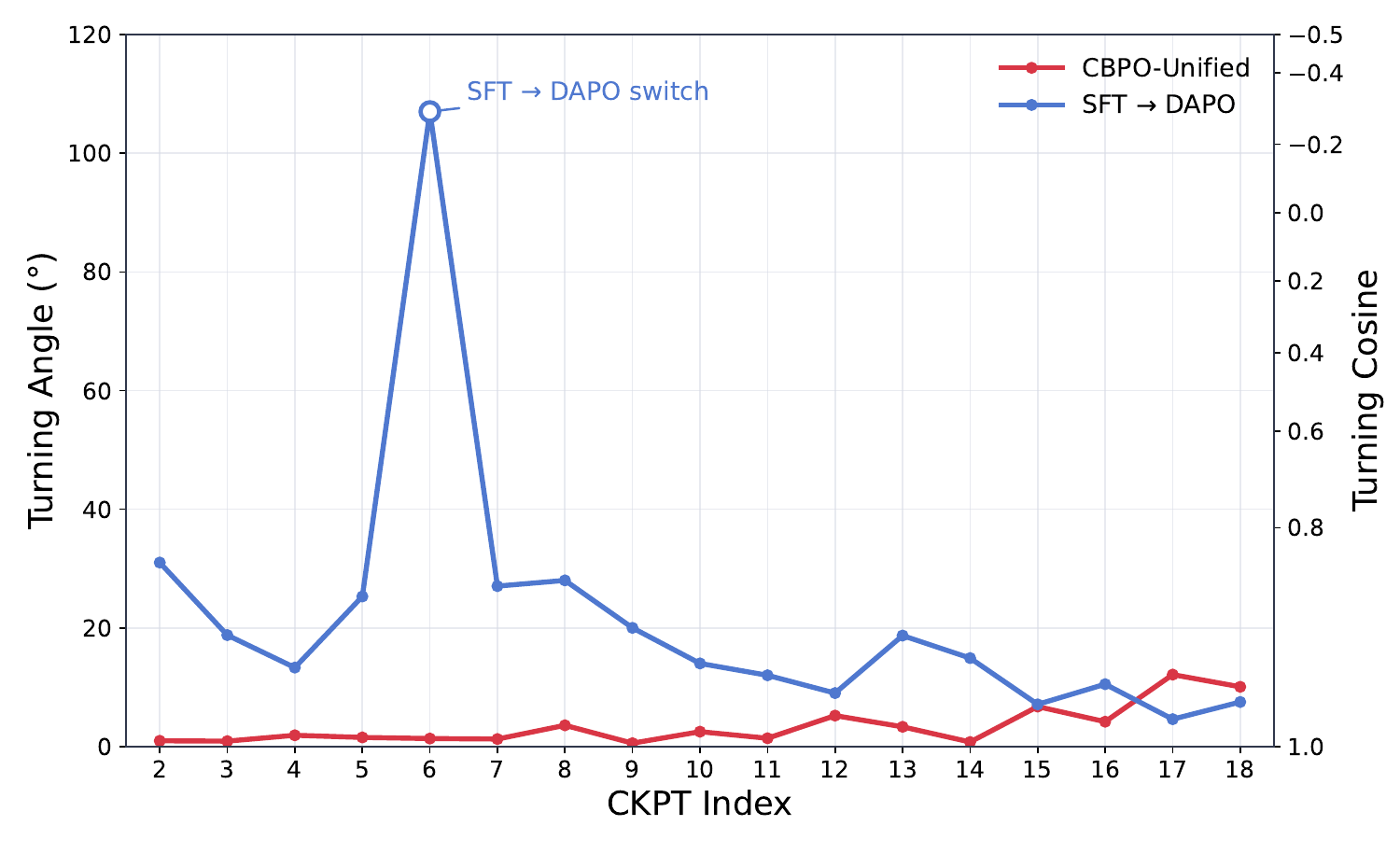}
  \vspace{-18pt}
  \caption{Checkpoint-wise turning of CBPO-Unified and SFT$\rightarrow$DAPO trajectories.}
  \label{fig:turning_by_checkpoint}
\end{figure}

Let $\mathbf z_i\in\mathbb R^2$ be the position of checkpoint $i$ in the PC1--PC2 plane used for the trajectory visualization.
For each method, the displacement ending at checkpoint $i$ is
\[
\boldsymbol{\delta}_i=\mathbf z_i-\mathbf z_{i-1}.
\]
At each interior checkpoint, the turning cosine is
\[
c_i=
\frac{\boldsymbol{\delta}_i^{\mathsf T}\boldsymbol{\delta}_{i+1}}
{\lVert\boldsymbol{\delta}_i\rVert_2\,\lVert\boldsymbol{\delta}_{i+1}\rVert_2}.
\]
The corresponding turning angle in degrees is
\[
\phi_i=\frac{180}{\pi}\arccos(c_i).
\]
A local turn requires both an incoming and an outgoing segment, so the horizontal axis contains only the interior node indices 2--18.
The left and right vertical axes in Fig.~\ref{fig:turning_by_checkpoint} report $\phi_i$ and $c_i$, respectively.
A small $\phi_i$ (equivalently, $c_i$ close to one) indicates locally straight movement.

CBPO-Unified maintains small local turns through most of training.
In contrast, SFT$\rightarrow$DAPO exhibits larger turns across many nodes and a pronounced peak of approximately $107^\circ$ at checkpoint 6, exactly where the objective changes from SFT to DAPO.
This peak is the largest single contribution to the two-stage trajectory's cumulative turn.
In the displayed two-dimensional paths, CBPO-Unified has smaller local turns, while the SFT$\rightarrow$DAPO path has its largest turn at the stage boundary.

\begin{figure}[H]
  \centering
  \includegraphics[width=0.94\textwidth]{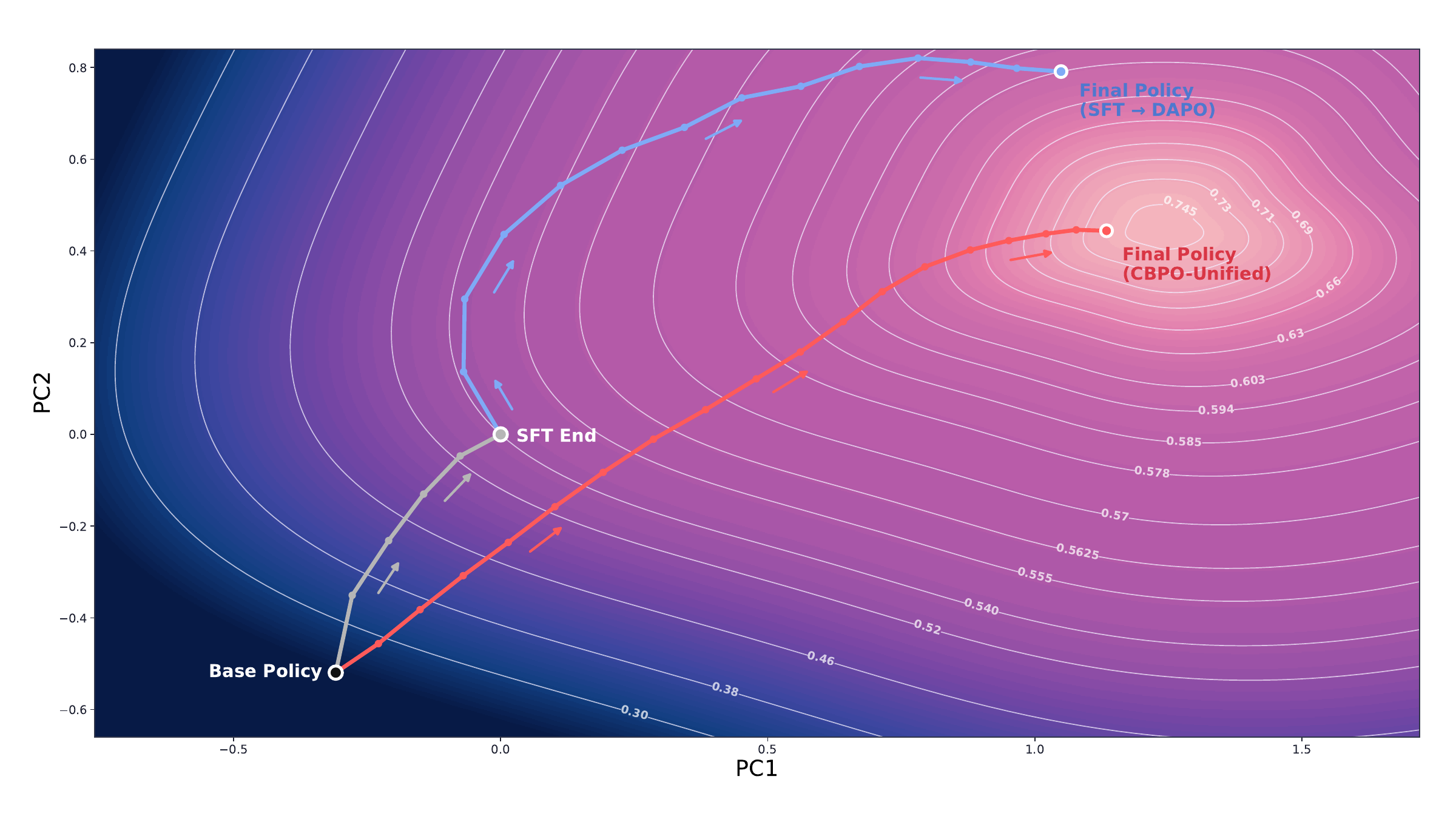}
  \caption{Optimization trajectories with five PCA components.}
  \label{fig:trajectory_pca5}
\end{figure}

\begin{figure}[H]
  \centering
  \includegraphics[width=0.94\textwidth]{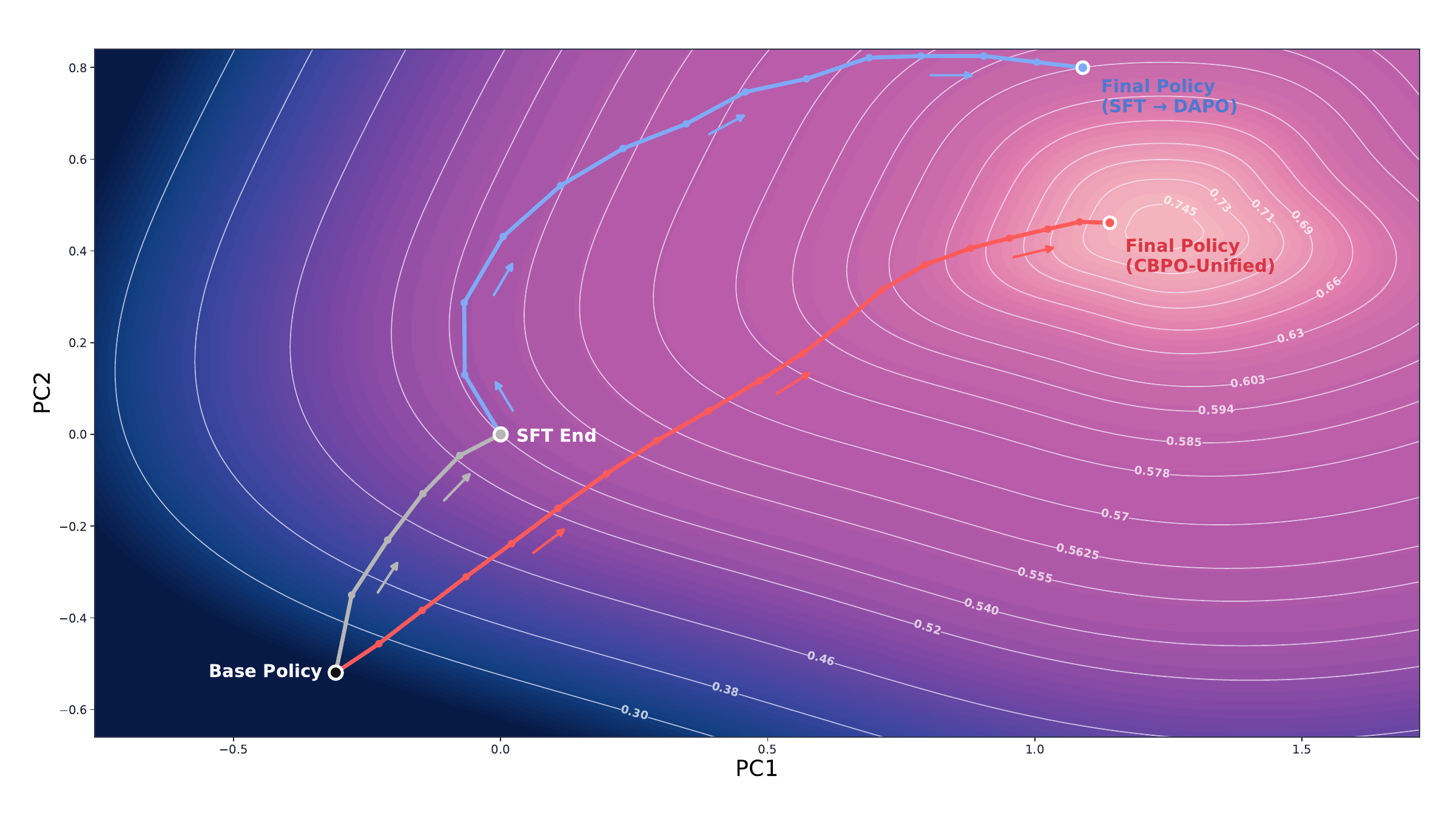}
  \caption{Optimization trajectories with ten PCA components.}
  \label{fig:trajectory_pca10}
\end{figure}

\paragraph{Additional PCA dimensions.}
Figures~\ref{fig:trajectory_pca5} and~\ref{fig:trajectory_pca10} show the trajectories with five and ten PCA components, respectively.
Table~\ref{tab:trajectory_geometry} reports the path geometry for the 2-, 5-, and 10-component projections.

Table~\ref{tab:trajectory_layer_scale} tests whether layer scale materially affects the geometry of the SFT$\rightarrow$DAPO and CBPO-Unified paths.
For the normalized variant, the displacement of the model parameters in each layer from the base policy is divided by the Frobenius norm of that layer's parameters in the base policy. The scale for each layer is fixed across checkpoints and shared by both methods.
Across 2, 5, and 10 PCA components, normalization changes $L/D$ by at most 0.0060 and cumulative turning by at most $2.28^\circ$.

\begin{table}[H]
  \centering
  \small
  \setlength{\tabcolsep}{5pt}
  \caption{Layer-scale sensitivity of the checkpoint trajectories. Each block reports cumulative explained variance for raw and layer-normalized displacements, respectively.}
  \label{tab:trajectory_layer_scale}
  \begin{tabularx}{0.95\linewidth}{@{}>{\raggedright\arraybackslash}X*{4}{>{\centering\arraybackslash}X}@{}}
    \toprule
    \textbf{Method} & \multicolumn{2}{c}{$\boldsymbol{L/D}$} & \multicolumn{2}{c}{\textbf{Cumulative turn}} \\
    \cmidrule(lr){2-3}\cmidrule(lr){4-5}
    & \textbf{Raw} & \textbf{Normalized} & \textbf{Raw} & \textbf{Normalized} \\
    \midrule
    \multicolumn{5}{@{}l}{\textit{2 components: 82.8744\% / 82.8637\%}} \\
    SFT$\rightarrow$DAPO & 1.2207 & 1.2267 & $368.75^\circ$ & $368.96^\circ$ \\
    CBPO-Unified & 1.0205 & 1.0214 & $58.20^\circ$ & $58.96^\circ$ \\
    \midrule
    \multicolumn{5}{@{}l}{\textit{5 components: 91.6378\% / 91.6359\%}} \\
    SFT$\rightarrow$DAPO & 1.1923 & 1.1980 & $295.67^\circ$ & $296.59^\circ$ \\
    CBPO-Unified & 1.0196 & 1.0223 & $69.76^\circ$ & $68.33^\circ$ \\
    \midrule
    \multicolumn{5}{@{}l}{\textit{10 components: 95.9950\% / 95.9948\%}} \\
    SFT$\rightarrow$DAPO & 1.1942 & 1.1999 & $290.89^\circ$ & $290.97^\circ$ \\
    CBPO-Unified & 1.0175 & 1.0191 & $71.22^\circ$ & $73.50^\circ$ \\
    \bottomrule
  \end{tabularx}
\end{table}

\begin{table}[H]
  \centering
  \small
  \setlength{\tabcolsep}{5pt}
  \caption{SFT$\rightarrow$CBPO-OnPolicy versus CBPO-Unified. Cumulative explained variance for raw displacements is 83.4765\%, 92.8952\%, and 96.2445\% for 2, 5, and 10 PCA components, respectively.}
  \label{tab:objective_switch_control}
  \begin{tabularx}{0.95\linewidth}{@{}l*{5}{>{\centering\arraybackslash}X}@{}}
    \toprule
    \multirow[c]{2}{*}{\textbf{Method}} & \multirow[c]{2}{*}{\shortstack{\textbf{PCA}\\\textbf{components}}} & \multirow[c]{2}{*}{$\boldsymbol{L/D}$} & \multirow[c]{2}{*}{\shortstack{\textbf{Cumulative}\\\textbf{turn}}} & \multicolumn{2}{c}{\textbf{Avg.}} \\
    \cmidrule(lr){5-6}
    & & & & \metrichead{avg@32} & \metrichead{pass@32} \\
    \midrule
    \multirow[c]{3}{*}{SFT$\rightarrow$CBPO-OnPolicy} & 2 & 1.1695 & $355.74^\circ$ & \multirow[c]{3}{*}{42.71} & \multirow[c]{3}{*}{65.22} \\
      & 5 & 1.1083 & $285.91^\circ$ & & \\
      & 10 & 1.0847 & $279.33^\circ$ & & \\
    \midrule
    \multirow[c]{3}{*}{CBPO-Unified} & 2 & 1.0205 & $58.20^\circ$ & \multirow[c]{3}{*}{44.56} & \multirow[c]{3}{*}{68.72} \\
      & 5 & 1.0196 & $69.76^\circ$ & & \\
      & 10 & 1.0175 & $71.22^\circ$ & & \\
    \bottomrule
  \end{tabularx}
\end{table}

\subsection{Reward-Temperature Sensitivity}
The CBPO-Unified study varies the reward-temperature coefficient $\beta$ used in the shared regression target.
Table~\ref{tab:beta_ablation} reports aggregate \texttt{avg@32} and \texttt{pass@32} for each value of $\beta$.
The default $\beta=0.2$ row reuses the seed-42 core CBPO-Unified run reported in Table~\ref{tab:replay_horizon}.

\begin{table}[H]
  \centering
  \small
  \setlength{\tabcolsep}{13pt}
  \vspace{-5pt}
  \caption{CBPO-Unified reward-temperature sensitivity for $\beta\in\{0.1,0.2,0.4,0.8\}$.}
  \label{tab:beta_ablation}
  \begin{tabularx}{\linewidth}{@{}*{3}{>{\centering\arraybackslash}X}@{}}
    \toprule
    $\boldsymbol{\beta}$ & \metrichead{avg@32} & \metrichead{pass@32} \\
    \midrule
    0.1 & 41.84 & 66.59 \\
    0.2 & \textbf{44.56} & \textbf{68.72} \\
    0.4 & 43.15 & 67.45 \\
    0.8 & 42.60 & 66.29 \\
    \bottomrule
  \end{tabularx}
  \vspace{-10pt}
\end{table}

\input{reward1_qualitative_appendix_clean.tex}

%% file: reward1_qualitative_appendix_clean.tex

\section{Qualitative Examples}
\label{app:qualitative-examples}

We present two groups of matched-prompt examples illustrating objective continuity and historical replay.
A reward of $1$ means that the rule-based verifier extracted a boxed final answer matching the stored target.
The original response content and formatting markers are preserved.
Each response is accompanied by its training stage or collection round, response index, and prompt hash.
The collection round specifies the sampling iteration.
The response index distinguishes the completions sampled for the same prompt in that round.
Response indices start at zero.
The prompt hash is computed with SHA-1 from the complete input after applying the conversation template, allowing the same formatted prompt to be matched across collection rounds.

\DefineVerbatimEnvironment{SampleText}{Verbatim}{fontsize=\footnotesize,breaklines=true,breakanywhere=true,breaksymbolleft={},breakautoindent=false,breakindent=0pt}

\subsection{Objective Continuity}
\label{app:objective-continuity-examples}

Each example compares a positive offline trajectory of SFT with the responses of the same prompt from the RL stage of the SFT$\rightarrow$DAPO and a response from CBPO-Unified.

\subsubsection{Example 1: Pascal's triangle}

\begin{tcolorbox}[enhanced jigsaw, breakable, title=Problem, colback=white, colframe=black!35, boxrule=0.4pt, arc=1mm, left=1mm, right=1mm, top=1mm, bottom=1mm]
\begin{SampleText}
In the row of Pascal's triangle that starts with 1 and then 10, what is the next number?
\end{SampleText}
\end{tcolorbox}

{\footnotesize\noindent SFT end, response index $0$.\par
\noindent Prompt hash: \nolinkurl{e468f6edb264ab48314fa410921c04ac2ed0f46f}.\par}

\begin{tcolorbox}[enhanced jigsaw, breakable, title=Offline trajectory (verifier reward 1), colback=white, colframe=black!35, boxrule=0.4pt, arc=1mm, left=1mm, right=1mm, top=1mm, bottom=1mm]
\begin{SampleText}
<think>
Okay, let's see. The user is asking about the next number in a row of Pascal's triangle that starts with 1 and then 10. Hmm, I need to recall how Pascal's triangle works. Each row starts and ends with 1, and each number in between is the sum of the two numbers directly above it. 

First, I should figure out which row they're referring to. The problem says the row starts with 1 and then 10. So, the first row is just 1. The second row is 1, 1. The third row is 1, 2, 1. The fourth row is 1, 3, 3, 1. Wait, but the user mentions a row that starts with 1 and then 10. So, maybe they're talking about a specific row where the first two numbers are 1 and 10. 

Wait, but in Pascal's triangle, each row is indexed starting from row 0. So row 0 is 1, row 1 is 1, 1, row 2 is 1, 2, 1, row 3 is 1, 3, 3, 1, and so on. But if the row starts with 1 and then 10, that would be a row where the second element is 10. Let me check. 

In row n, the elements are C(n, 0), C(n, 1), ..., C(n, n). So the second element is C(n, 1) = n. So if the second element is 10, then n must be 10. Therefore, the row is row 10. So the row would be 1, 10, ... and the next number would be C(10, 2) = 45. 

Wait, let me confirm. The first element is C(10, 0) = 1, the second is C(10, 1) = 10, the third is C(10, 2) = 45, and so on. So the next number after 10 is 45. 

But let me make sure I'm not mixing up the row numbers. If the row starts with 1 and then 10, that's row 10. So the numbers are 1, 10, 45, ... So the next number after 10 is 45. 

Alternatively, maybe the user is thinking of a different row. Let me check again. If the row starts with 1 and then 10, that's the second element. So the row is 1, 10, ... So the third element is C(10, 2) = 45. Therefore, the next number is 45. 

I think that's correct. So the answer should be 45. Let me just verify with a small example. For example, in row 3, which is 1, 3, 3, 1. The second element is 3, which is C(3,1) = 3. So yes, if the second element is n, then the row is row n. Therefore, if the second element is 10, the row is row 10, and the next element is C(10,2) = 45. Yep, that makes sense. So the answer is 45.
</think>

To determine the next number in the row of Pascal's triangle that starts with $1$ and then $10$, we begin by understanding how Pascal's triangle is structured.

Each row in Pascal's triangle is defined by the binomial coefficients. Specifically, the $n$-th row (starting from row 0) contains the coefficients $C(n, k)$ for $k = 0$ to $n$. The second element in the row (i.e., $C(n, 1)$) is equal to $n$. 

Given that the row starts with $1$ and then $10$, this means the second element is $10$. Therefore, the row corresponds to $n = 10$. The next number in the row is the third element, which is $C(10, 2)$.

$$
C(10, 2) = \frac{10 \times 9}{2} = 45
$$

Thus, the next number in the row is:

$$
\boxed{45}
$$
\end{SampleText}
\end{tcolorbox}

{\footnotesize\noindent SFT$\rightarrow$DAPO end, response index $0$.\par
\noindent Prompt hash: \nolinkurl{e468f6edb264ab48314fa410921c04ac2ed0f46f}.\par}

\begin{tcolorbox}[enhanced jigsaw, breakable, title=Two-stage RL response (verifier reward 0.1), colback=white, colframe=black!35, boxrule=0.4pt, arc=1mm, left=1mm, right=1mm, top=1mm, bottom=1mm]
\begin{SampleText}
<think>
Okay, let's see. The user is asking about the next number in the row of Pascal's triangle that starts with 1 and then 10. Hmm, first I need to recall how Pascal's triangle works. Each row starts and ends with 1, and each number in between is the sum of the two numbers directly above it. 

So, the first row is just 1. The second row is 1 and 1. The third row is 1, 2, 1. The fourth row is 1, 3, 3, 1. And so on. Now, the user mentioned a row that starts with 1 and then 10. Wait, so the row is 1, 10, ...? But in Pascal's triangle, each row is determined by the row number. For example, row 0 is 1, row 1 is 1, 1, row 2 is 1, 2, 1, row 3 is 1, 3, 3, 1, and so on. 

So, if the row starts with 1 and then 10, that would be row 4? Let me check. Row 0: 1. Row 1: 1, 1. Row 2: 1, 2, 1. Row 3: 1, 3, 3, 1. Row 4: 1, 4, 6, 4, 1. Wait, so row 4 starts with 1, then 4, 6, 4, 1. So if the row starts with 1 and then 10, that would be row 5? Let me check row 5: 1, 5, 10, 10, 5, 1. Oh, right! So the row starts with 1, then 5, 10, 10, 5, 1. Therefore, the next number after 10 would be 5. 

Wait, but the user said the row starts with 1 and then 10. So the first number is 1, then the second is 10. So that's row 5. The numbers in that row are 1, 5, 10, 10, 5, 1. So after 10, the next number is 5. So the answer should be 5. 

But let me make sure I'm not mixing up the rows. Let's think again. The rows are numbered starting from 0. So row 0: 1. Row 1: 1, 1. Row 2: 1, 2, 1. Row 3: 1, 3, 3, 1. Row 4: 1, 4, 6, 4, 1. Row 5: 1, 5, 10, 10, 5, 1. So yes, row 5 starts with 1, then 5, 10, 10, 5, 1. So if the row starts with 1 and then 10, that's row 5. The next number after 10 is 5. Therefore, the answer is 5. 

Alternatively, maybe the user is referring to a different row? But according to the standard Pascal's triangle, each row is determined by the row number. So if the row starts with 1 and then 10, that's row 5. So the next number is 5. So the answer is 5.
</think>

To determine the next number in the row of Pascal's triangle that starts with **1** and then **10**, we need to identify which row this corresponds to.

---

### Understanding Pascal's Triangle Structure

Pascal's triangle is constructed such that:
- Each row starts and ends with **1**.
- Each number in between is the **sum of the two numbers above it**.

The rows are numbered starting from **row 0**. Here's how the first few rows look:

- **Row 0**: 1  
- **Row 1**: 1, 1  
- **Row 2**: 1, 2, 1  
- **Row 3**: 1, 3, 3, 1  
- **Row 4**: 1, 4, 6, 4, 1  
- **Row 5**: 1, 5, 10, 10, 5, 1  

---

### Identifying the Row

The question states that the row starts with **1** and then **10**. This matches **Row 5**:

- **Row 5**: 1, 5, 10, 10, 5, 1

So, the row in question is **Row 5**.

---

### Finding the Next Number

In **Row 5**, the sequence is:
- 1, 5, 10, 10, 5, 1

The numbers are:
- 1 (first number)
- 5 (second number)
- 10 (third number)
- 10 (fourth number)
- 5 (fifth number)
- 1 (sixth number)

Since the row starts with 1 and then 10, the next number is the **fifth number** in the sequence, which is **5**.

---

### Final Answer

$$
\boxed{5}
$$
\end{SampleText}
\end{tcolorbox}

{\footnotesize\noindent SFT$\rightarrow$DAPO end, response index $1$.\par
\noindent Prompt hash: \nolinkurl{e468f6edb264ab48314fa410921c04ac2ed0f46f}.\par}

\begin{tcolorbox}[enhanced jigsaw, breakable, title=Two-stage RL response (verifier reward 0), colback=white, colframe=black!35, boxrule=0.4pt, arc=1mm, left=1mm, right=1mm, top=1mm, bottom=1mm]
\begin{SampleText}
<think>
Okay, let's see. The problem is asking for the next number in the row of Pascal's triangle that starts with 1 and then 10. Hmm, first, I need to recall how Pascal's triangle works. Each row starts and ends with 1, and each number in between is the sum of the two numbers directly above it. 

So, the first row is just 1. The second row is 1 and 1. The third row is 1, 2, 1. The fourth row is 1, 3, 3, 1. And so on. Each row starts and ends with 1, and the numbers in between are combinations. 

Wait, the question mentions a row that starts with 1 and then 10. So, the first number is 1, the second is 10. That must be the fifth row? Let me check. Let's list the rows:

Row 0: 1
Row 1: 1, 1
Row 2: 1, 2, 1
Row 3: 1, 3, 3, 1
Row 4: 1, 4, 6, 4, 1
Row 5: 1, 5, 10, 10, 5, 1
Row 6: 1, 6, 15, 20, 15, 6, 1

Wait, so row 5 is 1, 5, 10, 10, 5, 1. So the row that starts with 1 and then 10 is row 5. But the question says "the row of Pascal's triangle that starts with 1 and then 10". So, that would be row 5. But the next number after 10 would be 5? Wait, no, the row is 1, 5, 10, 10, 5, 1. So after the first 1 and then 5, the next number is 10. Wait, the question says "the row of Pascal's triangle that starts with 1 and then 10". Wait, that's not possible because the first number is 1, then the next number is 10. So maybe the row is row 5? Let me check again.

Wait, row 4 is 1, 4, 6, 4, 1. Row 5 is 1, 5, 10, 10, 5, 1. So the row that starts with 1 and then 10 is row 5. So the numbers are 1, 5, 10, 10, 5, 1. So the next number after 10 would be 5. But the question is asking for the next number in the row that starts with 1 and then 10. So the row is 1, 5, 10, 10, 5, 1. So after 10, the next number is 5. So the answer is 5. But wait, maybe the question is phrased differently. Let me check again.

Wait, the user says "the row of Pascal's triangle that starts with 1 and then 10". So the first number is 1, then the next number is 10. So that's the row where the second number is 10. So that would be row 5. So the numbers are 1, 5, 10, 10, 5, 1. So the next number after 10 would be 5. Therefore, the answer is 5. But let me confirm.

Alternatively, maybe the user is referring to the row where the numbers are 1, 10, ... but that would be row 5, but the first number is 1, then 10. Wait, row 5 is 1, 5, 10, 10, 5, 1. So the row that starts with 1 and then 10 is row 5. So the numbers are 1, 5, 10, 10, 5, 1. So the next number after 10 is 5. So the answer is 5. Therefore, the next number is 5.

Alternatively, maybe the user is thinking of the row where the numbers are 1, 10, ... but that would be row 5? Wait, no. Let's think about the rows again. The first row is row 0: 1. Row 1: 1, 1. Row 2: 1, 2, 1. Row 3: 1, 3, 3, 1. Row 4: 1, 4, 6, 4, 1. Row 5: 1, 5, 10, 10, 5, 1. So the row starting with 1 and then 10 is row 5. So the numbers are 1, 5, 10, 10, 5, 1. So after 10, the next number is 5. Therefore, the answer is 5.

But maybe the user is confused about the row. Let me check if there's another row. For example, if the row starts with 1 and then 10, that would be row 5. So the next number is 5. Therefore, the answer is 5. So I think the answer is 5.
\end{SampleText}
\end{tcolorbox}

{\footnotesize\noindent Collection round $320$, response index $3$.\par
\noindent Prompt hash: \nolinkurl{e468f6edb264ab48314fa410921c04ac2ed0f46f}.\par}

\begin{tcolorbox}[enhanced jigsaw, breakable, title=CBPO-Unified response (verifier reward 1), colback=white, colframe=black!35, boxrule=0.4pt, arc=1mm, left=1mm, right=1mm, top=1mm, bottom=1mm]
\begin{SampleText}
<think>
Okay, let's see. The question is about Pascal's triangle. The row starts with 1 and then 10. I need to find the next number in that row. Hmm, Pascal's triangle is built such that each number is the sum of the two numbers directly above it. But first, I should figure out which row this is.

Wait, Pascal's triangle starts with row 0. Row 0 is just [1]. Row 1 is [1, 1]. Row 2 is [1, 2, 1], and so on. So the row that starts with 1 and then 10... Let me think. The first row is [1], then the second row is [1, 1], third is [1, 2, 1], fourth is [1, 3, 3, 1], and so on. So if a row starts with 1 and then 10, that must be a higher row. Let me check the pattern.

In row n, the numbers are combinations of n choose k, where k goes from 0 to n. So the first number is C(n, 0) = 1, the second is C(n, 1) = n, the third is C(n, 2) = n(n-1)/2, and so on. So if the row starts with 1 and then 10, that means the second number is 10. So C(n, 1) = 10. Therefore, n = 10. So the row is row 10. 

Now, the next number after 10 in that row would be the third number, which is C(10, 2). Let me calculate that. C(10, 2) is 10*9/2 = 45. So the next number after 10 is 45. Wait, but let me double-check. The row starts with 1, then 10, then the next number is the third number, which is 45. So the answer should be 45. 

Alternatively, maybe I should verify the row structure. Let me think. If the row is [1, 10, ...], then the third number is the middle one. For row 10, the numbers are 1, 10, 45, 120, etc. So yes, the next number after 10 is 45. 

Another way: in Pascal's triangle, each row is the coefficients of the binomial expansion. So row 10 would be the coefficients of (a + b)^10. The coefficients are 1, 10, 45, 120, etc. So the next number after 10 is indeed 45. Therefore, the answer is 45.
</think>

To determine the next number in the row of Pascal's triangle that starts with $1$ and then $10$, we need to identify which row this is and then find the corresponding number in that row.

---

### Step 1: Identify the Row

In Pascal's triangle, each row is labeled by its index, starting from 0. The first row (row 0) is:
$$
[1]
$$
The second row (row 1) is:
$$
[1, 1]
$$
The third row (row 2) is:
$$
[1, 2, 1]
$$
And so on. The general form of the $n$-th row is:
$$
\text{Row } n = [ \binom{n}{0}, \binom{n}{1}, \binom{n}{2}, \dots, \binom{n}{n} ]
$$

We are told the row starts with $1$ and then $10$. That implies the second number is $10$, which corresponds to:
$$
\binom{n}{1} = 10 \Rightarrow n = 10
$$

So, the row is **row 10**.

---

### Step 2: Find the Next Number in the Row

The row is:
$$
[ \binom{10}{0}, \binom{10}{1}, \binom{10}{2}, \dots ]
$$

We already know the first two numbers are:
$$
\binom{10}{0} = 1, \quad \binom{10}{1} = 10
$$

The next number, which is the third number in the row, is:
$$
\binom{10}{2} = \frac{10 \cdot 9}{2} = 45
$$

---

### Final Answer

$$
\boxed{45}
$$
\end{SampleText}
\end{tcolorbox}

\subsubsection{Example 2: Mode of a stem-and-leaf plot}

\begin{tcolorbox}[enhanced jigsaw, breakable, title=Problem, colback=white, colframe=black!35, boxrule=0.4pt, arc=1mm, left=1mm, right=1mm, top=1mm, bottom=1mm]
\begin{SampleText}
The scores on a $110$-point test were organized in the stem-and-leaf plot shown. $9 | 6$ represents $96$ points. What is the mode of the scores? \begin{tabular}{c|lllllll}
\multicolumn{8}{c}{\underline{Points on the Test}}\\
5 &0 & 0 & & & & &\\
6 &3 & & & & & &\\
7 &7 & 8 & & & & &\\
8 &2 & 6 & 7 & 9 & 9 & 9 & 9\\
9 &1 & 4 & 4 & 4 & 6 & &\\
10 &0 & 0 & 0 & & & &\\
\end{tabular}
\end{SampleText}
\end{tcolorbox}

{\footnotesize\noindent SFT end, response index $1$.\par
\noindent Prompt hash: \nolinkurl{157af75b5df58c00c4e0ebc10d935711cd14b63e}.\par}

\begin{tcolorbox}[enhanced jigsaw, breakable, title=Offline trajectory (verifier reward 1), colback=white, colframe=black!35, boxrule=0.4pt, arc=1mm, left=1mm, right=1mm, top=1mm, bottom=1mm]
\begin{SampleText}
<think>
Okay, let's try to figure out the mode of the scores from this stem-and-leaf plot. First, I need to understand how the stem-and-leaf plot is structured. The problem says that 9 | 6 represents 96 points. So, the stem is the first digit(s), and the leaf is the last digit. 

Looking at the table, the rows are for different points. Let me break down each row:

1. The first row is 5 | 0 0. So that would be 50 and 50. So two scores of 50.
2. Next, 6 | 3. So that's 63. So one score of 63.
3. Then 7 | 7 8. So 77 and 78. Two scores of 77 and 78.
4. The next row is 8 | 2 6 7 9 9 9 9. So that's 82, 86, 87, 89, 89, 89, 89. So seven scores here: 82, 86, 87, and four 89s.
5. Then 9 | 1 4 4 4 6. So that's 91, 94, 94, 94, 96. Five scores here: 91, three 94s, and 96.
6. Finally, 10 | 0 0 0. So that's 100, 100, 100. Three scores of 100.

Now, I need to list out all the scores to see which ones appear most frequently. Let me list them all:

From row 5 (stem 5): 50, 50  
Row 6 (stem 6): 63  
Row 7 (stem 7): 77, 78  
Row 8 (stem 8): 82, 86, 87, 89, 89, 89, 89  
Row 9 (stem 9): 91, 94, 94, 94, 96  
Row 10 (stem 10): 100, 100, 100  

Now let's count the frequencies:

- 50: 2 times  
- 63: 1 time  
- 77: 1 time  
- 78: 1 time  
- 82: 1 time  
- 86: 1 time  
- 87: 1 time  
- 89: 4 times  
- 91: 1 time  
- 94: 3 times  
- 96: 1 time  
- 100: 3 times  

Looking at these counts, the scores with the highest frequency are 89 (4 times), 100 (3 times), 94 (3 times), and 50 (2 times). Wait, but the question asks for the mode, which is the most frequently occurring score. So the mode here would be the score that appears the most. 

Looking at the counts: 89 appears 4 times, which is more than 3 (for 100, 94, 50). So the mode is 89. 

Wait, but let me double-check. Are there any other scores with the same highest frequency? 100 and 94 both have 3 times. But 89 has 4 times, so 89 is the mode. 

Therefore, the mode of the scores is 89. 

**Final Answer**
The mode of the scores is \boxed{89}.
</think>

To determine the **mode** of the test scores from the given stem-and-leaf plot, we first need to understand the structure of the plot. The stem-and-leaf plot is organized such that each row represents a stem (the first digit or digits of the score) and the leaves (the last digit) represent the individual scores.

From the table:

- **Row 5 (stem 5):** 50, 50 → 2 scores
- **Row 6 (stem 6):** 63 → 1 score
- **Row 7 (stem 7):** 77, 78 → 2 scores
- **Row 8 (stem 8):** 82, 86, 87, 89, 89, 89, 89 → 7 scores (1 of each, except 89 appears 4 times)
- **Row 9 (stem 9):** 91, 94, 94, 94, 96 → 5 scores (3 of 94)
- **Row 10 (stem 10):** 100, 100, 100 → 3 scores

Now, we count the frequency of each score:

- 50: 2 times  
- 63: 1 time  
- 77: 1 time  
- 78: 1 time  
- 82: 1 time  
- 86: 1 time  
- 87: 1 time  
- 89: 4 times  
- 91: 1 time  
- 94: 3 times  
- 96: 1 time  
- 100: 3 times  

The **mode** is the score that appears most frequently. Among the listed scores, **89** appears **4 times**, which is the highest frequency. Therefore, the mode of the scores is:

$$
\boxed{89}
$$
\end{SampleText}
\end{tcolorbox}

{\footnotesize\noindent SFT$\rightarrow$DAPO end, response index $0$.\par
\noindent Prompt hash: \nolinkurl{157af75b5df58c00c4e0ebc10d935711cd14b63e}.\par}

\begin{tcolorbox}[enhanced jigsaw, breakable, title=Two-stage RL response (verifier reward 0.1), colback=white, colframe=black!35, boxrule=0.4pt, arc=1mm, left=1mm, right=1mm, top=1mm, bottom=1mm]
\begin{SampleText}
<think>
Okay, let's try to figure out the mode of the test scores from the stem-and-leaf plot. First, I need to understand how the stem-and-leaf plot is structured. The problem says that 9 | 6 represents 96 points. So, the stems are the tens place and the leaves are the ones place. 

Let me start by breaking down each row of the plot. 

The first row is 5 | 0 0. That should be 50 and 50, right? So two scores of 50.

Next row is 6 | 3. Since the stem is 6, the leaves are 3. So that's 63. But wait, the problem says that 9 | 6 is 96, so maybe the leaves are single digits. So, 6 | 3 would be 63. So that's one score of 63.

Third row is 7 | 7 8. So stem 7, leaves 7 and 8. That would be 77 and 78. So two scores here: 77 and 78.

Fourth row is 8 | 2 6 7 9 9 9 9. So stem 8, leaves 2, 6, 7, 9, 9, 9, 9. So that would be 82, 86, 87, 99, 99, 99, 99. So seven scores here: 82, 86, 87, and four 99s.

Fifth row is 9 | 1 4 4 4 6. Stem 9, leaves 1, 4, 4, 4, 6. So that's 91, 94, 94, 94, 96. So five scores: 91, 94, 94, 94, 96.

Sixth row is 10 | 0 0 0. Stem 10, leaves 0, 0, 0. So that's 100, 100, 100. Three scores of 100.

Now, let me list all the scores:

From 5: 50, 50

From 6: 63

From 7: 77, 78

From 8: 82, 86, 87, 99, 99, 99, 99

From 9: 91, 94, 94, 94, 96

From 10: 100, 100, 100

Now, to find the mode, which is the number that appears most frequently. Let me count each number:

50: 2 times

63: 1 time

77: 1 time

78: 1 time

82: 1 time

86: 1 time

87: 1 time

99: 4 times

91: 1 time

94: 3 times

96: 1 time

100: 3 times

So looking at the counts:

- 50: 2
- 63: 1
- 77: 1
- 78: 1
- 82: 1
- 86: 1
- 87: 1
- 99: 4
- 91: 1
- 94: 3
- 96: 1
- 100: 3

The highest frequency is 4 for 99, then 3 for 94 and 100, and 2 for 50. So the mode is 99. Wait, but let me check if there are any other numbers with the same frequency. The 94 and 100 both have 3, but 99 has 4, so 99 is the mode. 

But let me double-check the counts. For 99, there are four instances: in the 8 row, there are four 99s. Then in the 9 row, there's a 96, but 99 isn't there. So yes, 99 appears four times. The 94 appears three times: in the 9 row, three 94s. And 100 appears three times. So 99 is the mode with the highest frequency. 

Therefore, the mode is 99. 

But wait, let me check again. The 8 row has 82, 86, 87, and then four 99s. So 99 is there four times. The 9 row has 91, 94, 94, 94, 96. So 94 is three times. And 100 is three times. So 99 is the only one with four. So yes, the mode is 99. 

I think that's correct. So the answer should be \boxed{99}.
</think>

To determine the **mode** of the test scores from the given stem-and-leaf plot, we first need to interpret the plot and extract the individual scores.

---

### **Step 1: Interpreting the Stem-and-Leaf Plot**

Each row in the stem-and-leaf plot represents a "stem" (the tens digit) and "leaves" (the ones digit). The example given is:

- **9 | 6** represents **96**.

So, the breakdown of the scores is:

- **5 | 0 0** → 50, 50  
- **6 | 3** → 63  
- **7 | 7 8** → 77, 78  
- **8 | 2 6 7 9 9 9 9** → 82, 86, 87, 99, 99, 99, 99  
- **9 | 1 4 4 4 6** → 91, 94, 94, 94, 96  
- **10 | 0 0 0** → 100, 100, 100  

---

### **Step 2: Counting Frequency of Each Score**

We now count how many times each score appears:

- 50: 2 times  
- 63: 1 time  
- 77: 1 time  
- 78: 1 time  
- 82: 1 time  
- 86: 1 time  
- 87: 1 time  
- 99: 4 times  
- 91: 1 time  
- 94: 3 times  
- 96: 1 time  
- 100: 3 times  

---

### **Step 3: Identifying the Mode**

The **mode** is the number that appears most frequently. Among all the scores, **99** appears **4 times**, which is the highest frequency. No other number appears more than 3 times.

---

### **Final Answer**

$$
\boxed{99}
$$
\end{SampleText}
\end{tcolorbox}

{\footnotesize\noindent SFT$\rightarrow$DAPO end, response index $2$.\par
\noindent Prompt hash: \nolinkurl{157af75b5df58c00c4e0ebc10d935711cd14b63e}.\par}

\begin{tcolorbox}[enhanced jigsaw, breakable, title=Two-stage RL response (verifier reward 0), colback=white, colframe=black!35, boxrule=0.4pt, arc=1mm, left=1mm, right=1mm, top=1mm, bottom=1mm]
\begin{SampleText}
<think>
Okay, let's see. I need to find the mode of the scores from the stem-and-leaf plot. First, I should remember what a mode is. The mode is the number that appears most frequently in a data set. So, I need to look through all the scores and count how many times each number appears. 

The stem-and-leaf plot is given with different stems and leaves. Let me parse it step by step. 

Starting with the first row: 5 | 0 0. That means the stem is 5, and the leaves are 0 and 0. So the scores are 50 and 50. So there are two 50s here.

Next row: 6 | 3. The stem is 6, and the leaf is 3. So that's one score: 63.

Third row: 7 | 7 8. Stem 7, leaves 7 and 8. So scores are 77 and 78.

Fourth row: 8 | 2 6 7 9 9 9 9. Stem 8, leaves 2, 6, 7, 9, 9, 9, 9. So the scores are 82, 86, 87, 99, 99, 99, 99. Wait, let me check that again. The leaves are 2, 6, 7, 9, 9, 9, 9. So that's 82, 86, 87, and then four 9s. So that's four 9s here.

Fifth row: 9 | 1 4 4 4 6. Stem 9, leaves 1, 4, 4, 4, 6. So the scores are 91, 94, 94, 94, 96. So here, there are three 94s and one 91 and one 96.

Sixth row: 10 | 0 0 0. Stem 10, leaves 0, 0, 0. So the scores are 100, 100, 100. Three 100s.

Now, let me list all the scores:

From 5: 50, 50

From 6: 63

From 7: 77, 78

From 8: 82, 86, 87, 99, 99, 99, 99

From 9: 91, 94, 94, 94, 96

From 10: 100, 100, 100

Now, let me count each number's frequency:

50: 2

63: 1

77: 1

78: 1

82: 1

86: 1

87: 1

99: 4

91: 1

94: 3

96: 1

100: 3

So the counts are:

50: 2

63: 1

77: 1

78: 1

82: 1

86: 1

87: 1

99: 4

91: 1

94: 3

96: 1

100: 3

Now, the mode is the number with the highest count. Looking at these counts, 99 appears 4 times, which is the highest. Then 100 appears 3 times, 94 appears 3 times, and others have 1 or 2. So the mode is 99.

Wait, but let me double-check. Are there any other numbers with the same highest count? 99 is 4, 100 is 3, 94 is 3. So 99 is the only one with 4. So yes, the mode is 99.

But hold on, the problem says "the scores on a 110-point test". The maximum score is 100, but the stem-and-leaf plot includes up to 100. But the mode is 99, which is present four times. So that's the mode.

I think that's correct. Let me check again.

Yes, 99 appears four times, and no other number appears four times. So the mode is 99.
\end{SampleText}
\end{tcolorbox}

{\footnotesize\noindent Collection round $320$, response index $0$.\par
\noindent Prompt hash: \nolinkurl{157af75b5df58c00c4e0ebc10d935711cd14b63e}.\par}

\begin{tcolorbox}[enhanced jigsaw, breakable, title=CBPO-Unified response (verifier reward 1), colback=white, colframe=black!35, boxrule=0.4pt, arc=1mm, left=1mm, right=1mm, top=1mm, bottom=1mm]
\begin{SampleText}
<think>
Okay, let's see. I need to find the mode of the scores from the stem-and-leaf plot. First, I should remember what a mode is. The mode is the number that appears most frequently in a data set. So, I need to look at all the scores and count how many times each number appears.

The stem-and-leaf plot is given with different stems and leaves. Let me parse it step by step. The first row is 5 | 0 0. So that would be 50 and 50. Then the next row is 6 | 3. So that's 63. Then 7 | 7 8, which is 77 and 78. Then 8 | 2 6 7 9 9 9 9. So that's 82, 86, 87, 89, 89, 89, 89. Then 9 | 1 4 4 4 6. So 91, 94, 94, 94, 96. Finally, 10 | 0 0 0. So 100, 100, 100.

Let me list out all the scores:

From 5: 50, 50

From 6: 63

From 7: 77, 78

From 8: 82, 86, 87, 89, 89, 89, 89

From 9: 91, 94, 94, 94, 96

From 10: 100, 100, 100

Now, let's count each score's frequency.

Starting with 50: appears twice.

63: once.

77: once.

78: once.

82: once.

86: once.

87: once.

89: four times (since there are four leaves: 89, 89, 89, 89).

91: once.

94: three times (since there are three leaves: 94, 94, 94).

96: once.

100: three times (since there are three leaves: 100, 100, 100).

So now, let's list the counts:

- 50: 2
- 63: 1
- 77: 1
- 78: 1
- 82: 1
- 86: 1
- 87: 1
- 89: 4
- 91: 1
- 94: 3
- 96: 1
- 100: 3

Looking at these counts, the highest frequency is 4, which is for the number 89. So the mode is 89.

Wait, but let me double-check. Are there any other numbers with the same frequency? Let me check again:

- 50: 2
- 63: 1
- 77: 1
- 78: 1
- 82: 1
- 86: 1
- 87: 1
- 89: 4
- 91: 1
- 94: 3
- 96: 1
- 100: 3

Yes, 89 is the only one with 4. So the mode is 89. Therefore, the answer should be 89.

**Final Answer**
The mode of the scores is \boxed{89}.
</think>

To determine the **mode** of the scores from the given stem-and-leaf plot, we first need to interpret the data correctly.

---

### Step 1: Understand the Stem-and-Leaf Plot

The plot is organized as follows:

- **5 | 0 0** → 50, 50
- **6 | 3** → 63
- **7 | 7 8** → 77, 78
- **8 | 2 6 7 9 9 9 9** → 82, 86, 87, 89, 89, 89, 89
- **9 | 1 4 4 4 6** → 91, 94, 94, 94, 96
- **10 | 0 0 0** → 100, 100, 100

---

### Step 2: List All the Scores

From the above, the scores are:

- 50, 50  
- 63  
- 77, 78  
- 82, 86, 87, 89, 89, 89, 89  
- 91, 94, 94, 94, 96  
- 100, 100, 100

---

### Step 3: Count the Frequency of Each Score

Now, let's count how many times each score appears:

| Score | Frequency |
|-------|-----------|
| 50    | 2         |
| 63    | 1         |
| 77    | 1         |
| 78    | 1         |
| 82    | 1         |
| 86    | 1         |
| 87    | 1         |
| 89    | 4         |
| 91    | 1         |
| 94    | 3         |
| 96    | 1         |
| 100   | 3         |

---

### Step 4: Identify the Mode

The **mode** is the score that appears most frequently. From the table above, the score **89** appears **4 times**, which is the highest frequency.

---

### Final Answer

$$
\boxed{89}
$$
\end{SampleText}
\end{tcolorbox}

\subsection{Historical Replay}
\label{app:historical-replay-examples}

For each prompt, archived CBPO-Unified responses are displayed in increasing order of collection round.

\subsubsection{Example 1: Sibling counting}

\begin{tcolorbox}[enhanced jigsaw, breakable, title=Problem, colback=white, colframe=black!35, boxrule=0.4pt, arc=1mm, left=1mm, right=1mm, top=1mm, bottom=1mm]
\begin{SampleText}
Harry has 3 sisters and 5 brothers.  His sister Harriet has $\text{S}$ sisters and $\text{B}$ brothers.  What is the product of $\text{S}$ and $\text{B}$?
$\text{(A)}\ 8 \qquad \text{(B)}\ 10 \qquad \text{(C)}\ 12 \qquad \text{(D)}\ 15 \qquad \text{(E)}\ 18$
\end{SampleText}
\end{tcolorbox}

{\footnotesize\noindent Collection round $92$, response index $1$.\par
\noindent Prompt hash: \nolinkurl{c52eaec86d2a58f85ae62c76bf34e560283c2ebc}.\par}

\begin{tcolorbox}[enhanced jigsaw, breakable, title=Model response (verifier reward 0), colback=white, colframe=black!35, boxrule=0.4pt, arc=1mm, left=1mm, right=1mm, top=1mm, bottom=1mm]
\begin{SampleText}
<think>
Okay, let me try to figure out this problem. So, Harry has 3 sisters and 5 brothers. Then his sister Harriet has S sisters and B brothers. We need to find the product of S and B. The options are given, so maybe I can figure out S and B first.

First, let's think about Harry's sisters. Harry has 3 sisters. But wait, are those sisters including Harriet? Because Harriet is his sister, so if Harry has 3 sisters, that includes Harriet. So, if Harry has 3 sisters, then Harriet is one of them. So, Harriet's sisters would be the other 2 sisters. So S, which is the number of sisters Harriet has, would be 2. Wait, but let me make sure.

Wait, the problem says Harry has 3 sisters. So, that's Harry's sisters. So, Harry's sisters are 3 in total. Since Harriet is one of them, then Harriet has 3 - 1 = 2 sisters. So S is 2. That seems right.

Now, for the brothers. Harry has 5 brothers. So, the number of brothers Harry has is 5. But Harriet is his sister, so she isn't counted as a brother. So, Harriet's brothers would be the same as Harry's brothers. So B is 5. Therefore, the product S * B is 2 * 5 = 10. So the answer should be 10, which is option B.

Wait, but let me check again. Let me think differently. Maybe I need to consider the number of siblings in general.

Harry has 3 sisters and 5 brothers. So, total siblings for Harry would be 3 sisters + 5 brothers = 8 siblings. But Harry is one of them, so total people in the family would be 8 + 1 = 9? Wait, no. Wait, Harry is a boy, and his sisters are girls. So, if Harry has 3 sisters, that's 3 girls. His brothers are 5, so 5 boys. So total people in the family would be 3 girls + 5 boys + Harry = 9 people. But Harry is a boy, so the total is 3 + 5 + 1 = 9. But maybe that's not necessary here.

Now, Harriet is Harry's sister. So Harriet is one of the 3 sisters. Therefore, Harriet has 2 sisters (since there are 3 sisters in total, and she's one of them). So S = 2. Then, Harriet's brothers would be the 5 brothers Harry has. Because Harriet is a girl, her brothers are the same as Harry's brothers. So B = 5. Therefore, S * B = 2 * 5 = 10. So the answer is B. 

But let me check if there's any other way this could be interpreted. For example, if the problem was about Harriet's siblings, but maybe there's a different way to count. Suppose Harriet has S sisters. If Harriet is a girl, then her sisters are the other girls in the family. Since Harry has 3 sisters, that includes Harriet. So the number of sisters Harriet has is 3 - 1 = 2. So S = 2. 

For brothers, Harriet is a girl, so her brothers are the boys in the family. Harry has 5 brothers, so Harriet has 5 brothers. Therefore, B = 5. So 2 * 5 = 10. So the answer is B. 

Alternatively, maybe someone might think that Harriet's brothers are Harry's brothers minus someone else? But no, because Harriet is a sister, so her brothers are all the boys in the family. Since Harry has 5 brothers, that's all of them. So B is 5. 

So yes, the product is 10. So the answer is B.
\end{SampleText}
\end{tcolorbox}

{\footnotesize\noindent Collection round $199$, response index $3$.\par
\noindent Prompt hash: \nolinkurl{c52eaec86d2a58f85ae62c76bf34e560283c2ebc}.\par}

\begin{tcolorbox}[enhanced jigsaw, breakable, title=Model response (verifier reward 0.1), colback=white, colframe=black!35, boxrule=0.4pt, arc=1mm, left=1mm, right=1mm, top=1mm, bottom=1mm]
\begin{SampleText}
<think>
Okay, let's try to solve this problem step by step. So, the question is about Harry's sisters and brothers, and we need to find the product of the number of sisters Harriet has (S) and the number of brothers she has (B). The options are given, so maybe we can figure it out by reasoning through the relationships.

First, let's parse the information given. Harry has 3 sisters and 5 brothers. Wait, so Harry himself is a boy, right? Because he has sisters and brothers. So, if Harry has 3 sisters, that means there are 3 sisters in the family. And he has 5 brothers, which would mean there are 5 brothers in the family. But wait, Harry is one of those brothers, so the total number of boys in the family would be Harry plus his brothers. So, 5 brothers means there are 5 boys, including Harry. So, the total number of boys is 5, and the number of girls is 3. So, the family has 3 girls and 5 boys.

Now, the question is about Harriet. Harriet is one of Harry's sisters. So, Harriet is a girl. The problem says that Harriet has S sisters and B brothers. Let's think about that.

First, S is the number of sisters Harriet has. Since Harriet is a girl, her sisters would be all the other girls in the family. The total number of girls in the family is 3, so if Harriet is one of them, then the number of sisters she has is 3 - 1 = 2. Wait, because she's one sister, so the rest are her sisters. So S = 2.

Then, B is the number of brothers Harriet has. Harriet is a girl, so her brothers are all the boys in the family. The total number of boys is 5, so B = 5.

Therefore, the product S * B is 2 * 5 = 10. So, the answer should be 10, which is option B.

But let me double-check to make sure I didn't make a mistake. Let's verify again.

Harry has 3 sisters. So the total number of girls in the family is 3. Harriet is one of them, so she has 3 - 1 = 2 sisters. That seems right. Then, Harry has 5 brothers, so the total number of boys is 5. Harriet is a girl, so she has 5 brothers. That's correct. So S = 2 and B = 5. Multiplying them gives 10. So the answer is B.

Wait, but let me check if there's another way this could be interpreted. For example, is Harriet's number of brothers including Harry or not? But no, because Harry is a brother to Harriet. So if Harriet has brothers, they include Harry and all the other boys. Since there are 5 boys, including Harry, then Harriet has 5 brothers. So B is definitely 5. Therefore, the product is 2*5=10. Yep, that seems correct. So the answer should be B.

Another way to think about it: if Harry has 3 sisters, then each of his sisters is a girl, so Harriet is one of them. So her sisters are 3 - 1 = 2. Then, Harry has 5 brothers, so Harriet's brothers are 5. So S*B = 2*5=10. Yep, that's the same result. So I think that's solid. The answer is B.
</think>

To solve the problem, we begin by analyzing the information given:

---

### Step 1: Determine the number of girls and boys in the family

- **Harry has 3 sisters.** This means there are **3 girls** in the family.
- **Harry has 5 brothers.** Since Harry is a boy, this means there are **5 boys** in the family.

So, the family has:
- **3 girls**
- **5 boys**

---

### Step 2: Analyze Harriet's siblings

- **Harriet is one of Harry's sisters**, so she is a **girl**.
- **Number of sisters Harriet has (S):**  
  Since there are 3 girls in total and Harriet is one of them, the number of sisters Harriet has is $ 3 - 1 = 2 $.

- **Number of brothers Harriet has (B):**  
  Harriet is a girl, so her brothers are the **5 boys** in the family. Thus, $ B = 5 $.

---

### Step 3: Compute the product $ S \times B $

$$
S \times B = 2 \times 5 = 10
$$

---

### Final Answer

$$
\boxed{10}
$$
\end{SampleText}
\end{tcolorbox}

{\footnotesize\noindent Collection round $320$, response index $2$.\par
\noindent Prompt hash: \nolinkurl{c52eaec86d2a58f85ae62c76bf34e560283c2ebc}.\par}

\begin{tcolorbox}[enhanced jigsaw, breakable, title=Model response (verifier reward 1), colback=white, colframe=black!35, boxrule=0.4pt, arc=1mm, left=1mm, right=1mm, top=1mm, bottom=1mm]
\begin{SampleText}
<think>
Okay, let's try to solve this problem step by step. So, the question is about Harry's sisters and brothers, and we need to find the product of the number of sisters Harry has (S) and the number of brothers he has (B). The options are given, so maybe we can figure it out by reasoning through the information provided.

First, let's parse the information given. Harry has 3 sisters and 5 brothers. Wait, hold on. So, Harry is a boy, right? Because he has brothers. So, if Harry has 5 brothers, that means there are 5 boys who are his brothers. But then, how many sisters does Harry have? The problem says Harry has 3 sisters. But wait, sisters are female, so if Harry is a boy, his sisters would be the girls in his family. But the problem states that Harry has 3 sisters and 5 brothers. So, the sisters are part of his family, and the brothers are also part of his family.

But then, the question is about Harriet, who is Harry's sister. So, Harriet has S sisters and B brothers. Wait, but we need to figure out S and B for Harriet, then multiply them. Let me think.

First, let's figure out how many sisters Harry has. The problem says Harry has 3 sisters. So, that means there are 3 sisters in the family. But if Harry is a boy, then his sisters are the girls in the family. So, the total number of girls in the family is 3. Therefore, Harriet, being one of those sisters, would have 3 - 1 = 2 sisters? Wait, no. Wait, if Harry has 3 sisters, that means there are 3 sisters in total. So, if Harriet is one of them, then Harriet's sisters would be the other 2 sisters. So, S, which is the number of sisters Harriet has, would be 2? Because Harriet is a sister, so the number of sisters she has is total sisters minus 1 (herself). So, 3 sisters total, so Harriet has 3 - 1 = 2 sisters.

But wait, the problem says "Harriet has S sisters and B brothers." So, S is the number of sisters Harriet has. So, if there are 3 sisters in total, then Harriet has 2 sisters. So S = 2.

Now, what about B, the number of brothers Harriet has? Well, Harriet is a girl, so her brothers would be the boys in the family. Harry has 5 brothers. But wait, Harry is a boy, so the total number of boys in the family is 5 (Harry's brothers) plus himself? Wait, no. Wait, the problem says Harry has 5 brothers. So, that means Harry has 5 brothers, which would be other boys in the family. So, if Harry is a boy, then the total number of boys in the family is Harry plus his 5 brothers, which is 6 boys. Therefore, Harriet, being a girl, would have 6 brothers. So, B = 6.

Therefore, the product S * B is 2 * 6 = 12. So, the answer should be 12, which is option C. Let me check again to make sure I didn't make a mistake.

Wait, let me verify. Harry has 3 sisters and 5 brothers. So, total number of girls in the family is 3 (since Harry's sisters are the girls). So, Harriet is one of those 3 girls. Therefore, Harriet's sisters are 3 - 1 = 2. So S = 2.

For Harriet's brothers, since Harry has 5 brothers, and Harry is a boy, the total number of boys in the family is 5 (Harry's brothers) plus Harry himself, which is 6. Therefore, Harriet, being a girl, has 6 brothers. Therefore, B = 6. So, 2 * 6 = 12. That's option C. 

But let me check if there's another way to approach this. Maybe considering the family members. Let's think about the family composition. If Harry has 3 sisters and 5 brothers, then the total number of siblings is 3 sisters + 5 brothers + Harry himself? Wait, no. Wait, Harry is a boy, so his brothers are the other boys. So, if Harry has 5 brothers, that means there are 5 boys besides Harry. So, total number of boys is 1 (Harry) + 5 (brothers) = 6. The number of girls is 3 (Harry's sisters). Therefore, the total family members are 6 boys + 3 girls = 9 people.

Now, Harriet is one of the girls. So, Harriet's sisters are the other 2 girls. Therefore, S = 2. Harriet's brothers are all the boys, which is 6. Therefore, B = 6. So, 2 * 6 = 12. Yep, same answer.

Alternatively, maybe someone might think that Harry's brothers are the same as Harriet's brothers. But since Harriet is a girl, her brothers are all the boys in the family. So, Harry's brothers are part of that. So, yes, B is 6.

Another way: If Harry has 3 sisters, then Harriet is one of them, so Harriet has 2 sisters. If Harry has 5 brothers, that's 5 boys, including himself? Wait, no. Wait, Harry is a boy, so his brothers are the other boys. So, if Harry has 5 brothers, that means there are 5 boys besides Harry. Therefore, total boys are 6. So Harriet, being a girl, has 6 brothers. Therefore, S = 2, B = 6. Product is 12. So, answer is C.

I think that's solid. The answer should be \boxed{C}.
</think>

To solve the problem, we need to determine the product of the number of sisters Harriet has ($S$) and the number of brothers Harriet has ($B$).

---

### Step 1: Understand the Given Information

We are told:
- **Harry has 3 sisters** and **5 brothers**.

From this, we can infer:
- Harry is a **boy**, so the **brothers** he has are other boys in the family.
- **Harry's sisters** are the **girls** in the family, and there are **3** of them.

---

### Step 2: Determine the Number of Sisters Harriet Has ($S$)

Harriet is **one of Harry's sisters**, so there are:
- **3 sisters** in total (including Harriet).
- Therefore, **Harriet has $3 - 1 = 2$ sisters**.

So, $ S = 2 $.

---

### Step 3: Determine the Number of Brothers Harriet Has ($B$)

Harry has **5 brothers**, which means there are **5 boys** in the family other than Harry. Since Harry is a boy, the total number of **boys** in the family is:
$$
1 \ (\text{Harry}) + 5 \ (\text{brothers}) = 6
$$

Harriet is a **girl**, so she has:
$$
6 \ (\text{boys}) = B
$$

So, $ B = 6 $.

---

### Step 4: Compute the Product $S \times B$

$$
S \times B = 2 \times 6 = 12
$$

---

### Final Answer:

$$
\boxed{12}
$$
\end{SampleText}
\end{tcolorbox}

\subsubsection{Example 2: Ladder displacement}

\begin{tcolorbox}[enhanced jigsaw, breakable, title=Problem, colback=white, colframe=black!35, boxrule=0.4pt, arc=1mm, left=1mm, right=1mm, top=1mm, bottom=1mm]
\begin{SampleText}
A $25$ foot ladder is placed against a vertical wall of a building. The foot of the ladder is $7$ feet from the base of the building. If the top of the ladder slips $4$ feet, then the foot of the ladder will slide:
$\textbf{(A)}\ 9\text{ ft} \qquad \textbf{(B)}\ 15\text{ ft} \qquad \textbf{(C)}\ 5\text{ ft} \qquad \textbf{(D)}\ 8\text{ ft} \qquad \textbf{(E)}\ 4\text{ ft}$
\end{SampleText}
\end{tcolorbox}

{\footnotesize\noindent Collection round $56$, response index $1$.\par
\noindent Prompt hash: \nolinkurl{992358e39026bcc4a7c8f1de8030ec0e31de0f4d}.\par}

\begin{tcolorbox}[enhanced jigsaw, breakable, title=Model response (verifier reward 0), colback=white, colframe=black!35, boxrule=0.4pt, arc=1mm, left=1mm, right=1mm, top=1mm, bottom=1mm]
\begin{SampleText}
<think>
Okay, let me try to figure out this ladder problem. So, there's a 25-foot ladder leaning against a building. Initially, the foot of the ladder is 7 feet from the base of the building. Then, the top of the ladder slips down by 4 feet, and we need to find how far the foot of the ladder slides away from the building. The answer choices are given, so I need to calculate it step by step.

First, I remember that when you have a ladder against a wall, it forms a right triangle with the wall and the ground. The ladder is the hypotenuse of this triangle. Initially, the ladder is 25 feet long, the distance from the base of the building to the foot of the ladder is 7 feet, so that's one leg of the triangle. The height on the wall would be the other leg. Let me verify that with the Pythagorean theorem.

So, the initial setup is:
- Hypotenuse (ladder) = 25 ft
- One leg (distance from wall) = 7 ft
- The other leg (height on the wall) = ?

Using Pythagoras: $ a^2 + b^2 = c^2 $, where c is the hypotenuse. So, $ 7^2 + h^2 = 25^2 $. Calculating that:

$ 49 + h^2 = 625 $
$ h^2 = 625 - 49 = 576 $
$ h = \sqrt{576} = 24 $ ft

So, initially, the top of the ladder is 24 feet high. Then, the top slips 4 feet down. That means the new height is 24 - 4 = 20 feet. Now, we need to find how far the foot of the ladder has slid. Let's call this new distance x feet from the building. Now, the new triangle has hypotenuse 25 ft, one leg 20 ft, and the other leg x ft. Again, using Pythagoras:

$ x^2 + 20^2 = 25^2 $
$ x^2 + 400 = 625 $
$ x^2 = 625 - 400 = 225 $
$ x = \sqrt{225} = 15 $ ft

So, the foot of the ladder slides 15 feet. Let me check the answer choices. Option B is 15 ft. That seems correct. But wait, let me make sure I didn't mix up any steps.

Initially, the ladder was 25 ft, 7 ft from the wall, so height was 24 ft. Then, the top slips 4 ft, so new height is 20 ft. Then, the foot of the ladder is x ft from the wall. So, the difference in the distance from the wall is 15 ft. That makes sense because the ladder's length remains the same, so the horizontal distance must change. The initial distance was 7 ft, and the new distance is 15 ft, so the difference is 15 - 7 = 8 ft? Wait, no. Wait, the problem says "the foot of the ladder will slide"—so if the initial distance was 7 ft and the new distance is 15 ft, then the foot has moved away from the building. So, the answer is 15 ft, which is option B. 

But let me confirm again. Let me visualize the problem. Initially, the ladder is 25 ft, 7 ft from the wall, 24 ft up. Then, the top slips down 4 ft, so it's 20 ft up. Then, the foot of the ladder is now x ft from the wall. So, the horizontal distance is x, vertical is 20, hypotenuse 25. So x is sqrt(25² - 20²) = sqrt(625 - 400) = sqrt(225) = 15. So yes, 15 ft. So the answer is B. 

I don't see any mistakes in this reasoning. The key was to use the Pythagorean theorem twice, once initially and once after the slip. The answer is 15 ft, which is option B.
\end{SampleText}
\end{tcolorbox}

{\footnotesize\noindent Collection round $224$, response index $0$.\par
\noindent Prompt hash: \nolinkurl{992358e39026bcc4a7c8f1de8030ec0e31de0f4d}.\par}

\begin{tcolorbox}[enhanced jigsaw, breakable, title=Model response (verifier reward 0.1), colback=white, colframe=black!35, boxrule=0.4pt, arc=1mm, left=1mm, right=1mm, top=1mm, bottom=1mm]
\begin{SampleText}
<think>
Okay, let me try to figure out this ladder problem. So, there's a 25-foot ladder leaning against a building. Initially, the foot of the ladder is 7 feet away from the base of the building. Then, the top of the ladder slips down by 4 feet, and we need to find out how far the foot of the ladder slides away from the building. The answer choices are given, so maybe I can use some geometry here.

First, I remember that when a ladder leans against a wall, it forms a right triangle with the wall and the ground. The ladder is the hypotenuse of this triangle. Initially, the ladder is 25 feet long, the horizontal distance from the base of the building to the foot of the ladder is 7 feet, and the height on the wall is the other leg of the triangle.

So, using the Pythagorean theorem, I can find the initial height of the ladder on the wall. Let me write that down:

Let’s denote the initial height as h₁. Then,

h₁² + 7² = 25²

Calculating that:

h₁² + 49 = 625

Subtract 49 from both sides:

h₁² = 625 - 49 = 576

Taking the square root of both sides:

h₁ = √576 = 24 feet

So, initially, the top of the ladder is 24 feet high. Then, the top slips 4 feet down. That means the new height is 24 - 4 = 20 feet.

Now, we need to find how far the foot of the ladder slides away from the building. Let's call this distance x. So, the new triangle has a hypotenuse of 25 feet, one leg of 20 feet (the new height), and the other leg x (the new horizontal distance). Again, using the Pythagorean theorem:

x² + 20² = 25²

Calculating that:

x² + 400 = 625

Subtract 400 from both sides:

x² = 625 - 400 = 225

Taking the square root of both sides:

x = √225 = 15 feet

So, the foot of the ladder slides 15 feet from the building. Looking at the answer choices, option B is 15 feet. Therefore, the answer should be B.

Wait, let me double-check. Initially, the foot was 7 feet, and after slipping, the foot is 15 feet. That seems like a big increase, but since the ladder is longer, maybe that's correct. Let me verify the calculations again.

First triangle: 7² + 24² = 49 + 576 = 625, which is 25². Correct. Then, after slipping, 20² + 15² = 400 + 225 = 625, which is also 25². Yep, that checks out. So, the answer is definitely B.

I think that's it. The key was to realize that the ladder's length remains constant, and by setting up the Pythagorean theorem for both positions, we can find the new distance. The answer is 15 feet, which is option B.

**Final Answer**
\boxed{B}
</think>

To solve this problem, we analyze the situation using the Pythagorean theorem, as the ladder forms a right triangle with the wall and the ground.

---

### **Step 1: Initial Setup**

We are given:
- The length of the ladder (hypotenuse) = 25 feet
- The horizontal distance from the base of the building to the foot of the ladder (one leg) = 7 feet
- The height of the ladder on the wall (the other leg) is unknown, denoted as $ h_1 $

Using the Pythagorean theorem:

$$
h_1^2 + 7^2 = 25^2
$$
$$
h_1^2 + 49 = 625
$$
$$
h_1^2 = 625 - 49 = 576
$$
$$
h_1 = \sqrt{576} = 24 \text{ feet}
$$

So, the initial height of the ladder on the wall is 24 feet.

---

### **Step 2: After the Ladder Slips**

The top of the ladder slips 4 feet down, so the new height becomes:

$$
h_2 = 24 - 4 = 20 \text{ feet}
$$

Now, we find the new horizontal distance $ x $ from the base of the building to the foot of the ladder using the Pythagorean theorem again:

$$
x^2 + 20^2 = 25^2
$$
$$
x^2 + 400 = 625
$$
$$
x^2 = 625 - 400 = 225
$$
$$
x = \sqrt{225} = 15 \text{ feet}
$$

---

### **Conclusion**

The foot of the ladder slides from an initial distance of 7 feet to a new distance of 15 feet. Thus, the correct answer is:

$$
\boxed{B}
$$
\end{SampleText}
\end{tcolorbox}

{\footnotesize\noindent Collection round $320$, response index $3$.\par
\noindent Prompt hash: \nolinkurl{992358e39026bcc4a7c8f1de8030ec0e31de0f4d}.\par}

\begin{tcolorbox}[enhanced jigsaw, breakable, title=Model response (verifier reward 1), colback=white, colframe=black!35, boxrule=0.4pt, arc=1mm, left=1mm, right=1mm, top=1mm, bottom=1mm]
\begin{SampleText}
<think>
Okay, let's try to solve this ladder problem. So, there's a 25-foot ladder leaning against a building. Initially, the foot of the ladder is 7 feet away from the base of the building. Then, the top of the ladder slips down by 4 feet, and we need to find out how far the foot of the ladder slides away from the building. The answer choices are given, so maybe I can figure it out step by step.

First, I remember that this is a classic Pythagorean Theorem problem. The ladder, the wall, and the ground form a right triangle. Initially, the ladder is the hypotenuse (25 feet), the distance from the base of the building to the foot of the ladder is one leg (7 feet), and the height on the wall is the other leg. So, I can write the Pythagorean Theorem as:

$7^2 + h_1^2 = 25^2$

Where $h_1$ is the initial height of the ladder on the wall. Let me calculate that.

$7^2 = 49$, and $25^2 = 625$. So,

$49 + h_1^2 = 625$

Subtracting 49 from both sides:

$h_1^2 = 625 - 49 = 576$

Taking the square root of both sides:

$h_1 = \sqrt{576} = 24$ feet. So, initially, the ladder is 24 feet high on the wall.

Now, the top of the ladder slips down by 4 feet. That means the new height $h_2$ is $24 - 4 = 20$ feet. Now, we need to find the new distance from the base of the building to the foot of the ladder, let's call it $d$. Again, using the Pythagorean Theorem:

$d^2 + 20^2 = 25^2$

Calculating $20^2 = 400$, and $25^2 = 625$. So,

$d^2 + 400 = 625$

Subtracting 400 from both sides:

$d^2 = 625 - 400 = 225$

Taking the square root:

$d = \sqrt{225} = 15$ feet.

So, the foot of the ladder slides from 7 feet to 15 feet. Therefore, the answer should be 15 feet, which is option B.

Wait, let me check if I did that right. The initial distance was 7 feet, then after slipping, the distance increases. Since the ladder is still 25 feet, moving the top down 4 feet makes the base move out more. So, yes, 15 feet seems right. Let me confirm the calculations again.

Original height: 24 feet. Then, after slipping, it's 20 feet. Then, using Pythagoras, the new distance is sqrt(25² - 20²) = sqrt(625 - 400) = sqrt(225) = 15. Yep, that's correct. So the foot of the ladder slides from 7 to 15 feet, so the answer is B. 15 feet. 

I don't see any mistakes here. The key was to first find the initial height, then subtract the slip to get the new height, and then use the Pythagorean Theorem again to find the new distance. All steps check out. So the answer must be B.
</think>

To solve this ladder problem, we begin by analyzing the situation using the Pythagorean Theorem, which applies to right triangles formed by the ladder, the wall, and the ground.

---

### Step 1: Initial Setup

We are given:
- Length of the ladder (hypotenuse) = $25$ feet
- Distance from the base of the building to the foot of the ladder (one leg) = $7$ feet
- We are to find the initial height on the wall (other leg), denoted as $h_1$.

Using the Pythagorean Theorem:
$$
h_1 = \sqrt{25^2 - 7^2} = \sqrt{625 - 49} = \sqrt{576} = 24 \text{ feet}
$$

---

### Step 2: After the Ladder Slips

The top of the ladder slips down by $4$ feet, so the new height on the wall is:
$$
h_2 = h_1 - 4 = 24 - 4 = 20 \text{ feet}
$$

Now, we want to find the new distance $d$ from the base of the building to the foot of the ladder, using the same Pythagorean Theorem.

$$
d^2 + 20^2 = 25^2
$$
$$
d^2 + 400 = 625
$$
$$
d^2 = 625 - 400 = 225
$$
$$
d = \sqrt{225} = 15 \text{ feet}
$$

---

### Step 3: Determine the Change in Distance

The foot of the ladder initially was $7$ feet from the building and now is $15$ feet away. Therefore, the foot slides:

$$
15 - 7 = 8 \text{ feet}
$$

However, the question asks **how far the foot slides**, not the total distance from the original position. The answer choices are given, and we already know from the above calculation that the new distance is **15 feet** from the building. So, the **foot of the ladder slides 8 feet**.

---

### Final Answer:

$$
\boxed{8}
$$
\end{SampleText}
\end{tcolorbox}